\documentclass[twoside,11pt]{article}

\usepackage{jmlr2e_min}
\usepackage{amsmath, amsfonts, amssymb, bm, mathtools, bbm}
\usepackage{algorithm}
\usepackage{algorithmic}
\usepackage{enumerate}
\usepackage[shortlabels]{enumitem}
\usepackage{subcaption}

\DeclareMathOperator*{\argmax}{arg\,max}
\DeclareMathOperator*{\argmin}{arg\,min}
\renewcommand{\vec}[1]{\mathbf{#1}}

\usepackage{color}

\renewcommand{\tilde}{\widetilde}
\newcommand{\indep}{\perp \!\!\! \perp}

\usepackage{fancyhdr}
\fancypagestyle{FooBar}{%
    \fancyhead{}
    
}

\ShortHeadings{Sparsity-Agnostic Lasso Bandit}{Oh, Iyengar and Zeevi}
\firstpageno{1}

\begin{document}
\thispagestyle{FooBar}

\title{Sparsity-Agnostic Lasso Bandit}

\author{\name Min-hwan Oh \email minoh@snu.ac.kr \\
       \addr Seoul National University\\
       Seoul, 08826, South Korea
       \AND
       \name Garud Iyengar \email garud@ieor.columbia.edu \\
       \addr Columbia University\\
       New York, NY 10027, USA
       \AND
       \name Assaf Zeevi \email assaf@gsb.columbia.edu \\
       \addr Columbia University\\
       New York, NY 10027, USA
        }

\editor{ }

\maketitle

\begin{abstract}%
We consider a stochastic contextual bandit problem where the dimension $d$ of the feature vectors is potentially large, however, only a sparse subset of features of cardinality $s_0 \ll d$  affect the reward function.  
Essentially all existing algorithms for sparse bandits require a priori knowledge  of the value of the sparsity index $s_0$. This knowledge is almost never available in practice, and misspecification of this parameter can lead to severe deterioration in the performance of existing methods. 
The main contribution of this paper is to propose an algorithm that does \emph{not} require  prior knowledge of the sparsity index $s_0$ and establish tight regret bounds on its performance under mild conditions. 
We also comprehensively evaluate our proposed algorithm numerically and show that it consistently outperforms  existing methods, even when the correct sparsity index is revealed to them but is kept hidden from our algorithm.  
\end{abstract}


\section{Introduction}
In classical multi-armed bandits (MAB), one of the arms is pulled in each
round and  a reward corresponding to the chosen arm is revealed to the
decision-making agent.  The rewards are, typically, 
independent and identically distributed samples from an arm-specific
distribution.
The goal of the agent is to devise a strategy for pulling arms that maximizes
cumulative rewards, suitably balancing
between exploration and exploitation. Linear 
contextual bandits \citep{abe1999associative, auer2002using,
  chu2011contextual} and generalized linear contextual bandits
\citep{filippi2010parametric, li2017provably} are more recent important 
extensions of the basic MAB setting, where 
each arm $a$
is associated with a known feature vector $x_a \in \mathbb{R}^d$, and the
expected payoff of the arm 
is 
a (typically, monotone increasing)
function of the inner product $x_a^\top \beta^*$ for a fixed and unknown
parameter vector $\beta^* \in \mathbb{R}^d$. Unlike the traditional MAB problem, here
pulling
any one arm provides some information about the unknown parameter vector, and
hence, insight into the average reward of all the other arms. These contextual bandit
algorithms are applicable in a variety of problem settings, such as
recommender systems, assortment selection in online retail, and healthcare
analytics \citep{li2010contextual,oh2019thompson,tewari2017ads}, 
where the contextual information can be used for personalization and generalization.  

In most application domains highlighted above, the feature space is 
high-dimensional $(d \gg 1)$, yet typically only a
small subset of the features 
influence the expected reward. 
That is, the unknown parameter vector is \textit{sparse} with only elements
corresponding to the relevant features being non-zero,
i.e., the \emph{sparsity index} $s_ 0 = \| \beta^*\|_0  \ll d$,
where the zero norm $\|x\|_0$ counts non-zero entries in the vector $x$.
There is an emerging  body of literature on contextual bandit problems with sparse
linear reward functions \citep{abbasi2012online, gilton2017sparse,bastani2020online, wang2018minimax, kim2019doubly} which propose
methods to exploit the sparse structure under various conditions.  
However, there is a crucial shortcoming in almost all of these approaches:
the algorithms require \emph{prior}  knowledge of the sparsity index $s_0$,
information that is almost never available in practice.  
In the absence of such knowledge, the existing algorithms fail to fully leverage
the sparse structure, and their performance does not guarantee the
improvements in dimensionality-dependence which can be realized in the
sparse problem setting (and can lead to extremely poor performance if $s_0$ is underspecified).   
The purpose of this paper is to demonstrate that a relatively simple contextual bandit
algorithm that exploits $\ell_1$-regularized regression using 
Lasso \citep{tibshirani1996regression}
in a sparsity-agnostic manner,  is provably near-optimal insofar as its
regret performance (under suitable regularity). 
Our contributions are as follows:
\begin{enumerate}[(a)]
    \item We propose the first general
    sparse bandit 
      algorithm that does not require prior knowledge of the
      sparsity index $s_0$.     
    \item We establish that the regret bound
      of our proposed algorithm is $\mathcal{O}( s_0\sqrt{T \log (dT)} )$ for the two-armed case, which affords the most accessible exposition of the key analytical ideas. (Extensions to the general $K$-armed case are discussed later.)
      The regret bound scale in $s_0$ and $d$
      matches the equivalent terms in the \textit{offline} Lasso results
      (see the discussions in Section~\ref{sec:regret}). 
    \item We comprehensively evaluate our algorithm on numerical
      experiments and show that it consistently outperforms existing
      methods, even when these methods are granted prior knowledge of the
      correct 
      sparsity index (and can greatly outperform them if this information
      is misspecified).   
\end{enumerate}
The salient feature of our algorithm is that it does not rely on \emph{forced sampling} which was used by almost all 
previous work, e.g.,
\citet{bastani2020online, wang2018minimax, kim2019doubly}, 
to satisfy certain regularity of the empirical
Gram matrix. Forced sampling requires prior knowledge of
$s_0$ because such schemes, 
the key ideas of which go back to \citet{goldenshluger2013linear}, need to
be fine-tuned using  the {\it correct} sparsity index.  (See further
discussions in Section~\ref{sec:why_need_to_know_sparsity}.) 

The rest of the paper is organized as follows. In Section~\ref{sec:related_work}, 
we review the related literature and discuss the reason why the previously
proposed methods require 
knowledge of the
sparsity index~$s_0$.  
In Section~\ref{sec:prelim}, we present the problem
formulation. Section~\ref{sec:algorithm} describes our proposed
algorithm. 
In Section~\ref{sec:regret_analysis}, we 
describe the challenges 
when 
the sparsity information is unknown, and establish an  upper bound on the cumulative
regret for the two-armed sparse bandits.
Section~\ref{sec:experiments} contains the numerical experiments for the
two-armed sparse bandits. In 
Section~\ref{sec:K-arms}, we extend our analysis and numerical evaluations
to the $K$-armed sparse bandits. Section~\ref{sec:conclusion} presents
discussions and future directions. The complete proofs and additional
numerical results are provided in the appendix.

\section{Related Work}\label{sec:related_work}

\subsection{Review}

Linear bandits and generalized linear bandits  have been widely studied
\citep{abe1999associative, auer2002using, dani2008stochastic,
  rusmevichientong2010linearly, abbasi2011improved, filippi2010parametric,
  chu2011contextual, agrawal2013thompson, li2017provably, kveton2020randomized}. However, when
ported to the high-dimensional contextual bandit setting, these strategies
have difficulty exploiting 
sparse structure in the  unknown parameter vector, and hence, may incur
regret proportional to the full ambient dimension $d$ rather than the
sparse set of features of cardinality $s_0$.  
To exploit spare structure, \citet{abbasi2012online} propose a framework
to construct high probability confidence sets for online linear prediction
and establish a regret bound of $\tilde{\mathcal{O}}(\sqrt{s_0 d T})$,
where $\tilde{\mathcal{O}}$ hides logarithmic terms, when 
the sparsity index $s_0$ is \emph{known}.  
Furthermore,
their algorithm is not computationally efficient; an implementable version
of their framework is not yet known (Section 23.5 in  \citealt{LS19bandit-book}). It is worth noting that the
$\sqrt{d}$ dependence in the regret bound is unavoidable unless additional
assumptions are imposed; see Theorem 24.3 in 
\citet{LS19bandit-book}. \citet{gilton2017sparse} adapt 
Thompson sampling \citep{thompson1933likelihood} to sparse linear bandits;
however, they also assume a priori knowledge of a small superset of the
support for the parameter.

\citet{bastani2020online} address the contextual bandit problem with
high-dimensional features by using Lasso
\citep{tibshirani1996regression} to estimate the parameter of each arm
separately. To ensure compatibility of the empirical Gram matrices, they adapt the forced-sampling technique in \citet{goldenshluger2013linear} which is now tuned using the (a priori known) sparsity index, and is implemented for each arm at predefined time
points. They establish a regret bound of $\mathcal{O}( K s^2_0[\log d + \log T]^2)$ 
where $K$ is the number of arms. 
Note that they invoke several additional assumptions introduced in \citet{goldenshluger2013linear}, including 
a margin condition that ensures that the density of the context distribution is  bounded near the decision boundary,  and arm-optimality, which assumes a gap between the optimal and sub-optimal arms exists with some positive probability.  
In the same problem setting, \citet{wang2018minimax} propose an algorithm which uses forced-sampling
along with the  minimax concave penalty (MCP) estimator
\citep{zhang2010nearly} and improve the regret bound to $\mathcal{O}(K
s^2_0[s_0 + \log d] \log T)$. Note that \citet{bastani2020online} and
\citet{wang2018minimax} achieve a poly-logarithmic dependence on~$T$ in the regret, exploiting the arm optimality condition which assumes a gap between
the optimal and sub-optimal arms exists  with some probability.\footnote{The regret bounds in both \citet{bastani2020online} and \citet{wang2018minimax} have additional dependence $\mathcal{O}(1/p_*^3)$ 
where $p_*$ is the arm optimality lower bounding probability. Hence, in the worse case, the regret bounds have additional $\mathcal{O}(K^3)$ dependence.}
Since we do not assume such ``separability'' between arms, poly-logarithmic dependence on~$T$ is not attainable in our problem setting.
\citet{kim2019doubly} extend the Lasso bandit \citep{bastani2020online} to
linear bandit settings and propose a different approach to  address the
non-compatibility of the empirical Gram matrices by using a doubly-robust
technique \citep{bang2005doubly} that originates with the missing data (imputation) literature. They achieve $\mathcal{O}(s_0 \sqrt{T}\log(dT))$
regret. 

All of the aforementioned algorithms require that the
learning agent know the sparsity index $s_0$ of the unknown parameter (or a
non-trivial upper-bound on sparsity which is strictly less than
$d$).\footnote{Besides sparsity, some algorithms require further knowledge,
  such as arm optimality lower bounding probability
  \citep{bastani2020online, wang2018minimax}, which is also not readily available in practice.}
  That is, only when the
algorithm knows $s_0$, it can guarantee the regret bounds mentioned
above. Otherwise, the regret bounds would scale polynomially with $d$
instead of $s_0$ or potentially scale linearly with $T$. 
To our knowledge, the only work in sparse bandits which does not require
this prior knowledge of sparsity is the work by
\citet{carpentier2012bandit} although their algorithm still requires knowledge of 
the $\ell_2$-norm of the unknown parameter. Their analysis uses a
non-standard definition of noise and is restricted to the case where the
set of arms is the $\ell_2$ unit ball and fixed over time, a structure
they exploit in a significant manner, and which limits the scope of their
algorithm.  

\subsection{Why do existing sparse bandit algorithms require prior knowledge of the sparsity index?}\label{sec:why_need_to_know_sparsity}

The primary reason that a  priori  knowledge of sparsity index $s_0$ is assumed
throughout most of the literature is, roughly speaking, to ensure suitable
``size'' of confidence bounds and concentration.  For example,
\citet{abbasi2012online} require  the parameter $s_0$ to explicitly
construct a high probability confidence set with its radius proportional
to $s_0$ rather than $d$.   The recently proposed bandit algorithms of
\citet{bastani2020online, kim2019doubly} and the variant with MCP estimator in
\citet{wang2018minimax} employ a logic that is similar in spirit (though
different in execution). Specifically,   
 the compatibility condition is assumed
 to hold only for the theoretical Gram matrix, and the empirical Gram
 matrix may not satisfy such condition (the difficulty in controlling that
 is due to the non-i.i.d.~adapted samples  of the feature variables). As a
 remedy to this issue,  \citet{bastani2020online} and
 \citet{wang2018minimax} utilize the forced-sampling technique of
 \citet{goldenshluger2013linear} to obtain a ``sufficient'' number of
 i.i.d.~samples and use them to  show that the empirical Gram matrices
 concentrate in the vicinity of  the theoretical Gram matrix, and hence,
 satisfy the compatibility condition after a sufficient amount of
 forced-sampling. The forced-sampling duration needs to be predefined and
 scales at least polynomially in the sparsity index $s_0$ to ensure
 concentration of the Gram matrices. That is, if the algorithm does not
 know $s_0$, the forced-sampling duration will have to scale polynomially in
 $d$. 
 \citet{kim2019doubly} propose an alternative to forced sampling that
builds on doubly-robust techniques used in the  missing data literature;  
however, their algorithm
involves random arm selection with a probability that is calibrated using
$s_0$,  and initial uniform sampling whose duration  requires knowledge
of $s_0$ and scales polynomially with $s_0$, in order to establish their
regret bounds. The sensitivity to the sparsity index specification is
also evident in cases where its value is {\it misspecified}, which may
result in severe deterioration in the performance of the algorithms (see
further discussions in Section 5.1).   

The key observation in our analysis is that
i.i.d.~samples, which are the key output of the forced  samplings scheme,
are, in fact, \emph{not} required under some mild regularity conditions. 
We show that the  empirical Gram matrix
satisfies the required regularity after a sufficient number of rounds,
provided the theoretical Gram matrix is also regular; the
details of this analysis are in Section~\ref{sec:regret_analysis}. 
Numerical experiments support these findings, and moreover, demonstrate
that the performance of our proposed sparsity-agnostic algorithm can be superior to
forced-sampling-based schemes that 
are tuned with foreknowledge of
the sparsity index~$s_0$.  

\section{Preliminaries}\label{sec:prelim}

\subsection{Notation}
For a vector $x \in \mathbb{R}^d$, we use $\| x \|_1$ and $\| x \|_2$ to
denote its $\ell_1$-norm and $\ell_2$ norm respectively, the notation
$\|x\|_0$ is reserved for the cardinality of the set of non-zero entries
of that vector.  The minimum and maximum singular values of a matrix $V$
are written as $\lambda_{\min}(V)$ and $\lambda_{\max}(V)$ respectively.  
For two symmetric matrices $V$ and $W$ of the same dimensions, $V
\succcurlyeq W$ means that $V - W$ is positive semi-definite. 
For a positive integer $n$, we define $[n] = \{1,  ... , n\}$.
For a real-valued differentiable function $f$, we
use $\dot{f}$  to denote its first derivative.

\subsection{Generalized Linear Contextual Bandits}

We consider the stochastic generalized linear bandit problem with $K$
arms. Let $T$ be the problem horizon, namely the number of  rounds to be
played. In each round $t \in [T]$, the learning agent observes a context
consisting of a set of $K$ feature vectors $\mathcal{X}_t = \left\{X_{t,i} \in \mathbb{R}^d
  \mid i \in [K]\right\} $, where the tuple
$\mathcal{X}_t$ is drawn i.i.d.~over $t \in [T]$ from an unknown joint
distribution with probability density 
$p_\mathcal{X}$ with respect to the Lebesgue measure. Note that the
feature vectors for different arms are 
allowed to be correlated.  
Each feature vector $X_{t,i}$ is associated with an unknown stochastic
reward $Y_{t,i} \in \mathbb{R}$. The agent then selects one arm, denoted by
$a_t \in [K]$ and observes the reward $Y_t := Y_{t, a_t}$,
corresponding to the chosen arm's feature $X_t := X_{t,a_t}$, 
as a bandit feedback. 
The policy consists of the sequence of actions $\pi = \{a_t : t = 1, 2,...\}$ 
and is non-anticipating, namely each action only depends on past observations and actions.

In this work, we assume that the reward $Y_{t,i}$ of arm $i$ is given by a generalized linear model (GLM), i.e.
\begin{align*}
    Y_{t,i} = \mu(X_{t,i}^\top \beta^*) + \epsilon_{t,i}
\end{align*}
where $\mu: \mathbb{R} \rightarrow \mathbb{R}$ (also known
as  
\textit{inverse link function}) is a \emph{known} increasing function, $\beta^* \in \mathbb{R}^d$ is an \emph{unknown} parameter, and  each $\epsilon_{t,i}$ is an independent zero-mean noise.
Therefore, $\mathbb{E}[Y_{t,i}|X_{t,i} = x] = \mu(x^\top \beta^*)$  for all $i \in [K]$ and $t \in [T]$.
Widely used examples for $\mu$ are $\mu(z) = z$ which corresponds to the linear model,
and $\mu(z) = 1/(1+e^{-z})$ which corresponds to the logistic model.   
The parameter $\beta^*$ and the feature vectors $\{X_{t,i}\}$ are
potentially high-dimensional, i.e., $d \gg 1$,  but $\beta^*$  is
\textit{sparse}, that is, the number of non-zero elements in $\beta^*$, 
$s_0 = \|\beta^*\|_0 \ll d$. 
It is important to note that the agent \textit{does not} know $s_0$ or the
support of the unknown parameter~$\beta^*$.

We assume that there is an increasing sequence of sigma fields
$\{\mathcal{F}_t\}$ such that each $\epsilon_{t,i}$ is
$\mathcal{F}_t$-measurable with $\mathbb{E}[\epsilon_{t,i} |
\mathcal{F}_{t-1}] = 0$. In our problem, $\mathcal{F}_t$ is the sigma-field generated
by random variables of chosen actions $\{a_1, ..., a_{t}\}$, their features $\{X_{1,a_1}, ..., X_{t,a_t}\}$, and the corresponding rewards $\{Y_{1,a_1}, ..., Y_{t,a_t}\}$.
We assume the noise $\epsilon_{t,i}$ for all $i \in [K]$ is sub-Gaussian with
parameter $\sigma$, where $\sigma$ is a positive absolute  constant, i.e.,
$\mathbb{E}[e^{\alpha \epsilon_{t,i}}] \leq e^{\alpha^2 \sigma^2/2}$ for all
$\alpha \in \mathbb{R}$. In practice, for bounded reward $Y_{t,i}$,
the noise $\epsilon_{t,i}$ is also bounded and hence satisfies the
sub-Gaussian assumption with an appropriate $\sigma$ value.  

The agent’s goal is to maximize the cumulative expected reward $\mathbb{E}
[\sum_{t=1}^T \mu(X_{t,a_t}^\top \beta^*)]$ over $T$
rounds. 
Let $a^*_t =
\argmax_{i \in [K]} \big\{\mu(X_{t,i}^\top \beta^*)\big\}$ denote the optimal arm for each
round~$t$. Then, the expected cumulative {\it
  regret} of policy $\pi = \{a_1, ..., a_T\}$ is defined as
\begin{align*}
    \mathcal{R}^\pi(T) :=  \sum_{t=1}^T \mathbb{E}\left[ \mu(X_{t,a^*_t}^\top \beta^*) - \mu(X_{t,a_t}^\top \beta^*) \right] \,.
\end{align*}
Hence, maximizing the expected cumulative rewards of policy $\pi$ over $T$ rounds is
equivalent to minimizing the cumulative regret  $\mathcal{R}^\pi(T)$. 
Note that all the expectations and probabilities throughout the paper are with respect to feature vectors and noise unless explicitly stated otherwise.

\subsection{Lasso for Generalized Linear Models}
Consider an offline setting where
we have samples $Y_1, ..., Y_n$ and corresponding features   $X_1, ..., X_n$. The log-likelihood function of $\beta$ under the canonical GLM is
\begin{align*}
     \log\mathcal{L}_n(\beta) &:= \sum_{j=1}^n \left[ \frac{Y_j X_j^\top \beta - m(X_j^\top \beta)}{g(\eta)} - h(Y_j, \eta) \right] \,.
\end{align*}
Here, $\eta \in \mathbb{R}_+$ is a known scale parameter, $m(\cdot)$,
$g(\cdot)$ and $h(\cdot)$ are normalization functions,  and $m(\cdot)$ is
infinitely differentiable with the first derivative
\begin{align*}
\dot{m}(x^\top \beta^*) = \mathbb{E}[Y | X = x] = \mu(x^\top \beta^*) \,.    
\end{align*}
The Lasso \citep{tibshirani1996regression} estimate for the GLM
can be defined as
\begin{align}\label{eq:Lasso_glm}
    \hat{\beta}_n \in \argmin_{\beta } \big\{ \ell_n(\beta) + \lambda \|\beta\|_1 \big\}  
\end{align}
where $\ell_n(\beta) := -\frac{1}{n}\sum_{j=1}^n \left[ Y_j X_j^\top \beta - m(X_j^\top \beta) \right] $ and $\lambda > 0$ is a penalty parameter.
Lasso is  known to be an efficient (offline) tool for estimating the
high-dimensional linear regression parameter. The ``fast convergence''
property of Lasso is guaranteed when the above data are i.i.d.~and when the observed
covariates are not ``highly correlated.'' The restricted eigenvalue condition
\citep{bickel2009simultaneous, raskutti2010restricted}, the compatibility
condition \citep{van2009conditions}, and the restricted isometry property
\citep{candes2007dantzig} 
have all
been used to ensure that such high
correlations are  avoided. In sequential learning settings, however, these
conditions are often violated because the observations are adapted to the
past, and the feature variables of the chosen arms converge to a small
region of the  feature space as the learning agent updates its arm
selection policy.

\section{Proposed Algorithm}\label{sec:algorithm}
Our proposed 
\textsc{Sparsity-Agnostic (SA) Lasso Bandit} algorithm for high-dimensional 
GLM bandits is summarized in Algorithm~\ref{algo:SA_Lasso_bandit}.
As the name suggests, 
our algorithm does not require prior knowledge of the sparsity index~$s_0$. 
It relies on Lasso for parameter estimation, and  does
not explicitly use exploration strategies or forced-sampling. Instead, in
each round, we choose an arm which maximizes the inner product of a feature
vector and the Lasso estimate. After observing the reward, we update the
regularization parameter $\lambda_t$ and update the Lasso estimate
$\hat{\beta}_t$ which minimizes the penalized negative log-likelihood
function defined in \eqref{eq:Lasso_glm}.  

\textsc{SA Lasso Bandit} requires only one input parameter
$\lambda_0$. We show in Section~\ref{sec:regret_analysis} that $\lambda_0
= 2\sigma x_{\max}$ where $x_{\max}$ is  a bound on the $\ell_2$-norm
of the feature vectors $X_{t,i}$. Thus, $\lambda_0$ does  \emph{not}
depend on the sparsity index $s_0$ or the underlying parameter $\beta^*$.
(Note that, in comparison, \citet{kim2019doubly} require
three tuning parameters, 
and \citet{bastani2020online} and \citet{wang2018minimax}
require four tuning parameters, 
most of which are functions of the unknown sparsity index $s_0$.) 
It is worth noting that
tuning parameters, while helping to achieve low regret, are challenging to specify in online learning settings. In contrast, our proposed algorithm
is practical and easy to implement.

\begin{algorithm}
\begin{algorithmic}[1]
\caption{\textsc{SA Lasso Bandit}}
\label{algo:SA_Lasso_bandit}
    \STATE \textbf{Input parameter}: $\lambda_0$
    \FOR{all $t = 1$ to $T$}
        \STATE Observe $X_{t,i}$ for all $i \in [K]$
        \STATE Compute $a_t = \argmax_{i \in [K]} X^\top_{t,i} \hat{\beta}_t$ 
        \STATE Pull arm $a_t$ and observe $Y_t$
        \STATE Update $\lambda_t \leftarrow \lambda_0 \sqrt{\frac{4\log t + 2\log d}{t}}$
        \STATE Update $\hat{\beta}_{t+1} \leftarrow \argmin_{\beta}\left\{ \ell_t(\beta) + \lambda_t \|\beta\|_1 \right\}$ 
    \ENDFOR
\end{algorithmic}
\end{algorithm}

\noindent \textbf{Discussion of the algorithm.}
Algorithm~\ref{algo:SA_Lasso_bandit} may appear to be an
\emph{exploration-free} greedy algorithm (see e.g., \citealt{bastani2020mostly}), 
but this is not the case. 
To better understand this we will compare the steps in Algorithm~\ref{algo:SA_Lasso_bandit} to upper-confidence bound (UCB) algorithms. A UCB algorithm constructs a high-probability confidence ellipsoid around a
\textit{greedy} maximum likelihood estimate and chooses a parameter value within the ellipse
that maximizes the reward.
Once the UCB 
estimate is chosen, the
action selection is greedy with respect to the parameter
estimate.\footnote{Likewise, in Thompson sampling \citep{thompson1933likelihood}, the
agent chooses the
greedy action for the sampled parameter.}
The UCB algorithms regularize parameter estimates by carefully controlling the size of the confidence ellipsoid
to ensure convergence, thus, exploration is loosely equivalent to
regularizing the parameter estimate.
The algorithm we propose also computes the
parameter estimate by \textit{regularizing} the MLE with a sparsifying norm, and
then, as in UCB, takes a greedy action with respect to this regularized
parameter estimate. We adjust the penalty parameter associated with the
sparsifying norm over time at carefully specified rate in order to ensure that our estimate is
consistent as we collect more samples. 
(This adjustment and specification do not require knowledge of sparsity $s_0$.)
Incorrect choice for the penalty parameter would lead to large regret,
which is analogous to poor choice of confidence widths in UCB.

\section{Regret Analysis}\label{sec:regret_analysis}

\subsection{Regularity Condition}
In this section, we establish an upper bound on the expected regret of \textsc{SA~Lasso~Bandit} 
for the two-armed ($K=2$) generalized linear bandits. We focus on the two-arm case primarily for clarity
and accessibility of key analysis ideas, and later illustrate how this analysis extends to the $K$-armed case with $K \geq 3$ under suitable regularity (see Section~\ref{sec:K-arms}). 
We first provide a few definitions and assumptions used throughout the analysis, starting with assumptions standard in the (generalized) linear bandit literature.

\begin{assumption}[Feature set and parameter]\label{assum:feature_param_bounds}
There exists a positive constant $x_{\max}$ such that $\| x\|_2 \leq
x_{\max}$ for all $x \in \mathcal{X}_t$ and all $t$, and a positive
constant $b$ such that  $\|\beta^*\|_2 \leq b$. 
\end{assumption}

\begin{assumption}[Link function]\label{assum:link_func_bounds}
There exist $\kappa_0 > 0$ and $\kappa_1 < \infty$ such that the
derivative $\dot{\mu}(\cdot)$ of  the link function satisfies
$\kappa_0 \leq \dot{\mu}(x^\top \beta) \leq \kappa_1 $ for all $x$ and $\beta$.
\end{assumption}
Clearly for the linear link function, $\kappa_0 = \kappa_1 = 1$.
For the logistic link function, we have $\kappa_1 = 1/4$.
\begin{definition}[Active set and sparsity index]
The active set  $S_0 := \{ j : \beta^*_j \neq 0\}$ is the set of indices $j$ for which
$\beta^*_j$ is non-zero, 
and the sparsity index $s_0 = |S_0|$ 
denotes the cardinality of the active set $S_0$.  
\end{definition}
For the active set $S_0$, and  an arbitrary vector $\beta \in \mathbb{R}^d$, we can  define 
\begin{align*}
    \beta_{j,S_0} := \beta_j \mathbbm{1}\{ j \in S_0\}\,, \quad \beta_{j,S^c_0} := \beta_j \mathbbm{1}\{ j \notin S_0\} \,.
\end{align*}
Thus, $\beta_{S_0} = [\beta_{1,S_0}, ..., \beta_{d,S_0}]^\top$ has zero elements
outside the set $S_0$ and the components of $\beta_{S_0^c}$ can only be
non-zero in the complement of $S_0$. 
Let  $\mathbb{C}(S_0) $ denote the set of vectors
\begin{align}\label{eq:set_C}
    \mathbb{C}(S_0) := \{ \beta \in \mathbb{R}^d \mid  \|\beta_{S_0^c}\|_1 \leq 3\|\beta_{S_0}\|_1 \} \,.
\end{align}
Let $\mathbf{X} \in \mathbb{R}^{K \times d}$ denote the design matrix 
where each row is a feature vector for an arm.  
(Although we focus on $K = 2$ case in this section, the definitions and
the assumptions introduced here also apply to the case of $K \geq 3$.) 
Then, in keeping with the previous literature on sparse estimation and
specifically on sparse bandits
\citep{bastani2020online,wang2018minimax,kim2019doubly}, we assume that
the following compatibility condition is satisfied for the theoretical Gram matrix
$\Sigma := \frac{1}{K}\mathbb{E}[\mathbf{X}^\top \mathbf{X}]$. 

\begin{assumption}[Compatibility condition]\label{assum:comp_condition} For active set $S_0$, there exists compatibility constant $\phi^2_0 > 0$ such that
\begin{align*}
 \phi_0^2 \| \beta_{S_0} \|^2_1 \leq  s_0 \beta^\top  \Sigma \beta \quad \text{for all } \beta \in \mathbb{C}(S_0) \,.
\end{align*}
\end{assumption}
We add to this the following mild assumption that is more specific to our analysis. 
\begin{assumption}[Relaxed symmetry]\label{assum:rho}
For a joint distribution $p_\mathcal{X}$, there exists $\nu < \infty$ such that $\frac{p_\mathcal{X}(-\mathbf{x})}{p_\mathcal{X}(\mathbf{x})} \leq \nu$ for all $\mathbf{x}$.
\end{assumption}

\noindent \textbf{Discussion of the assumptions.} 
Assumptions~\ref{assum:feature_param_bounds} and~\ref{assum:link_func_bounds} are the standard regularity assumptions used in the GLM bandit literature \citep{filippi2010parametric,li2017provably,kveton2020randomized}.
It is important to note that unlike the existing GLM bandit algorithms 
which explicitly use the value of $\kappa_0$, our proposed algorithm does
not use $\kappa_0$  or $\kappa_1$
--- this information is only needed to establish the regret bound.
The compatibility condition in Assumption~\ref{assum:comp_condition} is
analogous to the standard positive-definite assumption on the Gram matrix
for the ordinary least squares estimator for linear models but is less
restrictive. 
The compatibility condition 
ensures that truly active components of the parameter vector are not ``too correlated.'' As
mentioned above, the compatibility condition is a standard assumption in
the sparse bandit literature
\citep{bastani2020online,wang2018minimax,kim2019doubly}. 
Assumption~\ref{assum:rho} states that the joint distribution
$p_\mathcal{X}$ can be skewed but this skewness is bounded. Obviously, if
$p_\mathcal{X}$ is symmetrical, we have $\nu = 1$. Assumption~\ref{assum:rho} is satisfied for a large class of continuous and discrete
distributions, e.g., elliptical distributions including Gaussian and
truncated Gaussian distributions, multi-dimensional uniform distribution, and Rademacher distribution. 
Note that in the non-sparse low dimensional setting (i.e., $d = s$), the relaxed symmetry in Assumption~\ref{assum:rho}  together with
the positive definiteness of the theoretical Gram matrix is equivalent to the covariate diversity condition introduced in \cite{bastani2020mostly}. 
However, in the sparse high-dimensional setting considered here, the relaxed symmetry does not imply diversity in all covariates. 
Consequently, the greedy parameter estimation approach proposed by \cite{bastani2020mostly} is not guaranteed to achieve a sublinear regret. As in the case of $\kappa_0$ and $\kappa_1$ in Assumption~\ref{assum:link_func_bounds}, the parameter $\nu$ is only needed to establish the regret bound, our proposed algorithm does not require knowledge of $\nu$.

\subsection{Regret Bound for \textsc{SA~Lasso~Bandit}}\label{sec:regret}

\begin{theorem}[Regret bound for two arms]\label{thm:GLM-Lasso-regret-bound_CC}
Suppose $K = 2$ and Assumptions~\ref{assum:feature_param_bounds}-\ref{assum:rho} hold. 
Let $\lambda_0 = 2 \sigma x_{\max}$. Then the expected cumulative regret of the \textsc{SA~Lasso~Bandit} policy $\pi$ over horizon~$T \geq 1$  is upper-bounded by
\begin{align*}
    \mathcal{R}^\pi(T) &\leq  4\kappa_1 + \frac{4 \kappa_1 x_{\max} b (\log(2d^2)+1)}{C_0(s_0)^2} + \frac{32 \kappa_1\nu \sigma x_{\max} s_0 \sqrt{  T \log (dT) }}{\kappa_0\phi^2_0}
\end{align*}
where $C_0(s_0) = \min\!\Big(\frac{1}{2}, \frac{\phi_0^2}{256s_0 \nu x_{\max}^2} \Big)$.
\end{theorem}
\vspace{0.1cm}

\noindent \textbf{Discussion of Theorem~\ref{thm:GLM-Lasso-regret-bound_CC}.}
In terms of key problem primitives, Theorem~\ref{thm:GLM-Lasso-regret-bound_CC} establishes $\mathcal{O}( s_0\sqrt{T
  \log (dT) } )$ regret without any prior knowledge on $s_0$. 
The bound shows that the regret of our algorithm grows at most
logarithmically in feature dimension $d$.  
The key takeaway from this theorem is that \textsc{SA~Lasso~Bandit} is sparsity-agnostic and is able to achieve
``correct'' dependence on parameters $d$ and $s_0$. 
That is, based on the offline Lasso convergence results  under the compatibility
condition (e.g., Theorem~6.1 in \citealt{buhlmann2011statistics}), 
we believe that the dependence on $d$ and $s_0$ in Theorem~\ref{thm:GLM-Lasso-regret-bound_CC}
is best possible.\footnote{Since the horizon $T$ does not exist in offline Lasso results, it is not straightforward to see whether $\sqrt{T}$ dependence can be improved comparing only with the offline Lasso results. 
Clearly, without an additional assumption on the separability of the arms, we know that poly-logarithmic scalability in $T$ is not feasible.
We briefly discuss our conjecture in comparison with the lower bound result in the non-sparse linear bandits in Section~\ref{sec:regret_re} where we discuss the regret bound under the RE condition.}

The regret bound in Theorem~\ref{thm:GLM-Lasso-regret-bound_CC} is
tighter than the previously known bound in the same problem setting \citep{kim2019doubly}  although direct comparison is not
immediate, given the  difference in assumptions involved --- compared to \citet{kim2019doubly}, we require Assumption~\ref{assum:rho} whereas they assume the sparsity index $s_0$ is known.
Having said that, the numerical experiments in Section~\ref{sec:experiments}
support our theoretical claims and provide additional evidence that our
proposed algorithm compares very favorably  to other existing methods
(which are tuned with the knowledge of the correct $s_0$), and moreover,
the performance is not sensitive to the assumptions that were
imposed primarily for technical tractability purposes. Note that the input parameter $\lambda_0 = 2\sigma x_{\max}$
depends on $\sigma$ and $x_{\max}$ which are inputs required by 
all parametric bandit methods, and hence our algorithm does not require 
any additional information relative to that.

As mentioned earlier, the previous work on sparse bandits
\citep{bastani2020online,wang2018minimax,kim2019doubly} requires the
knowledge of the sparsity index $s_0$. In the absence of such knowledge, if sparsity is
underspecified, then these algorithms would suffer a regret linear in
$T$. On the other hand, if the sparsity is overspecified, the regret of
these algorithms may scale with $d$ instead of $s_0$.  Our proposed
algorithm does not require such prior knowledge, hence there is no risk of
under-specification or over-specification, and yet our analysis provides a sharper regret guarantee.  
Furthermore, our result also suggests that even when the sparsity is known, random sampling to satisfy the compatibility condition, invoked by all existing sparse bandit algorithms
to date, can be wasteful since said conditions may be already satisfied even in the absence of such sampling.  
This finding is also supported by the numerical experiments in Section~\ref{sec:experiments} and Section~\ref{sec:experiments_K-arms}.
We provide the outline of the proof and the key lemmas 
in the following section.

\subsection{Challenges and Proof Outlines}\label{sec:challenges_proof_outlines}
There are two essential challenges that prevent us from 
fully benefiting from the fast convergence property of Lasso:
\begin{enumerate}[label=(\roman*)]
    \item The samples induced by our bandit policy are not i.i.d., therefore the standard Lasso oracle
      inequality does not hold. 
    \item Empirical Gram matrices do not necessarily satisfy the
      compatibility condition even under Assumption~\ref{assum:comp_condition}. This is because the selected feature
      variables for which the rewards are observed do not provide an
      ``even'' representation for the entire distribution.  
\end{enumerate}
To resolve (i), we provide a Lasso oracle inequality for the GLM with
non-i.i.d.~adapted samples under the compatibility condition in Lemma~\ref{lemma:oracle_ineq_GLM_L1}. 
For (ii), we aim to provide a remedy
without using the knowledge of sparsity or without using i.i.d.~samples. 
Hence, this poses a greater
challenge.  
In Section~\ref{sec:re_condition_matrix_concentration}, we address this
issue by showing that the empirical Gram matrix 
behaves ``nicely''  even when we choose arms adaptively
without deliberate random sampling. In particular, 
we show that adapted Gram matrices can be controlled by the theoretical Gram matrix,
and the empirical Gram matrix concentrates properly
around the adapted Gram matrix as we collect more samples. 
Connecting this matrix concentration to the corresponding compatibility constants, 
we show that the empirical Gram matrix satisfies the compatibility condition with high probability.

\subsubsection{Lasso Oracle Inequality for GLM with Non-i.i.d.~Data.}\label{sec:lasso_oracle_ineq_GLM}

We present an oracle inequality for the Lasso estimator for the GLM with
non-i.i.d.~data. This is a generalization
of the standard Lasso oracle inequality \citep{buhlmann2011statistics, van2008high} 
that allows for adapted sequences of observations. 
This is also a generalization of Proposition~1 in \citet{bastani2020online} 
to the GLM.
This convergence result may be of independent interest.

\begin{lemma}[Oracle inequality] \label{lemma:oracle_ineq_GLM_L1}
Let $\{X_\tau: \tau \in [t]\}$ be an adapted sequence such that each
$X_{\tau}$ may depend on $\{X_s: s < \tau\}$. 
Suppose the compatibility condition holds for the empirical covariance matrix
$\hat{\Sigma}_t = \frac{1}{t}\sum_{\tau=1}^t X_\tau X_\tau^\top$ with
active set $S_0$ and compatibility constant
~$\phi_t$. For $\delta \in
(0,1)$, define the regularization parameter 
\begin{equation*}
    \lambda_t := 2\sigma x_{\max} \sqrt{\frac{2[\log(2/\delta) + \log d]}{t}} \,.
\end{equation*}
Then with probability at least $1-\delta$, the Lasso estimate $\hat{\beta}_t$
defined in \eqref{eq:Lasso_glm} satisfies
\begin{align*}
        \| \hat{\beta}_t - \beta^* \|_1 \leq \frac{4 s_0 \lambda_t }{\kappa_0 \phi^2_t} \,.
\end{align*}
\end{lemma}
Note that here we assume that the compatibility condition holds for the
empirical Gram matrix $\hat{\Sigma}_t$. 
In the next section, we show that this holds with high probability.
The Lasso oracle inequality  holds without further  assumptions on the
underlying parameter $\beta^*$ or its support. Therefore, if we show that
$\hat{\Sigma}_t$ satisfies the compatibility condition without the knowledge
of  $s_0$, then the remainder of the result does not require 
this knowledge as well.

\subsubsection{Compatibility Condition and Matrix Concentration.}\label{sec:re_condition_matrix_concentration}

We first define the generic compatibility constant for matrix $M$ 
with respect to $S_0$.
\begin{definition}
The compatibility constant of $M$ over the set $A$ is given by
\begin{align*}
    \phi^2(M,A) := \min_\beta \left\{ \frac{a \beta^\top M \beta}{\|\beta_{A}\|^2_1} : \|\beta_{A^c}\|_1 \leq 3\|\beta_{A}\|_1 \neq 0 \right\},
\end{align*}
where $a$ denotes the cardinality of the set $A$.
\end{definition}
Hence, $M$ satisfies the compatibility condition if $\phi^2(M,A) > 0$. 
Although one can define a compatibility constant with respect to any index set,
in this section,
we will focus on the active index set $S_0$ of
the parameter $\beta^*$. 
 Also, note that the constant 3 in the inequality is for ease of
 exposition and may be replaced by a different value, but then one has to
 adjust the choice of the regularization parameter accordingly. Now, under
 Assumption~\ref{assum:comp_condition}, the theoretical Gram matrix
 $\Sigma = \frac{1}{K}\mathbb{E}[\mathbf{X}^\top \mathbf{X}]$ satisfies
 the compatibility condition i.e., $\phi_0^2 = \phi^2(\Sigma, S_0) > 0$. 

\begin{definition}
We define the \textit{adapted} Gram matrix as $\Sigma_t := \frac{1}{t}\sum_{\tau=1}^t \mathbb{E}[X_\tau X_\tau^\top | \mathcal{F}_{\tau-1}]$ and the \textit{empirical} Gram matrix as $\hat{\Sigma}_t := \sum_{\tau=1}^t X_\tau X_\tau^\top$. 
\end{definition}

For each term $\mathbb{E}[X_\tau X_\tau^\top | \mathcal{F}_{\tau-1}]$ in $\Sigma_t$, the past observations $\mathcal{F}_{\tau-1}$ affects how the feature vector $X_\tau$ is chosen. More specifically, our algorithm uses $\mathcal{F}_{\tau-1}$ to compute $\hat{\beta}_\tau$ and then chooses arm $a_\tau$ such that its feature $x_{a_\tau}$  maximizes $x_{a_\tau}^\top \hat{\beta}_\tau$. Therefore, we can rewrite $\Sigma_t$ as
\begin{align*}\label{eq:adapted_cov_matrix}
    \Sigma_t
    = \frac{1}{t}\sum_{\tau=1}^t\sum_{i = 1}^2 \mathbb{E}_{\mathcal{X}_\tau}\big[X_{\tau, i} X_{\tau,i}^\top  \mathbbm{1}\{X_{\tau, i} = \argmax_{X \in \mathcal{X}_\tau} X^\top \hat{\beta}_{\tau}\} \mid \hat{\beta}_{\tau} \big]\,. 
\end{align*}
Assumption~\ref{assum:comp_condition} guarantees that the compatibility condition is  satisfied by the
theoretical Gram matrix $\Sigma$; however, we need to show the empirical Gram
matrix $\hat{\Sigma}_t$ satisfies the compatibility condition. In our analysis, we use the adapted
Gram matrix $\Sigma_t$ as a bridge between $\Sigma$ and 
$\hat{\Sigma}_t$. We first lower-bound the
compatibility constant
$\phi^2(\Sigma_t,S_0)$ in terms of 
$\phi^2(\Sigma,S_0)$ so that we can show that $\Sigma_t$ satisfies the
compatibility condition as long as $\Sigma$ satisfies the
compatibility condition. Then, we show that $\hat{\Sigma}_t$ concentrates
around $\Sigma_t$ with high probability and that such matrix concentration
guarantees the compatibility condition of $\hat{\Sigma}_t$.

In Lemma~\ref{lemma:cov_matrix_lowerbound_asymmetry}, we show that
the adapted Gram matrix $\Sigma_t$ can be controlled
in terms of the theoretical Gram matrix
$\Sigma$, which allows us
to link the compatibility constant of $\Sigma$ to compatibility constant of $\Sigma_t$.
Note that Lemma
\ref{lemma:cov_matrix_lowerbound_asymmetry} shows the result for any fixed
vector $\beta$; hence, it  can be applied to $\mathbb{E}[X_\tau X_\tau^\top |
\mathcal{F}_{\tau-1}] $. 

\begin{lemma}\label{lemma:cov_matrix_lowerbound_asymmetry}
For a fixed vector $\beta \in \mathbb{R}^d$, we have
\begin{equation*}
     \sum_{i=1}^2 \mathbb{E}_{\mathcal{X}_t}\Big[X_{t,i} X_{t,i}^\top \mathbbm{1}\{X_{t,i} =
     \argmax_{\mathclap{X \in \mathcal{X}_t}} X^\top \beta\}\Big]
     \succcurlyeq \nu^{-1}\Sigma,
   \end{equation*}
   where $\nu$ the degree of asymmetry of the distribution
   $p_{\mathcal{X}}$ defined in Assumption~\ref{assum:rho}.
\end{lemma}
Therefore, we have $\Sigma_t \succcurlyeq \nu^{-1}\Sigma$
  which implies that
  $\phi^2(\Sigma_t, S_0) \geq \frac{\phi^2(\Sigma, S_0)}{\nu} > 0$, i.e., $\Sigma_t$ satisfies the compatibility
  condition. Note that both $\Sigma$ and $\Sigma_t$ can be singular.
In Lemma~\ref{lemma:Sigma_hat_concentration},
we show that $\hat{\Sigma}_t$ concentrates to
$\Sigma_t$ with high probability. This result is crucial in our analysis since it allows the matrix concentration without using i.i.d.~samples. The proof of Lemma~\ref{lemma:Sigma_hat_concentration} utilizes a new Bernstein-type inequality
for adapted samples (Lemma~\ref{lemma:maximal_bernstein_ineq} in the
appendix) which may be of independent interest. 
\begin{lemma}[Matrix concentration]\label{lemma:Sigma_hat_concentration}
For $t \geq \frac{2\log(2d^2)}{C_0(s_0)^2}$ where $C_0(s_0) = \min\!\Big(\frac{1}{2}, \frac{\phi_0^2}{256 s_0 \nu x_{\max}^2 } \Big)$, we have
\begin{align*}
    \mathbb{P}\left( \| \Sigma_t - \hat{\Sigma}_t \|_{\infty} \geq  \frac{ \phi^2_0}{32  s_0\nu } \right) 
    &\leq \exp\!\left(-\frac{t C_0(s_0)^2}{2}\right) \,.
\end{align*}
\end{lemma}
Then, we invoke the following corollary to use the matrix concentration results to ensure the compatibility condition for $\hat{\Sigma}_t$.
\begin{corollary}[Corollary 6.8,
  \citet{buhlmann2011statistics}]\label{cor:gram_mat_infinity_norm_compatibility} 
Suppose that $\Sigma_0$-compatibility condition holds for the index set
$S$ with cardinality $s = |S|$, with compatibility constant
$\phi^2(\Sigma_0, S)$, and that $\| \Sigma_1 - \Sigma_0 \|_\infty \leq
\Delta$, where $32s \Delta \leq \phi^2(\Sigma_0,S)$. 
Then, for the set $S$, the $\Sigma_1$-compatibility condition holds as
well, with $\phi^2(\Sigma_1, S) \geq \phi^2(\Sigma_0, S)/2$. 
\end{corollary} 
In order to satisfy the hypotheses in
Lemma~\ref{lemma:Sigma_hat_concentration} and
Corollary~\ref{cor:gram_mat_infinity_norm_compatibility}, we 
define the \textit{initial} period $t < T_0  :=
\frac{2\log(2d^2)}{C_0(s_0)^2}$ during which the compatibility
condition for the empirical Gram matrix is not guaranteed, 
and the event
\begin{align*}
    \mathcal{E}_t := \left\{ \| \Sigma_t - \hat{\Sigma}_t \|_{\infty} \leq  \frac{ \phi^2_0}{32  s_0\nu } \right\} \,.
\end{align*}
Then for all $t \geq \lceil T_0 \rceil$ and 
$\Sigma_t$ for which event $\mathcal{E}_t$ holds, we have
\begin{align*}
\phi^2_t := \phi^2(\hat{\Sigma}_t, S_0) \geq \frac{\phi^2(\Sigma_t, S_0)}{2} \geq \frac{\phi^2_0}{ 2 \nu } > 0\,.
\end{align*}
Hence, the compatibility condition is satisfied for the empirical Gram matrix without using sparsity information. 

\subsubsection{Proof Sketch of Theorem~\ref{thm:GLM-Lasso-regret-bound_CC}}

We combine the results above to analyze the regret bound of \textsc{SA~Lasso~Bandit} shown in Theorem~\ref{thm:GLM-Lasso-regret-bound_CC}. First, we divide the time horizon $[T]$ into three groups:
\begin{enumerate}[label=(\alph*)]
\item $(t \leq T_0)$. Here the compatibility condition is not guaranteed to hold.
\item $(t > T_0)$ such that  $ \mathcal{E}_t$ holds.
\item $(t > T_0)$ 
  such that
  $ \mathcal{E}_t$ does not hold.
\end{enumerate}
 These sets are disjoint, hence we bound the regret contribution from
 each  separately and obtain an upper bound on the  overall regret. It is
 important to note that \textsc{SA~Lasso~Bandit} Algorithm does not rely
 in any way on this partitioning --  it is introduced purely for
 the purpose of analysis. Set  (a) is the  initial period over which
 we do not have guarantees for the compatibility condition. Therefore, we
 cannot apply the Lasso convergence result; hence we can incur
 $\mathcal{O}(s_0^2 \log d)$ regret. Set (b) is where the compatibility
 condition is satisfied; hence the Lasso oracle inequality in
 Lemma~\ref{lemma:oracle_ineq_GLM_L1} can apply. In fact, this group can
 be further divided to two cases: (b-1) when the high-probability Lasso
 result holds and (b-2) when it does not, where the regret of (b-2) can be
 bounded by $\mathcal{O}(1)$. For (b-1), using the Lasso convergence
 result and summing the regret over the time horizon gives $\mathcal{O}(
 s_0\sqrt{T\log (dT)} )$ regret, which is the leading factor in the regret
 bound of Theorem~\ref{thm:GLM-Lasso-regret-bound_CC}.  
 Lastly, (c) contains the failure events of
 Lemma~\ref{lemma:Sigma_hat_concentration} whose regret is
 $\mathcal{O}(s_0^2)$. 
 The proofs of the lemmas are
 in Appendix~\ref{app:Theorem1}, followed by the complete proof of
 Theorem~\ref{thm:GLM-Lasso-regret-bound_CC} in
 Appendix~\ref{app:Theorem1main}.

\subsection{Regret under the Restricted Eigenvalue Condition}\label{sec:regret_re}
In our analysis so far, we have presented the main results under the
compatibility condition in order to be consistent with previous results in
the sparse bandit literature. 
In this section, we present the regret bound for \textsc{SA~Lasso~Bandit}
under the restricted eigenvalue (RE) condition and briefly discuss its
implication in terms of potentially matching lower bounds.  
Similar to the analysis under the compatibility condition, we assume that 
the RE condition is satisfied only for the theoretical Gram matrix $\Sigma
= \frac{1}{K}\mathbb{E}[\mathbf{X}^\top \mathbf{X}]$. 
\begin{assumption}[RE condition]\label{assum:re_condition}
For active set $S_0$ and $\Sigma$, there exists restricted eigenvalue $\phi_1 > 0$ such that
$\phi_1^2 \| \beta \|^2_2 \leq  \beta^\top  \Sigma \beta$ for all $\beta 
\in \mathbb{C}(S_0)$ defined in  \eqref{eq:set_C}.
\end{assumption}
The RE condition is very similar to the compatibility condition in
Assumption~\ref{assum:comp_condition} but uses the $\ell_2$ norm instead of
the $\ell_1$ norm. Based on
this condition, we can show the following regret bound. 

\begin{theorem}[Regret bound under RE condition]\label{thm:GLM-Lasso-regret-bound_re}
Suppose $K = 2$ and Assumptions~\ref{assum:feature_param_bounds}, \ref{assum:link_func_bounds}, \ref{assum:rho}, and \ref{assum:re_condition} hold. 
Then the expected cumulative regret of the \textsc{SA~Lasso~Bandit} policy is  $\mathcal{O}\big( \sqrt{ s_0 T \log (dT)} \big)$.
\end{theorem}

Theorem~\ref{thm:GLM-Lasso-regret-bound_re} establishes $\mathcal{O}\big(
\sqrt{ s_0 T \log (dT)} \big)$ regret without any prior knowledge on
$s_0$. The regret upper-bound based on  the RE condition still enjoys
logarithmic dependence on $d$ and furthermore sub-linear dependence on
$s_0$. 
Compared to Theorem~\ref{thm:GLM-Lasso-regret-bound_CC}, the regret bound
in Theorem~\ref{thm:GLM-Lasso-regret-bound_re} is smaller by $\sqrt{s_0}$
factor, which is again consistent with the offline Lasso results under the
RE condition (Theorem ~7.19 in \citealt{wainwright2019high}). The
difference in the regret bounds in
Theorem~\ref{thm:GLM-Lasso-regret-bound_CC} and
Theorem~\ref{thm:GLM-Lasso-regret-bound_re} is due to the RE condition
being slightly stronger than the compatibility condition. 

The RE condition is more directly analogous (as compared to the compatibility
condition) to the standard positive-definiteness assumption for covariance
matrices in GLM bandits \citep{li2017provably}. That is, the RE condition
is equivalent to positive-definite covariance when $s_0 = d$ , i.e.,
non-sparse settings. 
\citet{li2017provably} showed $\mathcal{O}\big((\log
T)^{3/2}\sqrt{dT\log K} \big)$  regret bound of  for GLM bandits, which
matches the $\Omega(\sqrt{dT})$ minimax lower bound established
\citep{chu2011contextual} for linear bandits with finite arms, up to
logarithmic factors.
Therefore, in sparse settings, we conjecture that $\mathcal{O}\big( \sqrt{
  s_0 T \log (dT)} \big)$ regret is \textit{best possible} up to
logarithmic factors under the RE condition  (and so is $\mathcal{O}\big(
s_0\sqrt{  T \log (dT)} \big)$ regret under the compatibility
condition). While we present these conjectures, we do not claim our
results are minimax. In fact, we discuss in Section~\ref{sec:conclusion} that the entire notion of
minimax regret is much more delicate in sparse bandits.

\section{Numerical Experiments}\label{sec:experiments}

\begin{figure*}[t]
    \begin{subfigure}[b]{0.32\textwidth}
        \includegraphics[width=\textwidth]{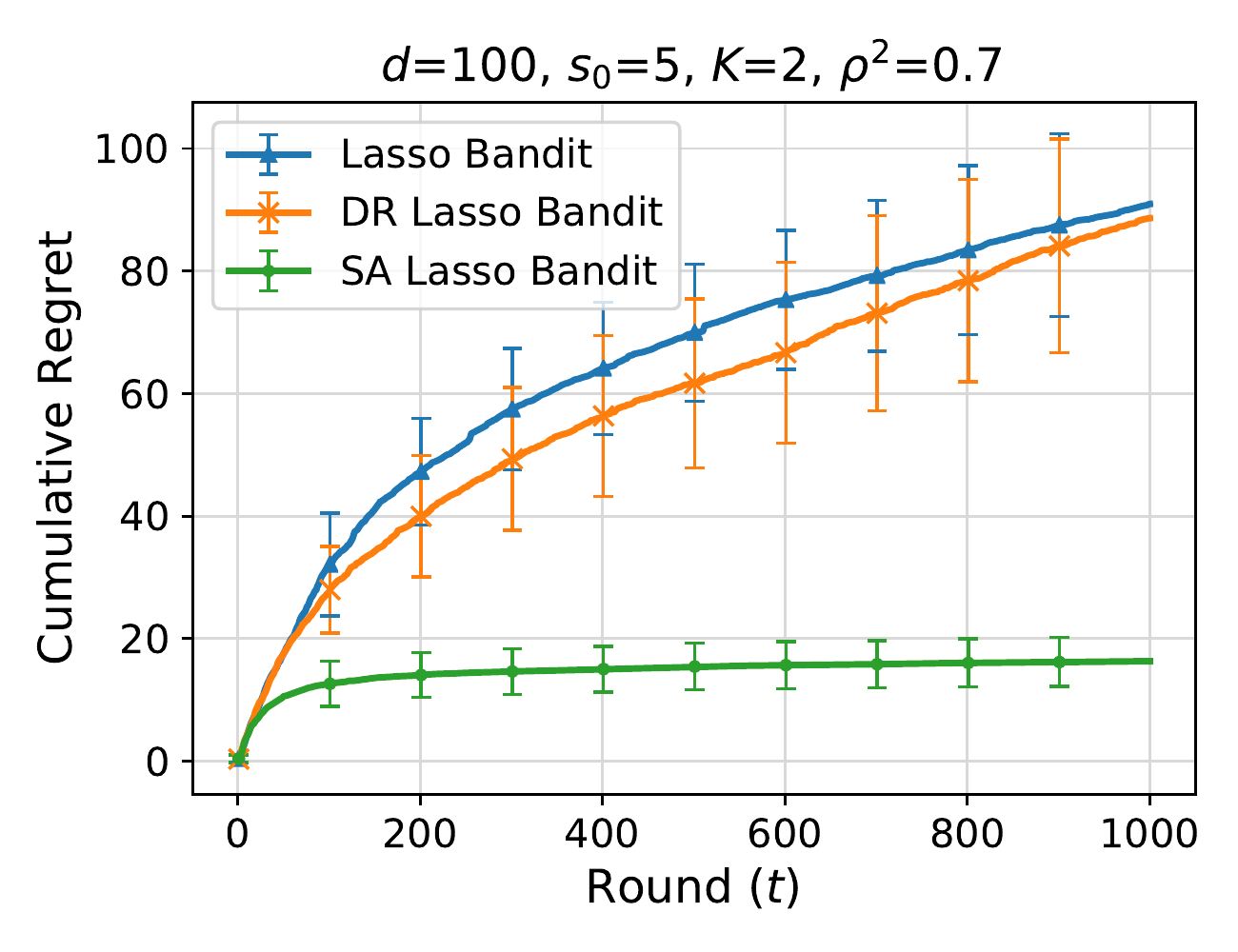}
    \end{subfigure}
    \begin{subfigure}[b]{0.32\textwidth}
        \includegraphics[width=\textwidth]{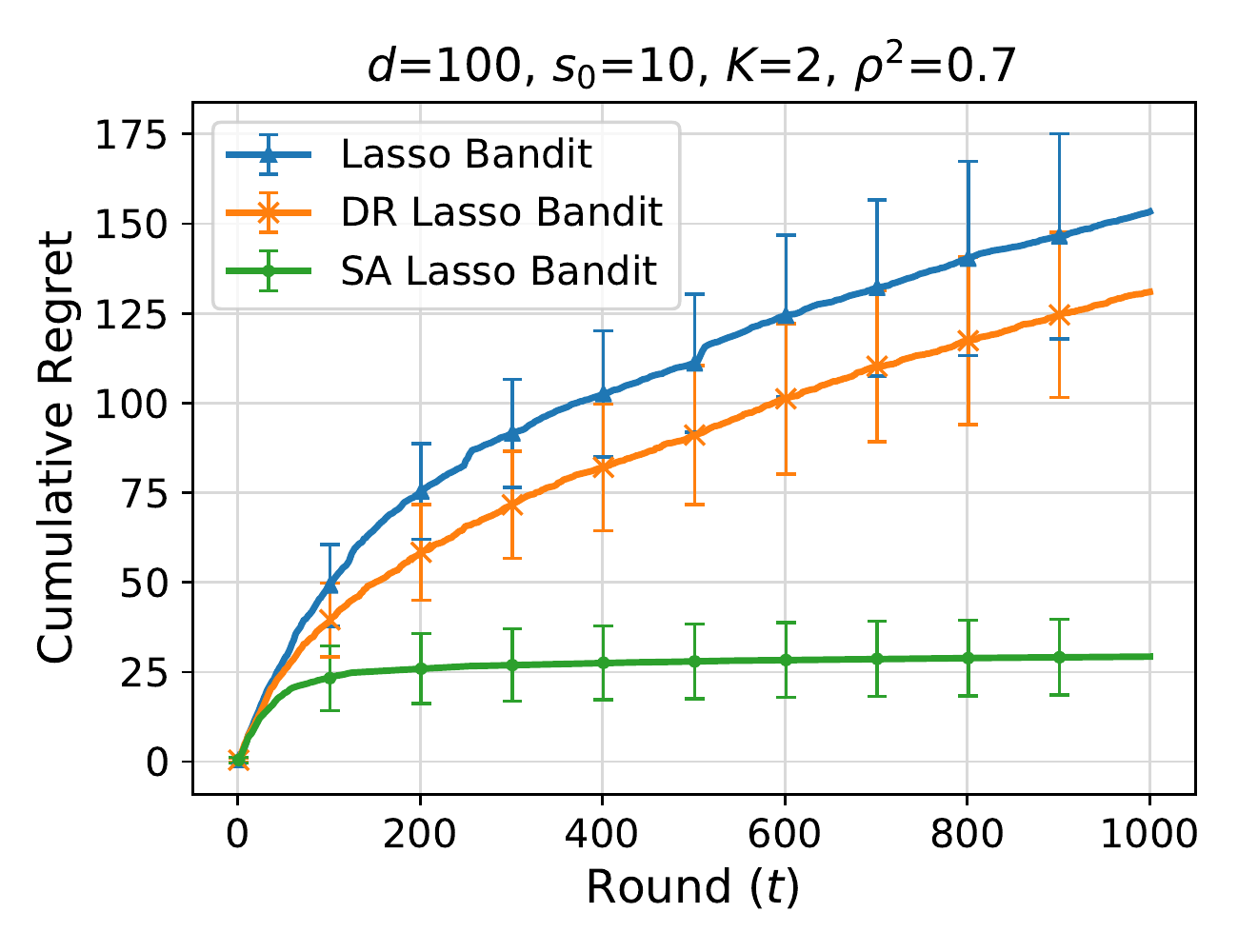}
    \end{subfigure}
    \begin{subfigure}[b]{0.32\textwidth}
        \includegraphics[width=\textwidth]{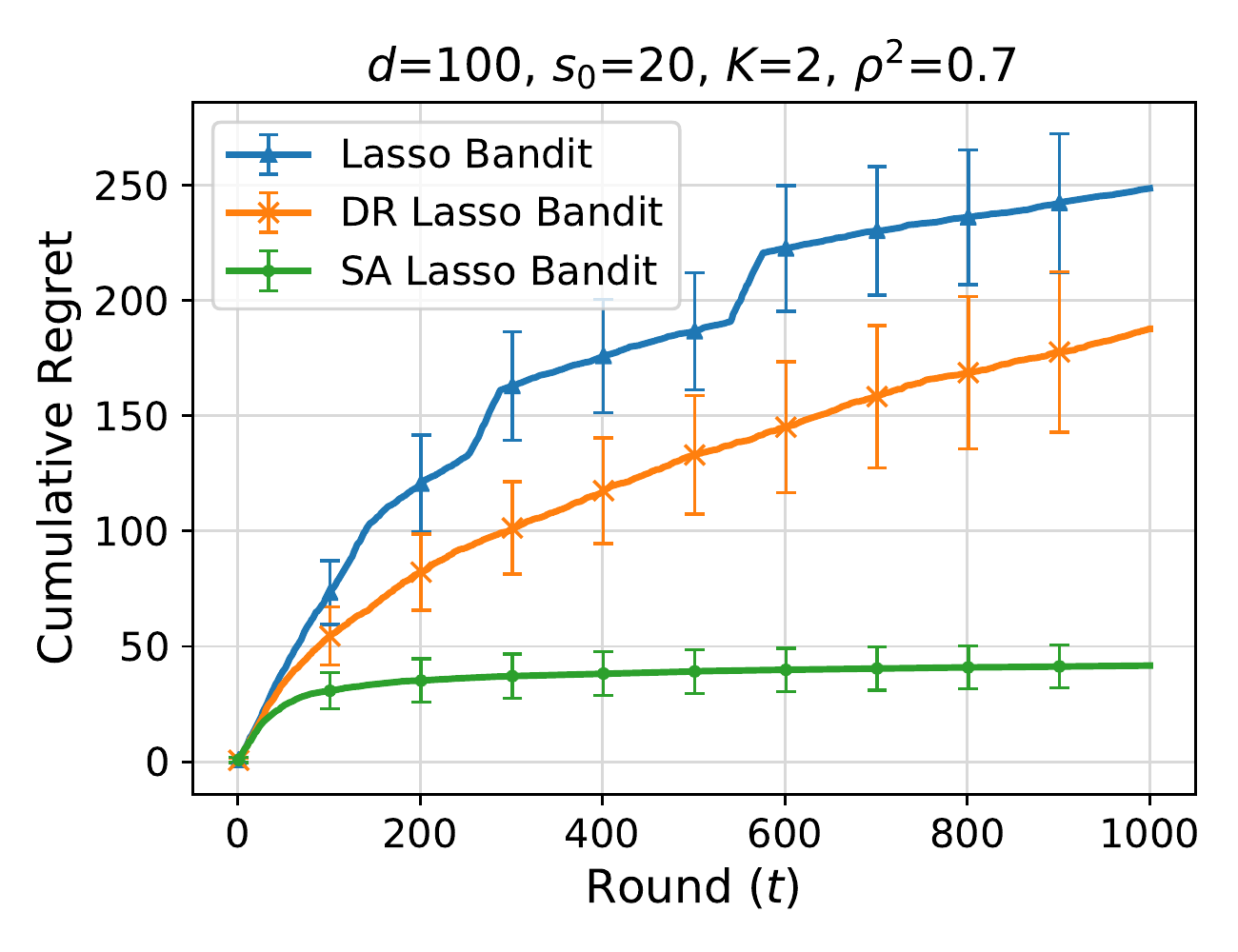}
    \end{subfigure}\\
    \begin{subfigure}[b]{0.32\textwidth}
        \includegraphics[width=\textwidth]{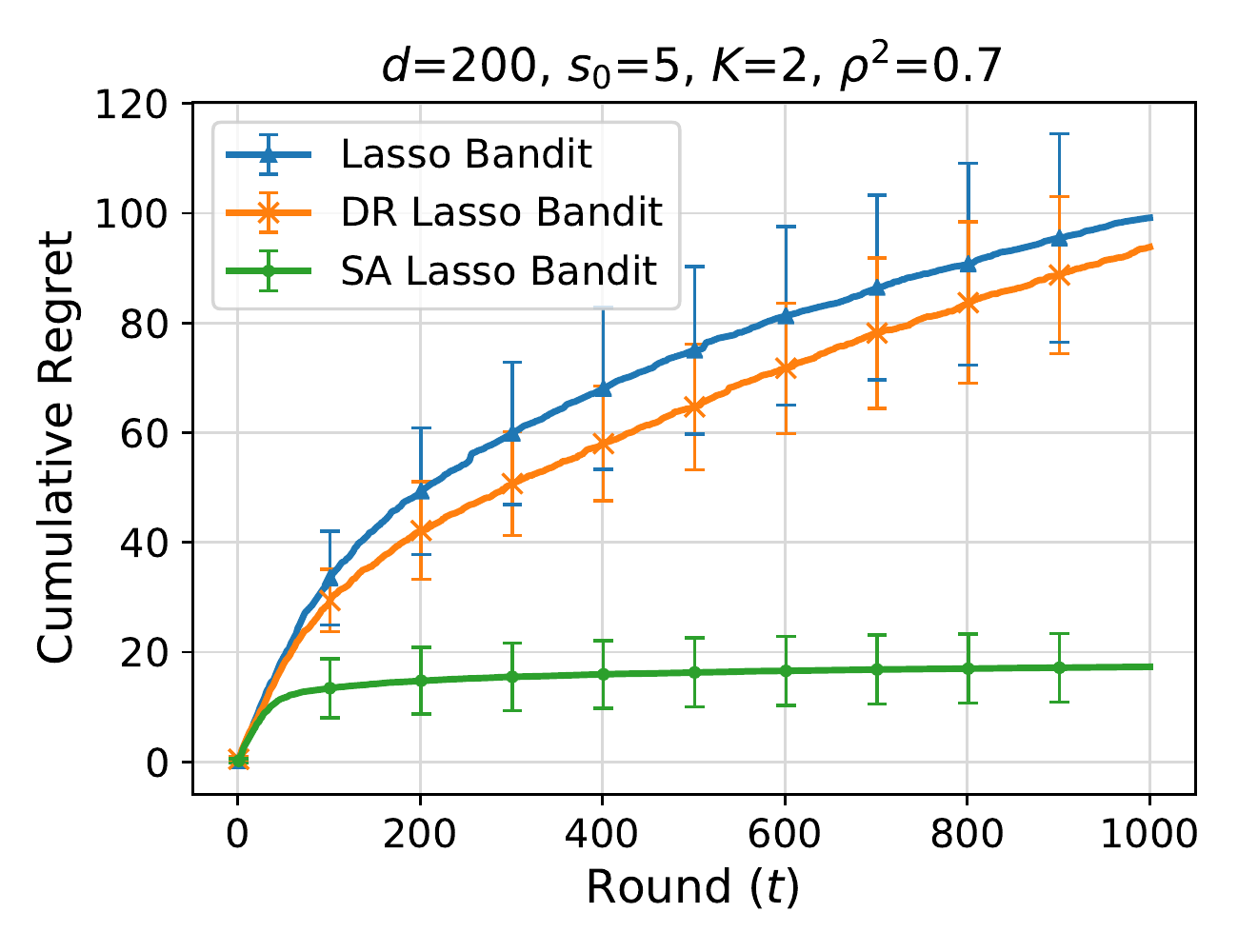}
    \end{subfigure}
    \begin{subfigure}[b]{0.32\textwidth}
        \includegraphics[width=\textwidth]{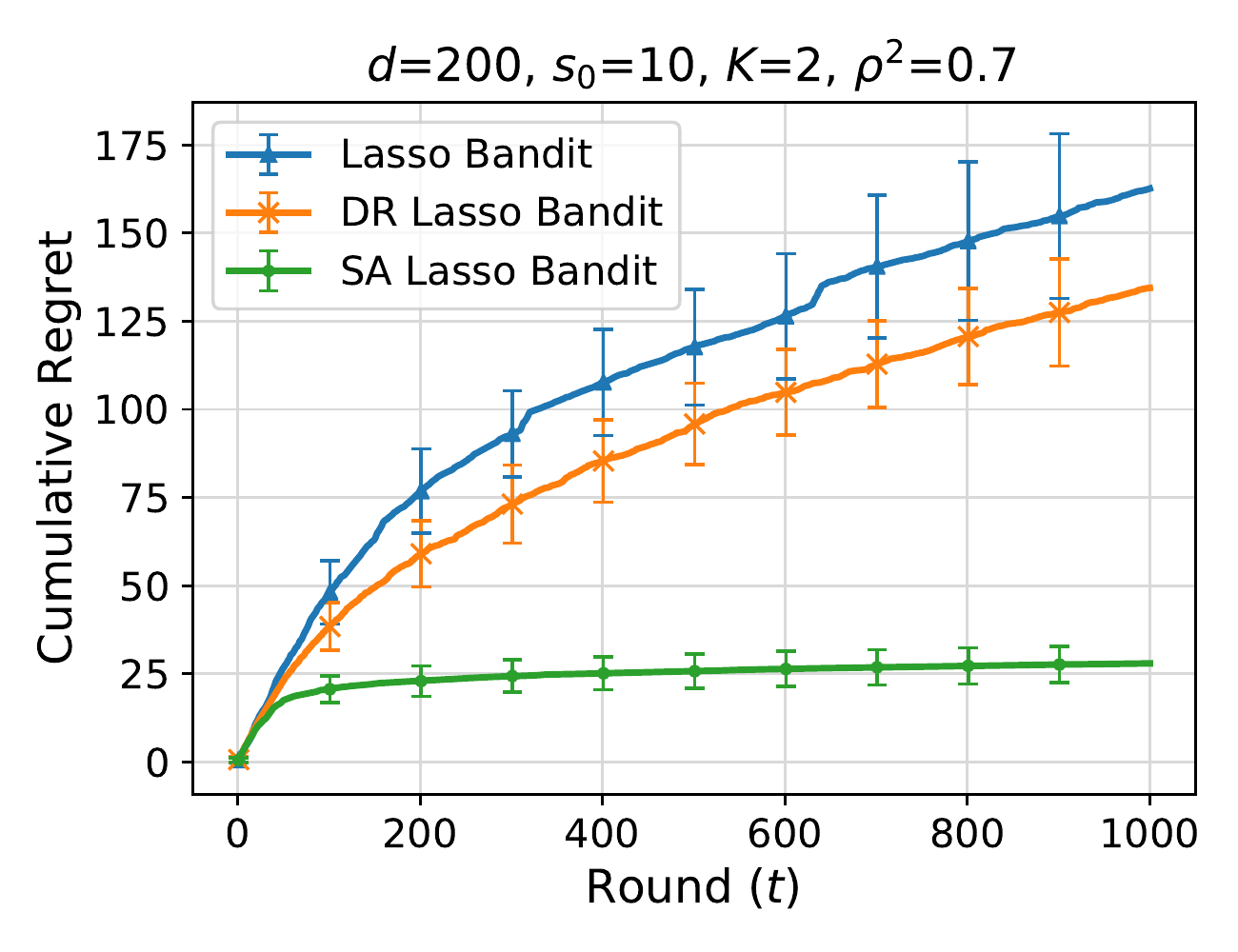}
    \end{subfigure}
    \begin{subfigure}[b]{0.32\textwidth}
        \includegraphics[width=\textwidth]{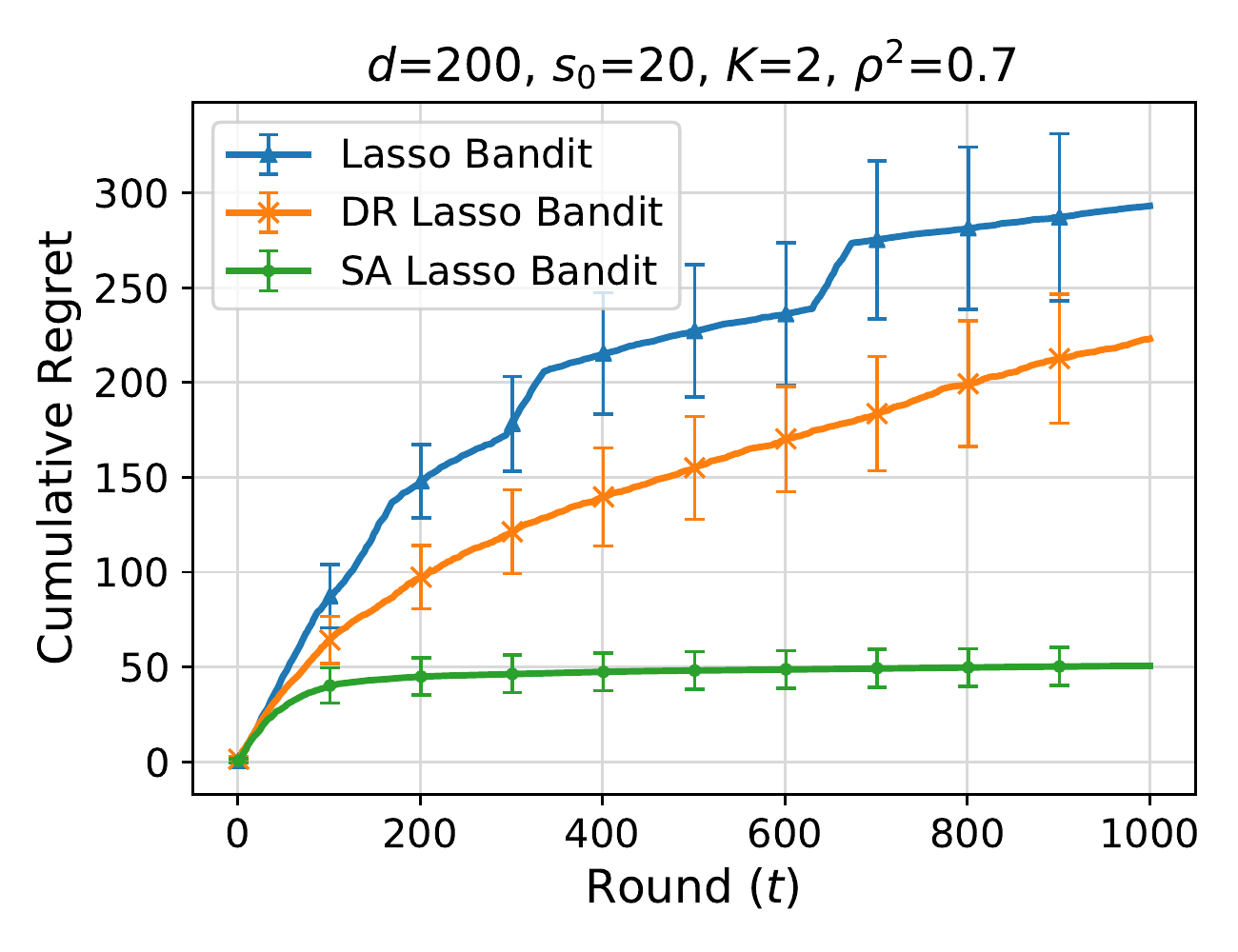}
    \end{subfigure}
    \caption{\small The plots show the $t$-round cumulative regret of \textsc{SA Lasso Bandit} (Algorithm~\ref{algo:SA_Lasso_bandit}), \textsc{DR Lasso Bandit} \citep{kim2019doubly}, and \textsc{Lasso Bandit} \citep{bastani2020online} for $K = 2$, $d = 100$ (first row) and $d = 200$ (second row) with varying sparsity $s_0 \in \{5, 10, 20\}$ under  strong correlation, $\rho^2 = 0.7$.} 
     \label{fig:two_arm_strong_corr}
\end{figure*}

We conduct numerical experiments to evaluate \textsc{SA Lasso Bandit} and compare with existing sparse bandit algorithms: \textsc{DR Lasso Bandit} \citep{kim2019doubly} and \textsc{Lasso Bandit}  \citep{bastani2020online} in two-armed contextual bandits. 
We follow the experimental setup of \citet{kim2019doubly} to evaluate algorithms under different levels of correlation between arms. Although we consider $K = 2$ case in this section, the experimental setup introduced here also applies to numerical evaluations for $K \geq 3$ armed case in Section~\ref{sec:K-arms}.
For each dimension $i \in  [d]$, we sample each element of the feature vectors $[X_{t,1}^{(i)}, ..., X_{t,K}^{(i)}]$ from multivariate Gaussian distribution $\mathcal{N}(\vec{0}_K, V)$ where covariance matrix $V$ is defined as $V_{i,i} = 1$ for all diagonal elements $i \in [K]$ and $V_{i,j} = \rho^2$ for all off-diagonal elements $i \neq j \in [K]$. Hence, for $\rho^2 > 0$, feature vectors for each arm are allowed to be correlated.
We consider different levels of correlation with $\rho^2 = 0.7$ (strong correlation) in Figure~\ref{fig:two_arm_strong_corr} and $\rho^2 = 0.3$ (weak correlation) in Figure~\ref{fig:two_arm_weak_corr} as well as $\rho^2 = 0$ (no correlation) in the appendix. 
In these sets of experiments, we consider feature dimensions $d = 100$ and $d = 200$. For comparison, we use a linear reward with the linear link function $\mu(z) = z$ since both \textsc{Lasso Bandit} and \textsc{DR Lasso Bandit} are proposed in linear reward settings. 
We generate $\beta^*$ with varying sparsity $s_0 = \| \beta^* \|_0$. For a given $s_0$, we generate each non-zero element of $\beta^*$ from a uniform distribution in $[0,1]$.
For noise, we sample $\epsilon_{t} \sim \mathcal{N}(0,1)$ independently for all rounds.
For each case with different experimental configurations, we conduct 20
independent runs, and
report the average of the cumulative regret for each of the algorithms.
The error bars represent the standard deviations. 

\begin{figure*}[t]
    \begin{subfigure}[b]{0.32\textwidth}
        \includegraphics[width=\textwidth]{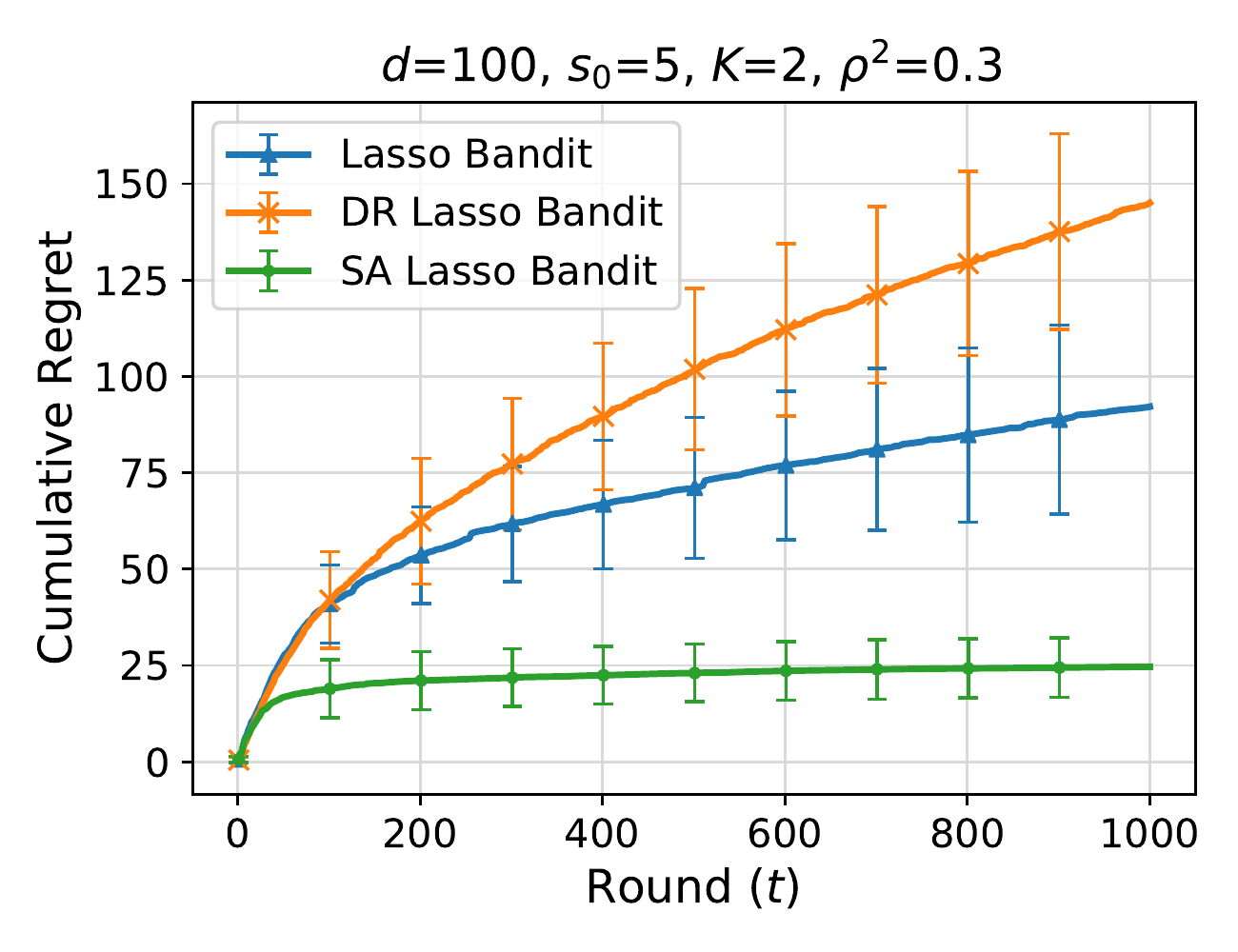}
    \end{subfigure}
    \begin{subfigure}[b]{0.32\textwidth}
        \includegraphics[width=\textwidth]{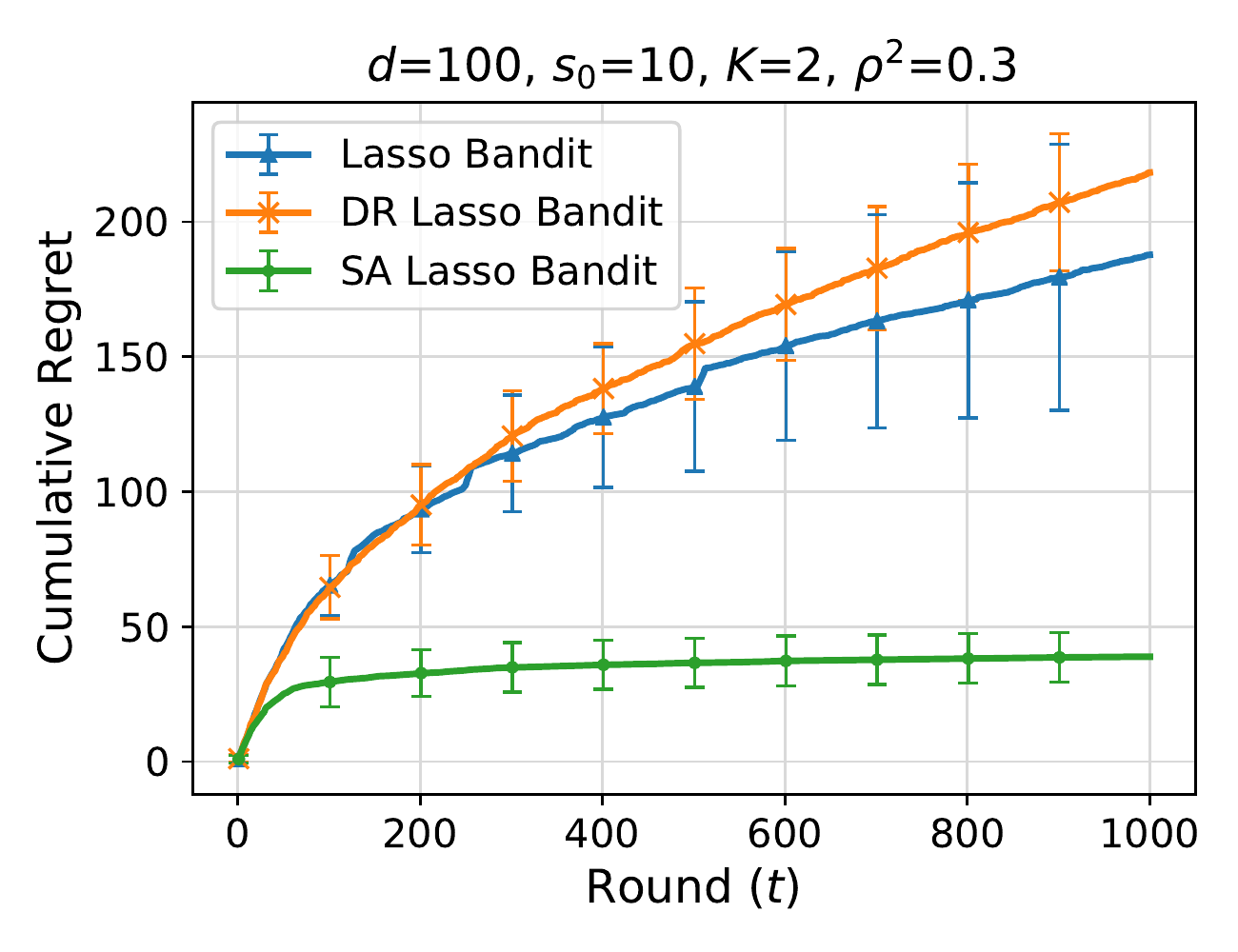}
    \end{subfigure}
    \begin{subfigure}[b]{0.32\textwidth}
        \includegraphics[width=\textwidth]{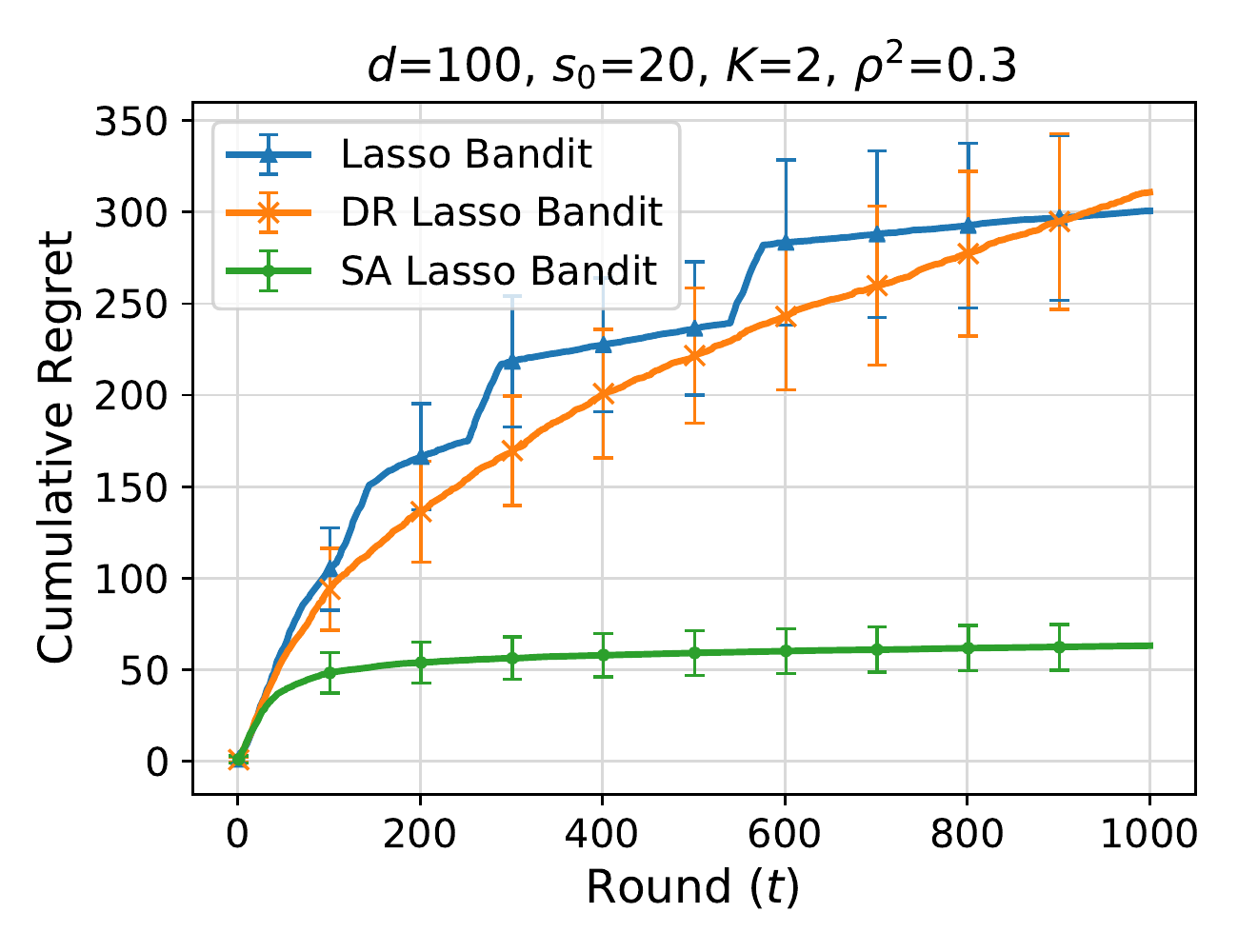}
    \end{subfigure}\\
    \begin{subfigure}[b]{0.32\textwidth}
        \includegraphics[width=\textwidth]{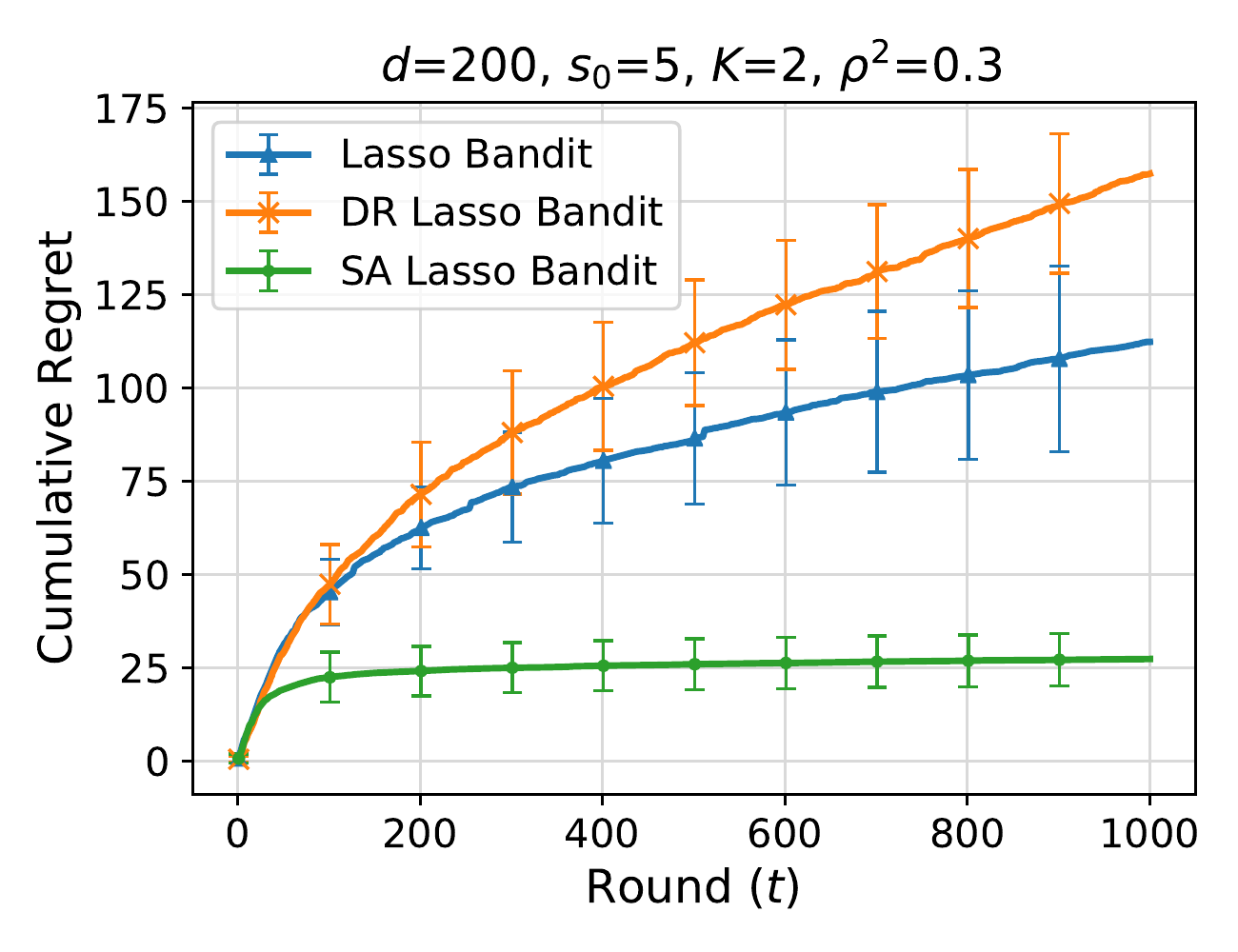}
    \end{subfigure}
    \begin{subfigure}[b]{0.32\textwidth}
      \includegraphics[width=\textwidth]{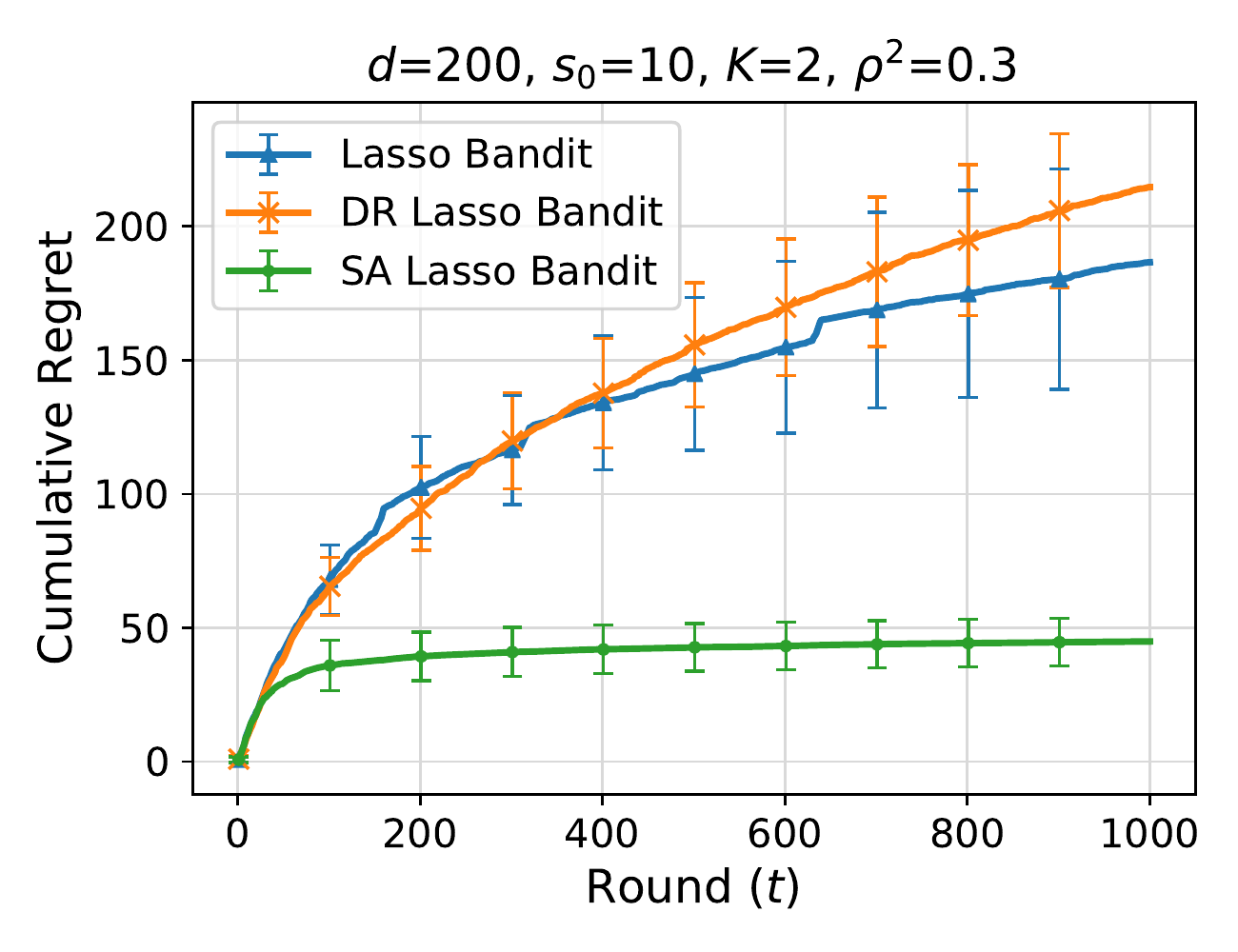}
    \end{subfigure}
    \begin{subfigure}[b]{0.32\textwidth}
      \includegraphics[width=\textwidth]{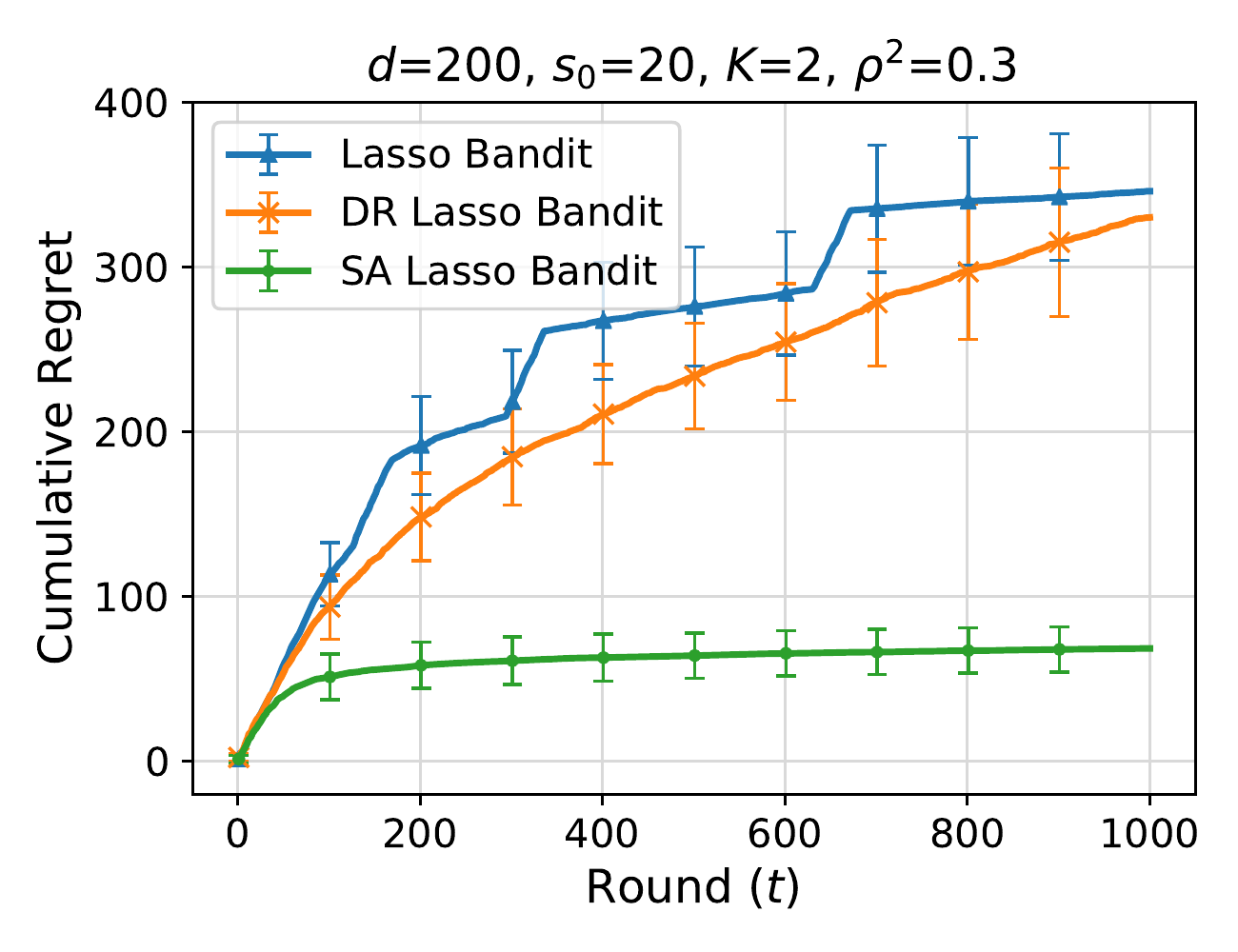}
    \end{subfigure}
    \caption{\small The plots show the $t$-round cumulative regret of \textsc{SA Lasso Bandit} (Algorithm~\ref{algo:SA_Lasso_bandit}), \textsc{DR Lasso Bandit} \citep{kim2019doubly}, and \textsc{Lasso Bandit} \citep{bastani2020online} for $K = 2$, $d = 100$ (first row) and $d = 200$ (second row) with varying sparsity $s_0 \in \{5, 10, 20\}$ under weak correlation, $\rho^2 = 0.3$.} 
     \label{fig:two_arm_weak_corr}
\end{figure*}

\textsc{DR Lasso Bandit} is proposed for the same problem setting as
ours. Therefore, it does not require any modifications for
experiments. However, the problem setting of \textsc{Lasso Bandit} is
different from ours: it assumes that the context variable is the same for
all arms but 
each arm has a different parameter.
We follow the setup in 
\citet{kim2019doubly},  and adapt \textsc{Lasso
Bandit to our setting by defining} a $Kd$-dimensional context vector
$X_t = [X_{t,1}^\top, ...,
X_{t,K}^\top]^\top  \in \mathbb{R}^{Kd}$ and  a $K d$-dimensional parameter
$\beta^*_i$ for each arm $i$ where $\beta^*_i =
[{\beta^*}^\top\mathbbm{1}(i=1), ..., {\beta^*}^\top\mathbbm{1}(i=K)]^\top
\in \mathbb{R}^{Kd}$; thus, $X_t^\top \beta^*_i = X_{ti}^\top \beta^*$s. Note
that despite the concatenation, the effective 
dimension of the unknown parameter $\beta^*_i$ remains the same as far as
estimation is concerned. We defer the other details of the experimental
setup and additional results to the appendix.

It is important to note that we report the performances of the benchmarks
(\textsc{DR Lasso Bandit} and \textsc{Lasso Bandit}) assuming that they
have access to correct sparsity index~$s_0$; however, this information is hidden
from our algorithm. Despite this advantage, the experiment
results shown in Figure~\ref{fig:two_arm_strong_corr} and
Figure~\ref{fig:two_arm_weak_corr} demonstrate that 
\textsc{SA Lasso Bandit} outperforms the other methods by significant
margin consistently across various problem instances.  
We also verify that the performance of our proposed algorithm is the least
sensitive to the details of the problem instances, and scales well with
changes in the instance. 
The regret of our algorithm appears to  
scale linearly with the 
sparsity index $s_0$, 
while its dependence on the feature dimension $d$ appears to be very minimal in
most of the instances, which is consistent with our theoretical findings. 
We also observe that a higher correlation between arms (feature vectors)
improves the overall performances of the algorithms. This finding is
stronger in 
the experiments for the $K$-armed case. We discuss
this phenomenon in detail in Section~\ref{sec:K-arms}. 

\section{Extension to \textit{K} Arms}\label{sec:K-arms}
Thus far, we have presented our main results in two-armed bandit settings
which highlight the main challenges of sparse bandit problems without
prior knowledge of sparsity. 
In this section, we extend our regret analysis to the case of $K \geq 3$
arms. Also, we present additional numerical experiments for $K$-armed
bandits.  

\subsection{Regret Analysis for \textit{K} Arms}
Recall that \textsc{SA~Lasso~Bandit} is valid for any number of arms;
hence, no modifications are required to extend the algorithm to 
$K \geq 3$ arms.
The analysis of \textsc{SA Lasso Bandit} for the $K$-armed case tackles
largely the same challenges described in
Section~\ref{sec:challenges_proof_outlines}: the need for a Lasso
convergence result for adapted samples and ensuring the compatibility
condition without knowing $s_0$ (and without relying on
i.i.d.~samples). The former challenge is again taken care of by the Lasso
convergence result in Lemma~\ref{lemma:oracle_ineq_GLM_L1}. However, the
latter issue is more subtle in the $K$-armed case than in the two-armed
case. In particular, when controlling the adapted Gram matrix $\Sigma_t$
with the theoretical Gram matrix $\Sigma$, the Gram matrix for the
unobserved feature vectors could be incomparable with the Gram matrix for the
observed feature vectors. 
For this issue,  we introduce an additional regularity condition, which we denote as the ``balanced covariance'' condition.

\begin{assumption}
[Balanced covariance]\label{assum:balanced_cov}
Consider a permutation $(i_1, ..., i_K)$ of $(1, ..., K)$. For any integer $k \in \{2,...,K-1\}$ and fixed vector  $\beta$, there exists $C_\mathcal{X} < \infty$ such that
\begin{align*}
    \mathbb{E}\left[ X_{i_k} X^\top_{i_k}  \mathbbm{1}\{ X_{i_1}^\top \beta < ... < X_{i_K}^\top \beta \} \right]
    &\preccurlyeq C_\mathcal{X}  \mathbb{E}\left[ ( X_{i_1} X^\top_{i_1} + X_{i_K} X^\top_{i_K}) \mathbbm{1}\{ X_{i_1}^\top \beta < ... < X_{i_K}^\top \beta \} \right] .
\end{align*}
\end{assumption}
In Algorithm~\ref{algo:SA_Lasso_bandit} we observe only the reward corresponding to arm $i_1$, and Assumption~\ref{assum:rho} implies that we have some control on the arm $i_K$.
This balanced covariance condition implies that there is
``sufficient randomness'' in the observed features compared to non-observed features. 
The exact value of $C_\mathcal{X}$ depends on 
the joint distribution of $\mathcal{X}$ including the correlation between
arms. In general, the more positive the correlation, the smaller
$C_\mathcal{X}$  (obviously, with an extreme case of perfectly correlated
arms having a constant $C_\mathcal{X}$ independent of any problem
parameters). 
When the arms are
independent and identically distributed, Assumption~\ref{assum:balanced_cov} holds with $C_\mathcal{X}=\mathcal{O}(1)$
for  
both the multivariate Gaussian distribution and a uniform
distribution on a sphere, and 
for an arbitrary independent distribution for each arm,
Assumption~\ref{assum:balanced_cov}  holds for $C_{\mathcal{X}}=
\binom{K-1}{K_0}$ where $K_0 = \lceil (K-1)/2 \rceil$.
It is important to
note that even in this pessimistic case, $C_\mathcal{X}$ does not exhibit
dependence on dimensionality $d$ or the sparsity index $s_0$. These are formalized
in Proposition~\ref{prop:c_x_bound} in Appendix~\ref{sec:K-arm-regret-analysis}.\footnote{While it is
  not our primary goal to derive general  tight bounds on $C_\mathcal{X}$,
  we acknowledge that the bound on $C_\mathcal{X}$ for an 
  arbitrary
  distribution for independent arms is very loose, and is the result of conservative analysis driven
  by lack of information on $p_\mathcal{X}$. 
  Numerical evaluation on distributions other than Gaussian and uniform
  distributions, detailed  in Section~\ref{sec:K-arms}, buttress this point and
  indicate that the dependence on $K$ is no greater than linear.} 
This balanced covariance condition is somewhat similar to
``positive-definiteness'' condition for observed contexts in the 
bandit literature (e.g.,
\citet{goldenshluger2013linear,bastani2020mostly}). However, notice that
we allow the covariance matrices on both sides of the inequality to be
singular. Hence, the positive-definiteness condition for observed context
in our setting may not hold even when the balanced covariance condition
holds.   
While this condition admittedly originates from our proof technique, it
also provides potential insights on learnability of problem
instances. That is, $C_\mathcal{X}$ close to infinity implies that the
distribution of feature vectors is heavily skewed toward a particular
direction. Hence, learning algorithms may require many more samples to
learn the unknown parameter, 
leading to larger regret. It is important
to note that our algorithm does not require any prior information on
$C_\mathcal{X}$. 
The regret bound for the $K$-armed sparse bandits under
Assumption~\ref{assum:balanced_cov}  is as follows.

\begin{theorem}[Regret bound for $K$ arms]\label{thm:GLM-Lasso-regret-bound_CC_K-arm}
Suppose $K \geq 3$ and Assumptions~\ref{assum:feature_param_bounds}-\ref{assum:rho}, and \ref{assum:balanced_cov} hold. 
Let $\lambda_0 = 2 \sigma x_{\max}$. Then the expected cumulative regret of the \textsc{SA~Lasso~Bandit} policy $\pi$ over horizon $T \geq 1$  is upper-bounded by
\begin{align*}
    \mathcal{R}^\pi(T) &\leq  4\kappa_1 + \frac{4 \kappa_1 x_{\max} b (\log(2d^2)+1)}{C_1(s_0)^2} + \frac{64 \kappa_1\nu C_\mathcal{X}\sigma x_{\max} s_0 \sqrt{  T \log (dT) }}{\kappa_0\phi^2_0} 
\end{align*}
where $C_1(s_0) = \min\!\Big(\frac{1}{2}, \frac{\phi_0^2}{256s_0 \nu C_\mathcal{X} x_{\max}^2} \Big)$.
\end{theorem}

Theorem~\ref{thm:GLM-Lasso-regret-bound_CC_K-arm} establishes $\mathcal{O}\big( s_0 \sqrt{  T \log (dT)} \big)$ regret without prior knowledge on $s_0$, achieving the same rate as Theorem~\ref{thm:GLM-Lasso-regret-bound_CC} in terms of the key problem primitives. 
Since both  multivariate  Gaussian distributions and uniform distributions satisfy Assumption~\ref{assum:rho} with $\nu = 1$ and Assumption~\ref{assum:balanced_cov} with $C_\mathcal{X} = \mathcal{O}(1)$, 
the regret bound in Theorem~\ref{thm:GLM-Lasso-regret-bound_CC_K-arm}  still holds under Assumptions~\ref{assum:feature_param_bounds}-\ref{assum:comp_condition} for these distributions. 
Therefore, to our knowledge, this is the first sparsity-agnostic regret bound for a general $K$-armed high-dimensional contextual bandit algorithm even for the Gaussian distribution or uniform distribution.

The proof of Theorem~\ref{thm:GLM-Lasso-regret-bound_CC_K-arm} largely follows that of Theorem~\ref{thm:GLM-Lasso-regret-bound_CC}. 
The main difference is how we control the adapted Gram matrix~$\Sigma_t$
with the theoretical Gram matrix~$\Sigma$. Under the balanced covariance
condition, we can ensure the lower bound of the adapted Gram matrix as a
function of the theoretical Gram matrix, which is analogous to the result
in Lemma~\ref{lemma:cov_matrix_lowerbound_asymmetry}. In particular, we
can show that for a fixed vector $\beta \in \mathbb{R}^d$,  
\begin{equation*}
     \sum_{i=1}^K \mathbb{E}_{\mathcal{X}_t}\Big[X_{t,i} X_{t,i}^\top \mathbbm{1}\{X_{t,i} =
     \argmax_{\mathclap{X \in \mathcal{X}_t}} X^\top \beta\}\Big]
     \succcurlyeq (2 \nu
       C_\mathcal{X})^{-1} \Sigma \,.
\end{equation*}
The formal result is presented in
Lemma~\ref{lemma:cov_matrix_lowerbound_asymmetry_K-arm} in Appendix~\ref{sec:K-arm-regret-analysis}
along with its proof. Next, we again invoke the matrix concentration result
in Lemma~\ref{lemma:Sigma_hat_concentration} to connect the compatibility
constant of empirical Gram matrix $\hat{\Sigma}_t$ to that of $\Sigma_t$,
and eventually to the theoretical Gram matrix $\Sigma$. Thus, we ensure
the compatibility condition of $\hat{\Sigma}_t$. The additional 
regret 
in the $K$-armed case as compared to the two-armed case is
essentially a scaling by $C_\mathcal{X}$ to ensure the balanced covariance
condition.

\subsection{Numerical Experiments for \textit{K} Arms}\label{sec:experiments_K-arms}

We now validate the performance of \textsc{SA Lasso Bandit} in $K$-armed sparse bandit settings via additional numerical experiments and provide comparison with the existing sparse bandit algorithms. The setup of the experiments is identical to the setup described in Section~\ref{sec:experiments}. We perform evaluations under various instances. In particular, we focus on the performances of algorithms as the number of arms increases. Additionally, to investigate the effect of the balanced covariance condition, we evaluate algorithms on features drawn from a non-Gaussian elliptical distribution, for which we do not have a tight bound of $C_\mathcal{X}$ as well as the multi-dimensional uniform distribution.

\begin{figure*}[t]
    \begin{subfigure}[b]{0.245\textwidth}
        \includegraphics[width=\textwidth]{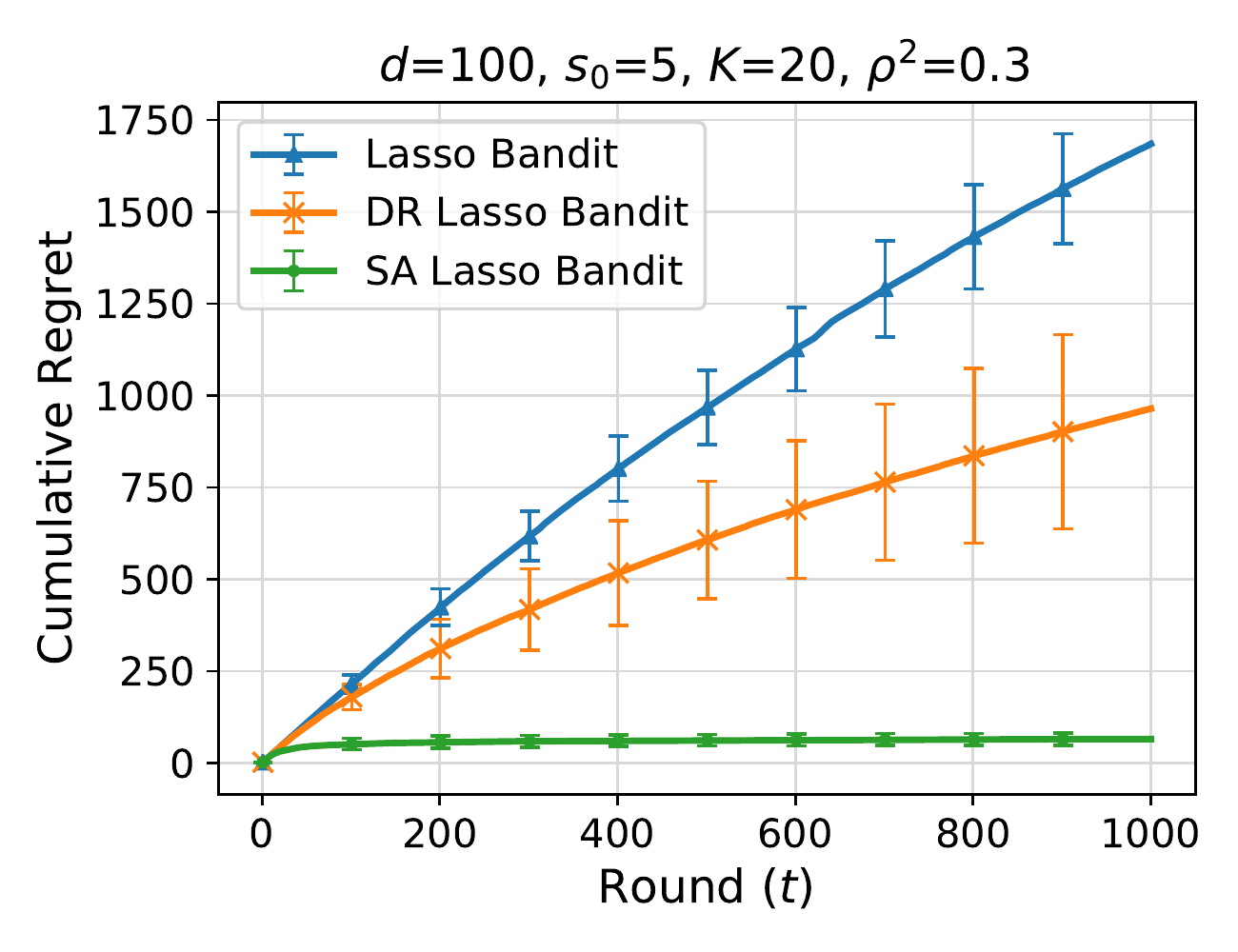}
    \end{subfigure}
    \begin{subfigure}[b]{0.245\textwidth}
        \includegraphics[width=\textwidth]{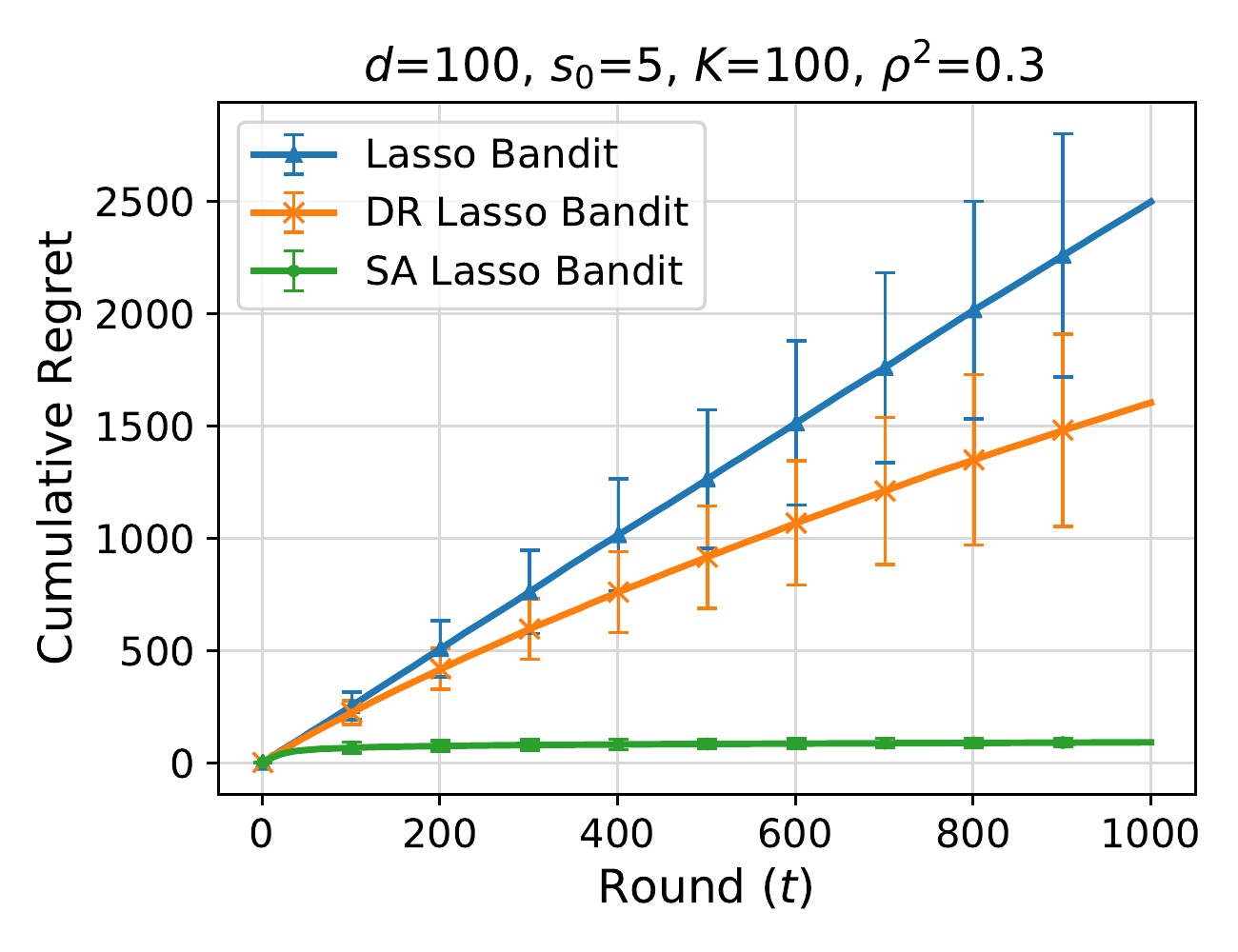}
    \end{subfigure}
    \begin{subfigure}[b]{0.245\textwidth}
        \includegraphics[width=\textwidth]{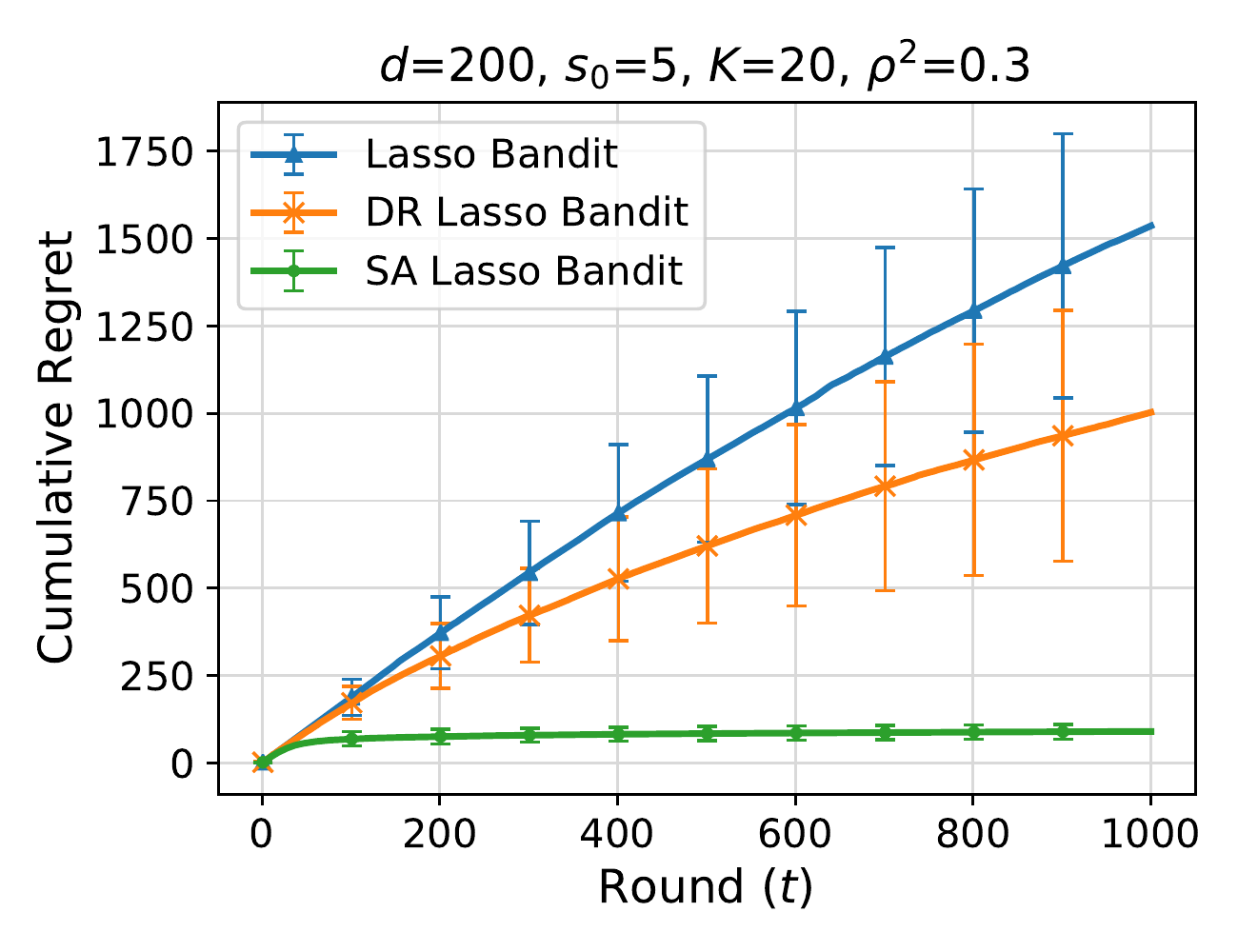}
    \end{subfigure}
    \begin{subfigure}[b]{0.245\textwidth}
        \includegraphics[width=\textwidth]{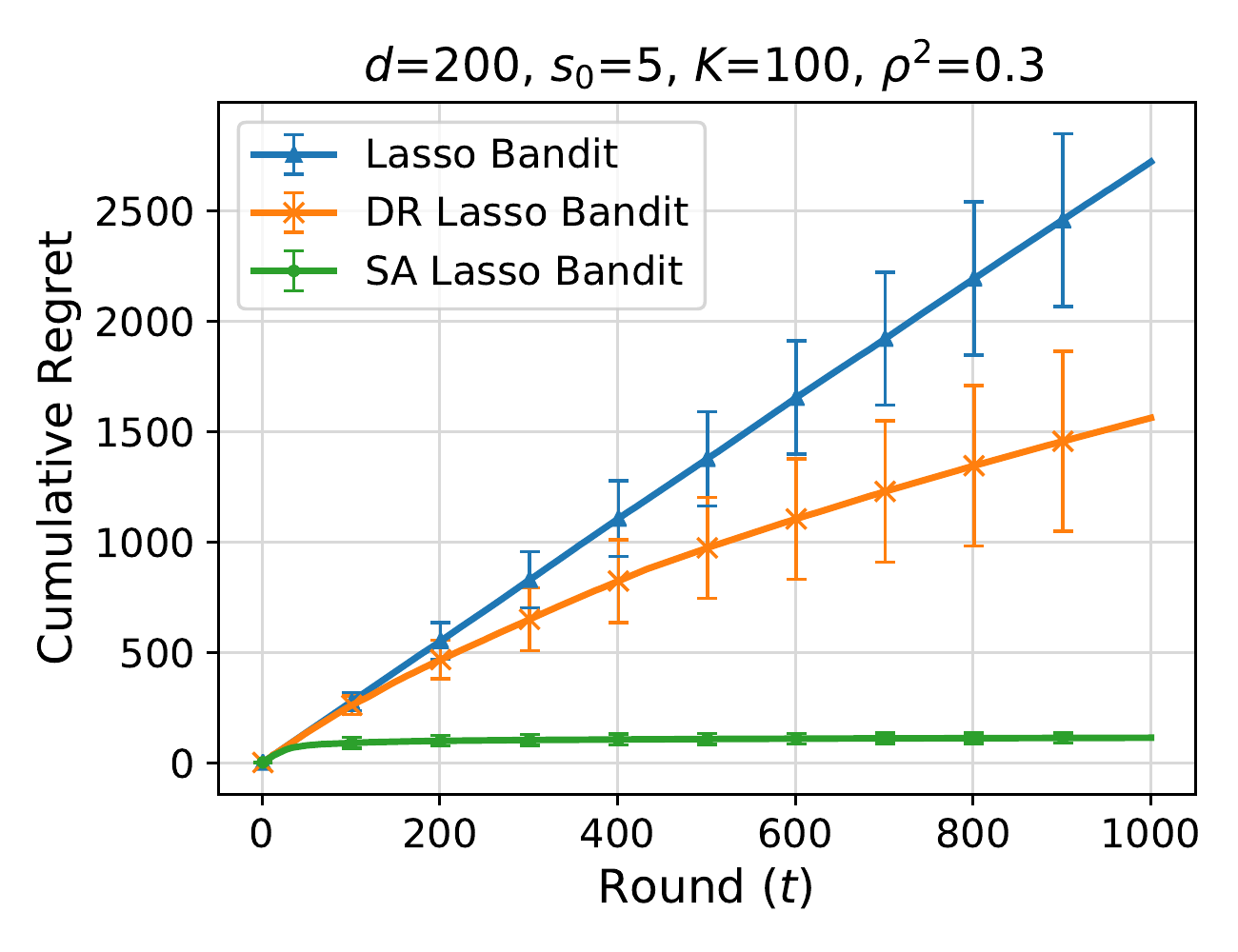}
    \end{subfigure}\\
    \begin{subfigure}[b]{0.245\textwidth}
        \includegraphics[width=\textwidth]{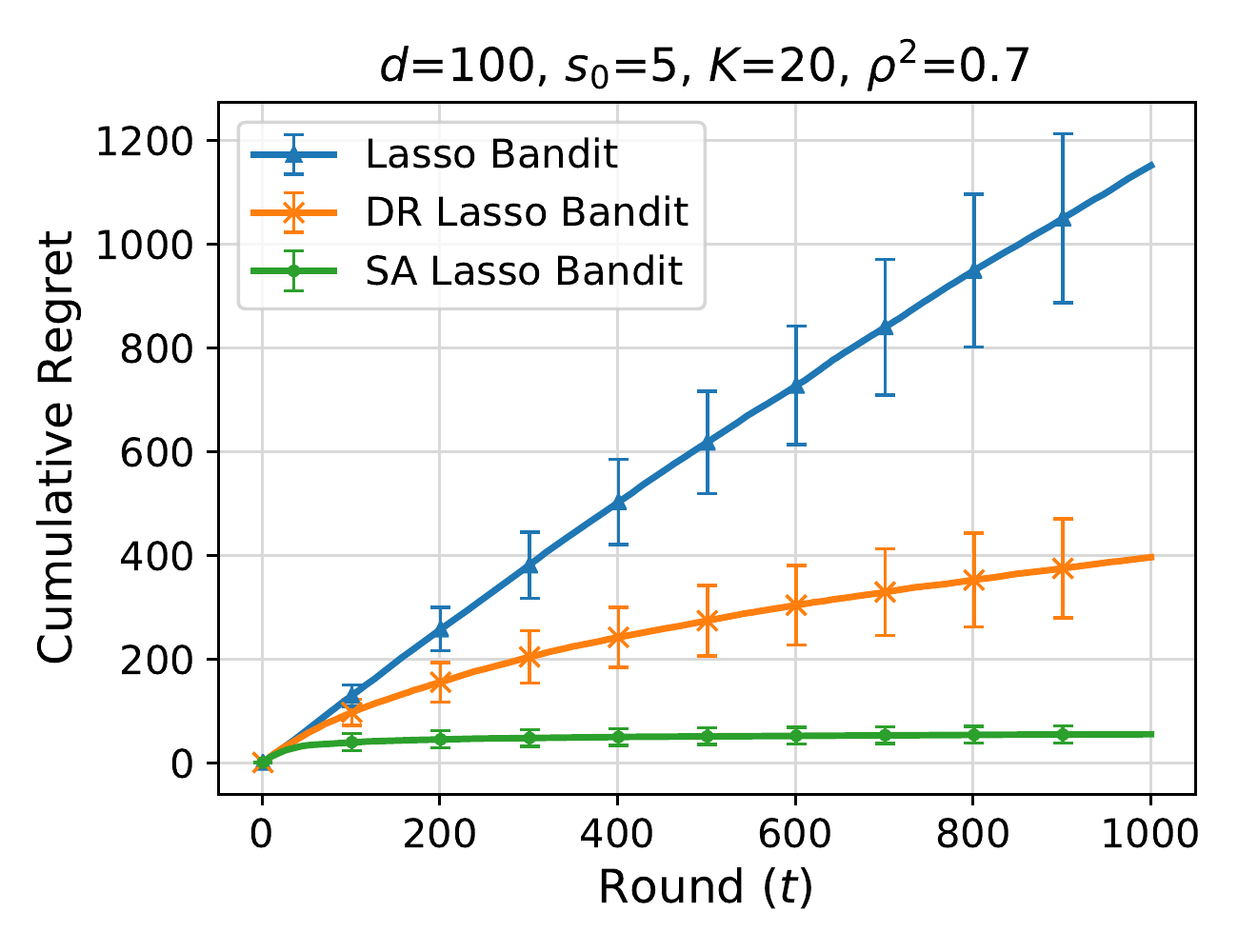}
    \end{subfigure}
    \begin{subfigure}[b]{0.245\textwidth}
        \includegraphics[width=\textwidth]{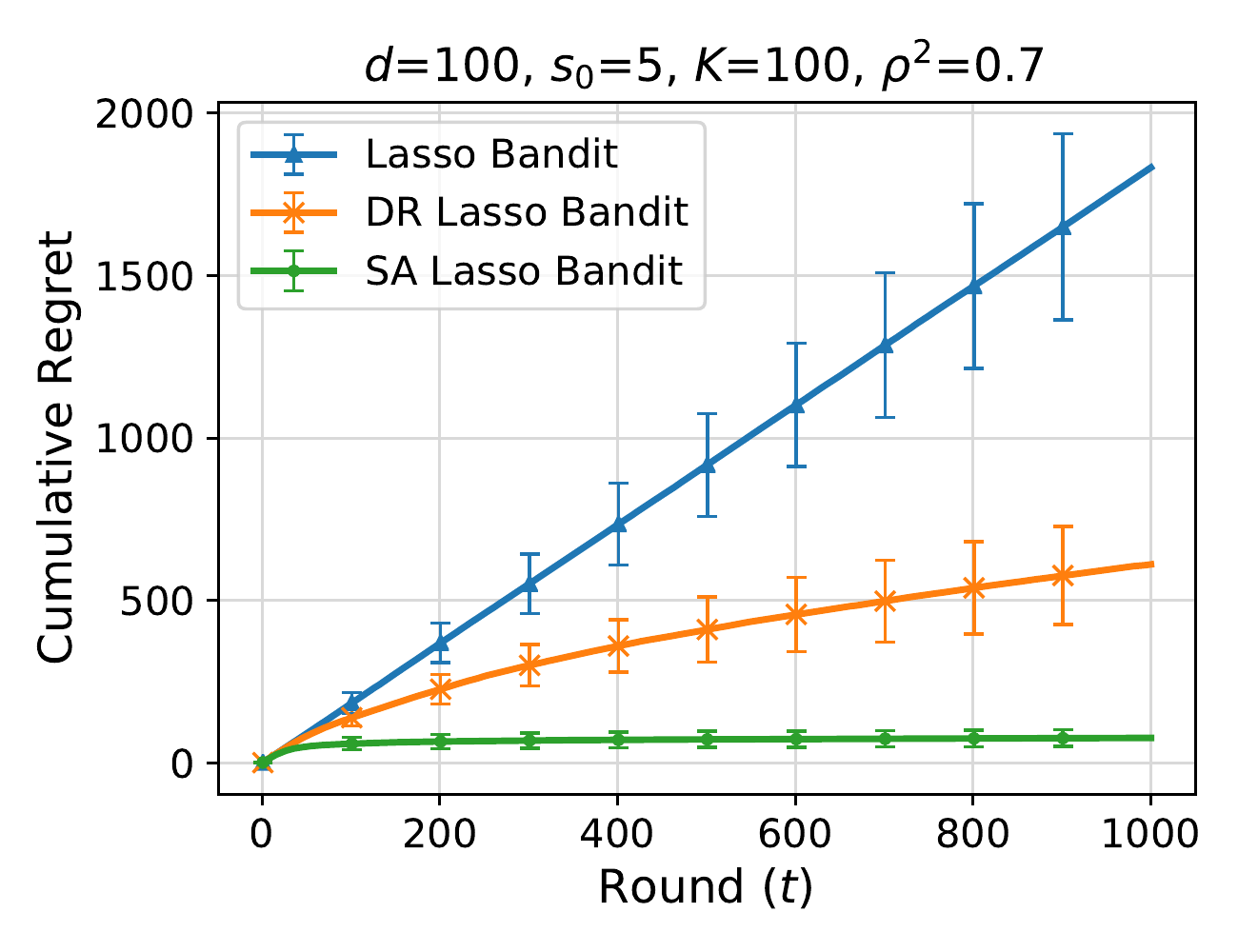}
    \end{subfigure}
    \begin{subfigure}[b]{0.245\textwidth}
        \includegraphics[width=\textwidth]{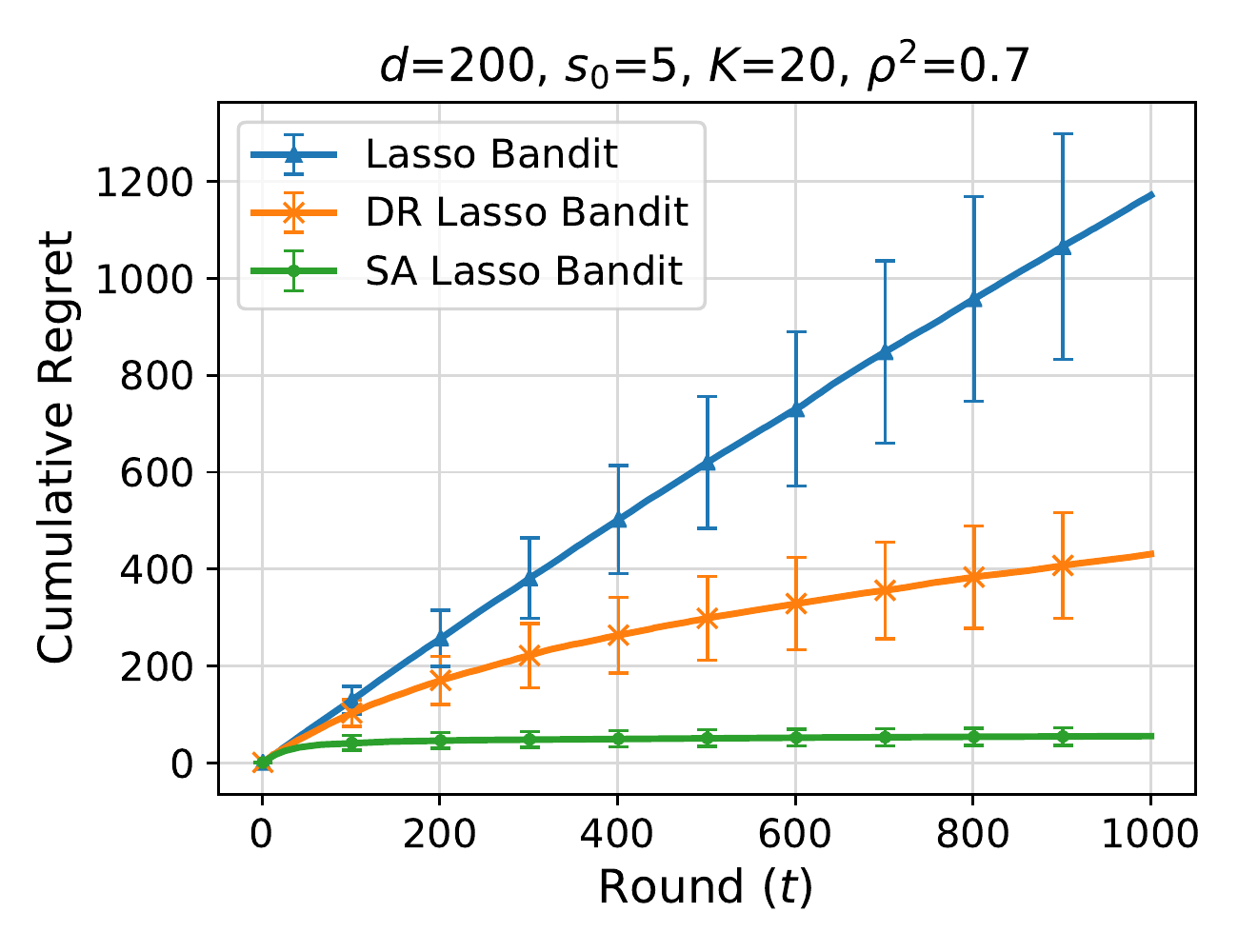}
    \end{subfigure}
    \begin{subfigure}[b]{0.245\textwidth}
        \includegraphics[width=\textwidth]{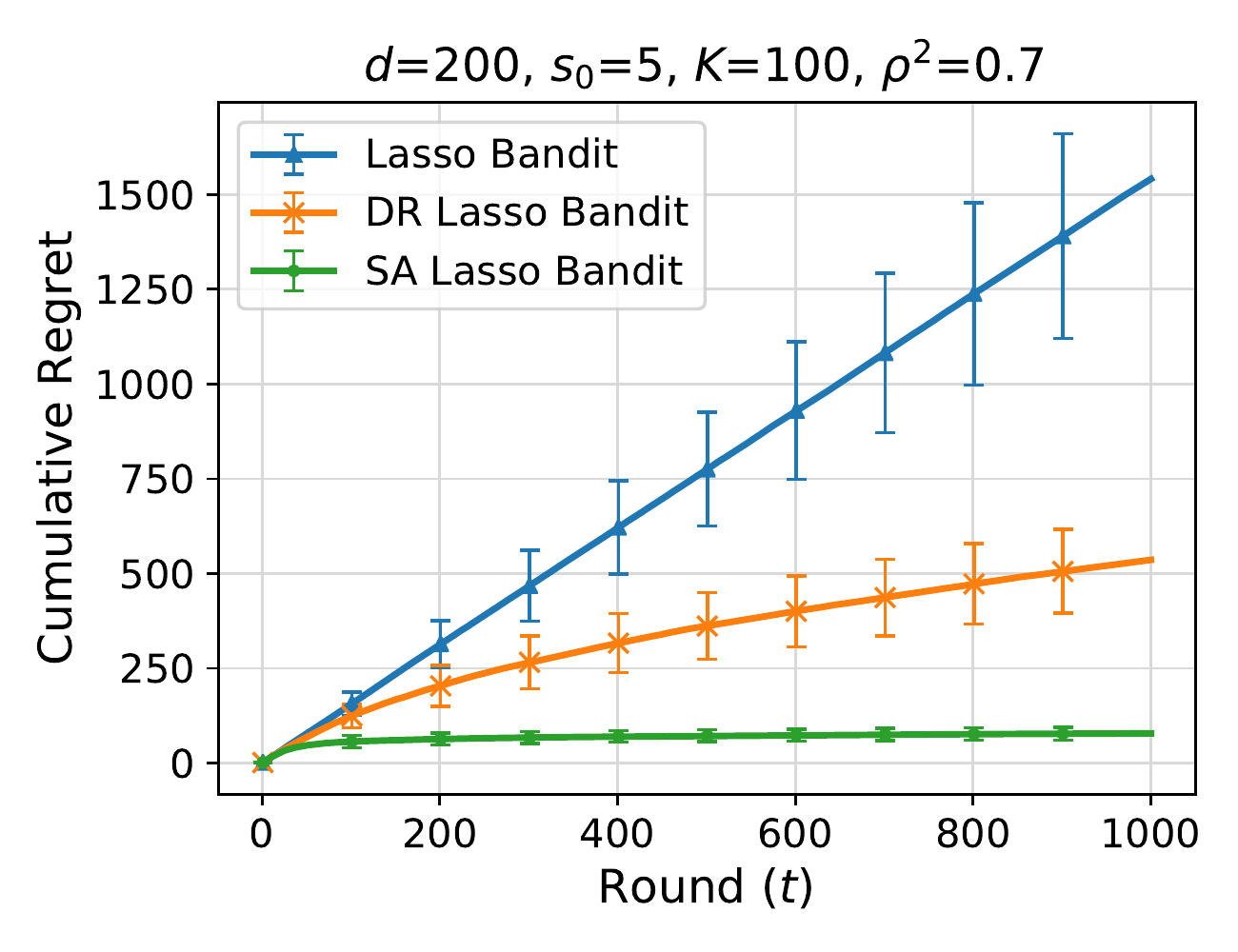}
    \end{subfigure}\\
     \begin{subfigure}[b]{0.245\textwidth}
        \includegraphics[width=\textwidth]{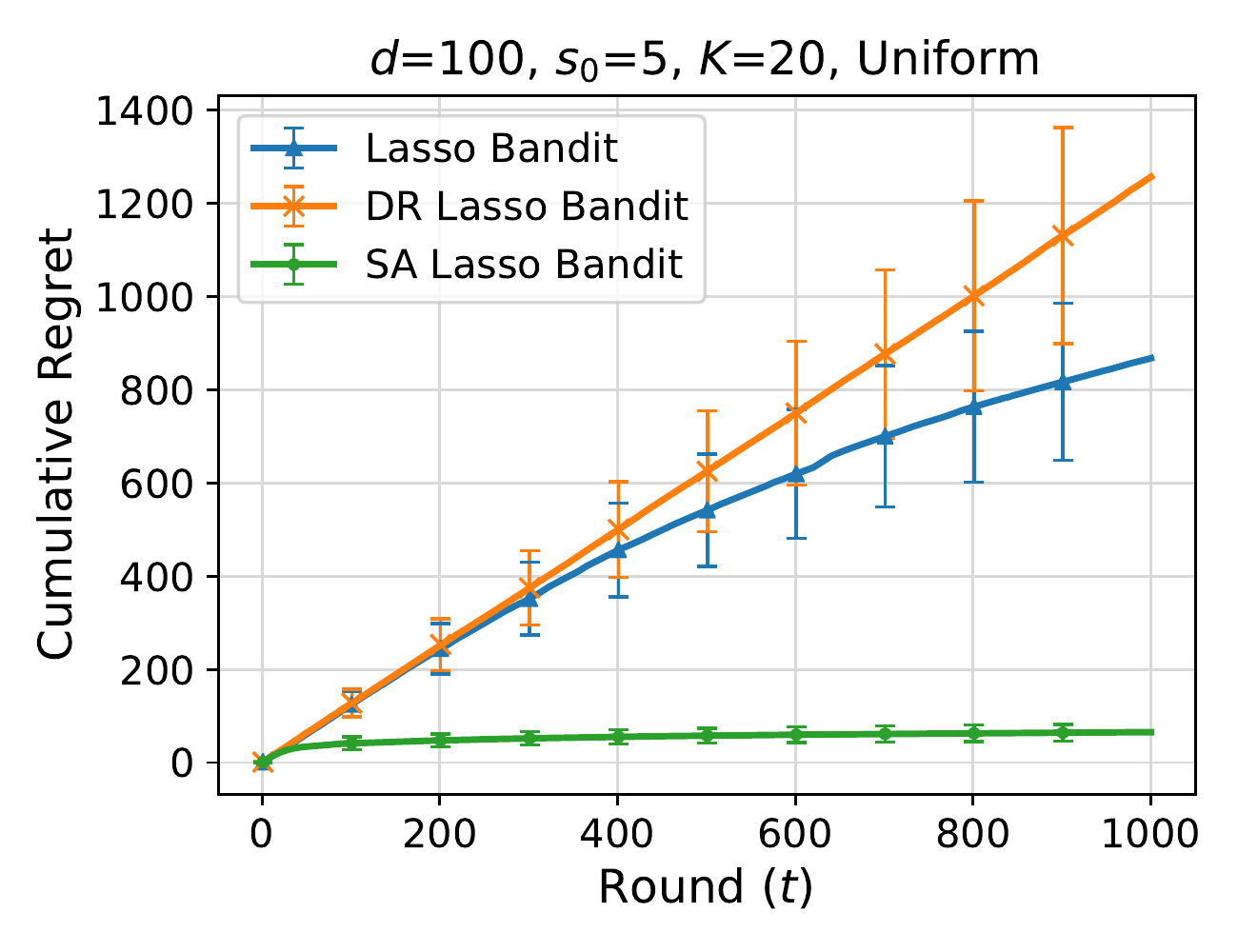}
    \end{subfigure}
    \begin{subfigure}[b]{0.245\textwidth}
        \includegraphics[width=\textwidth]{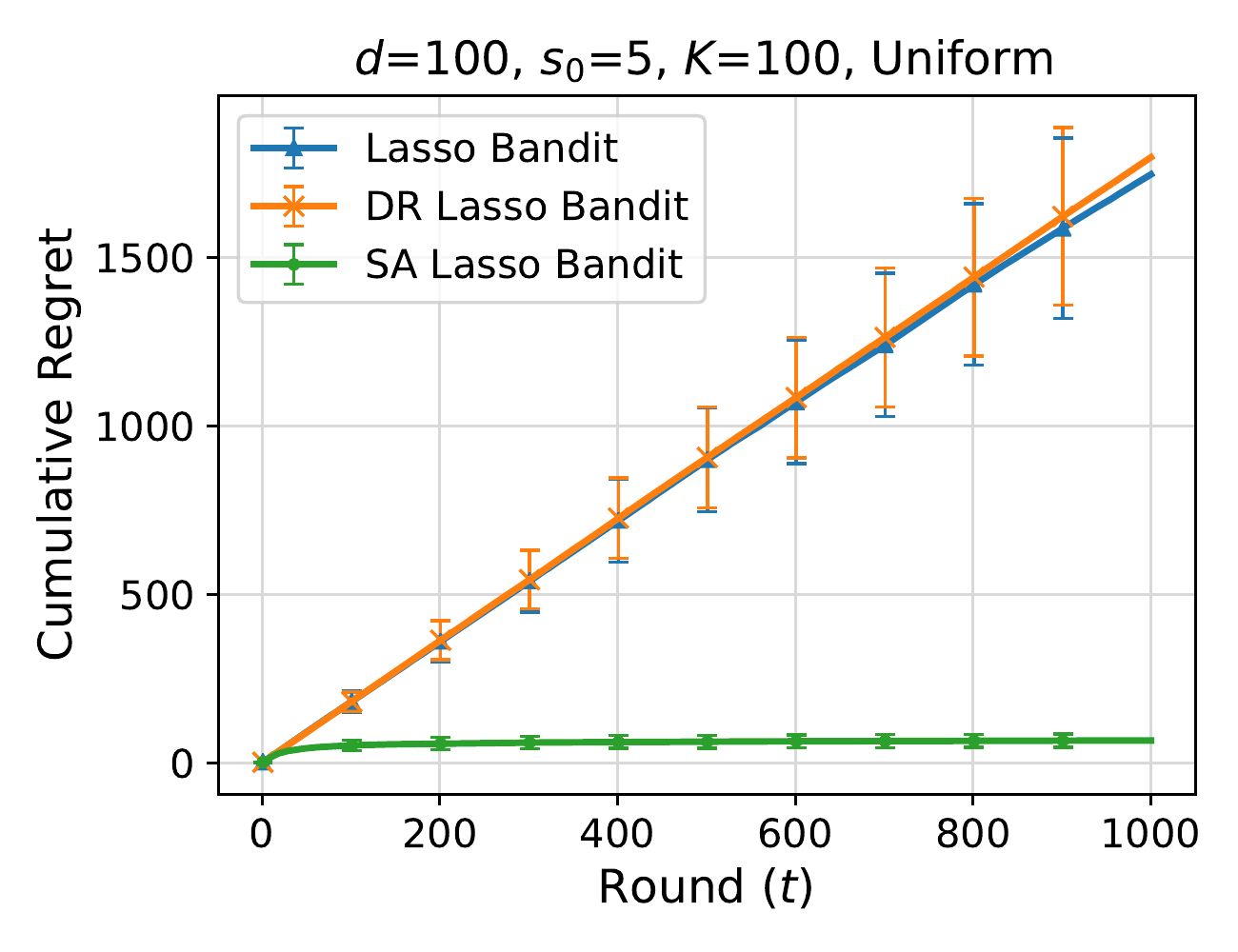}
    \end{subfigure}
    \begin{subfigure}[b]{0.245\textwidth}
        \includegraphics[width=\textwidth]{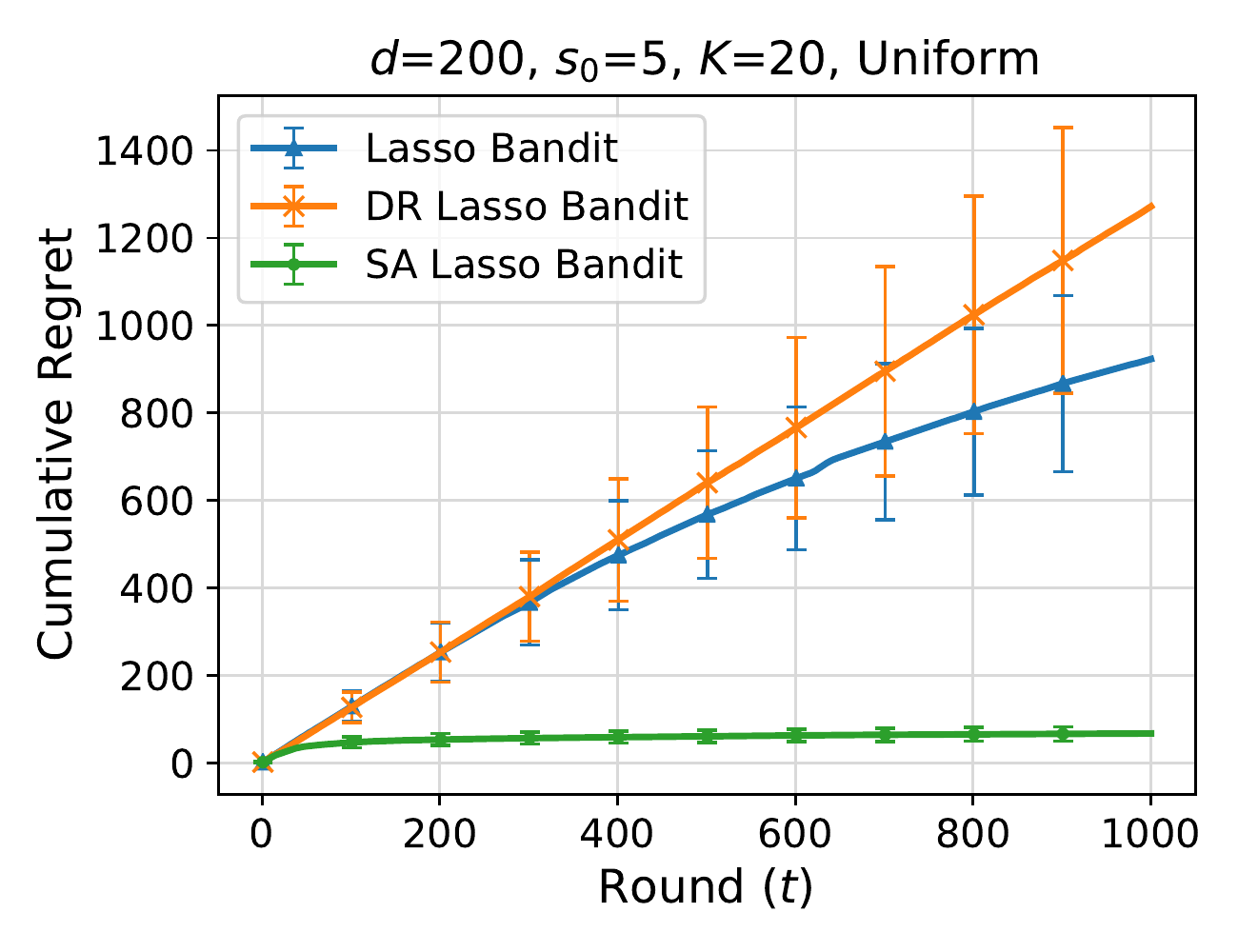}
    \end{subfigure}
        \begin{subfigure}[b]{0.245\textwidth}
        \includegraphics[width=\textwidth]{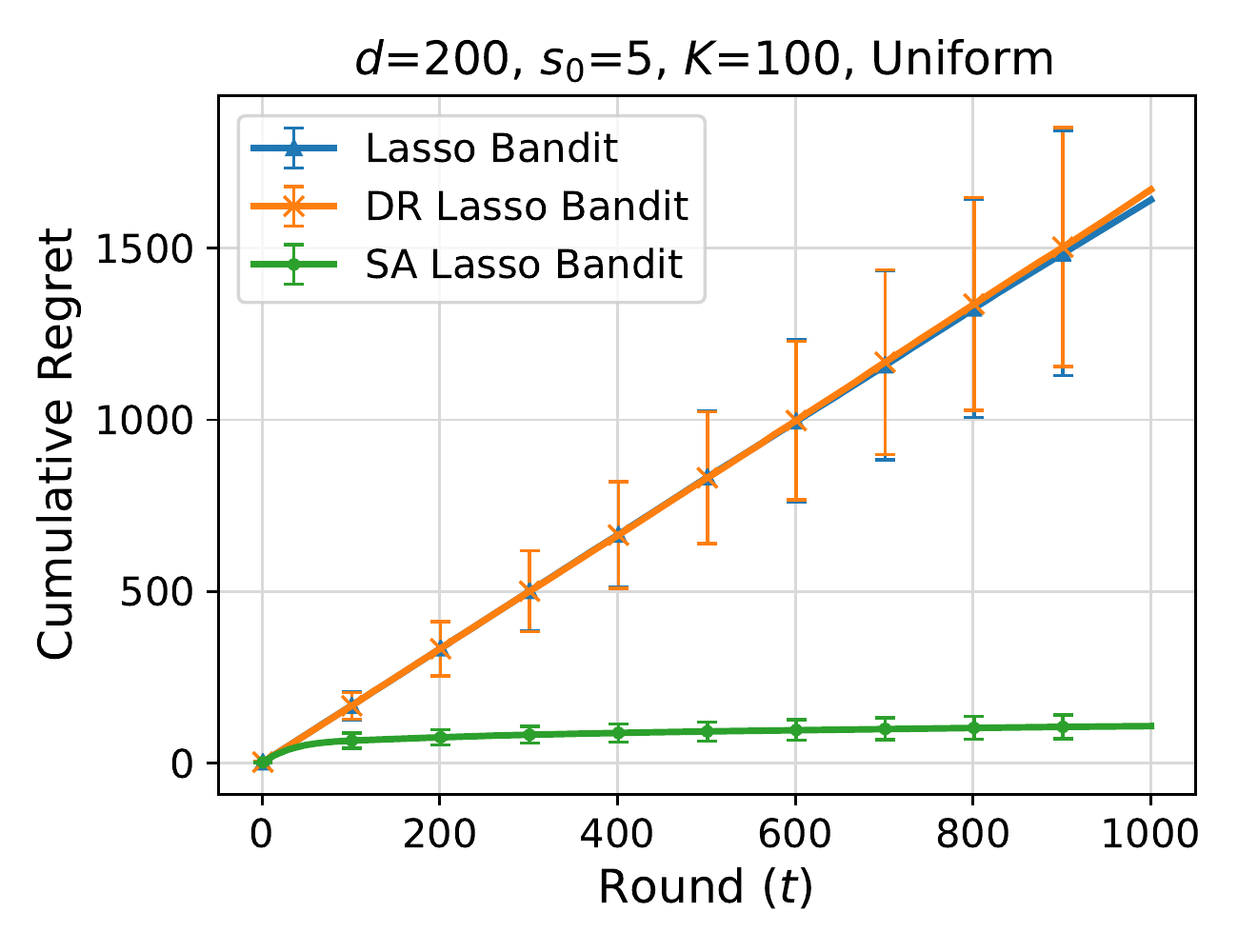}
    \end{subfigure}\\
     \begin{subfigure}[b]{0.245\textwidth}
        \includegraphics[width=\textwidth]{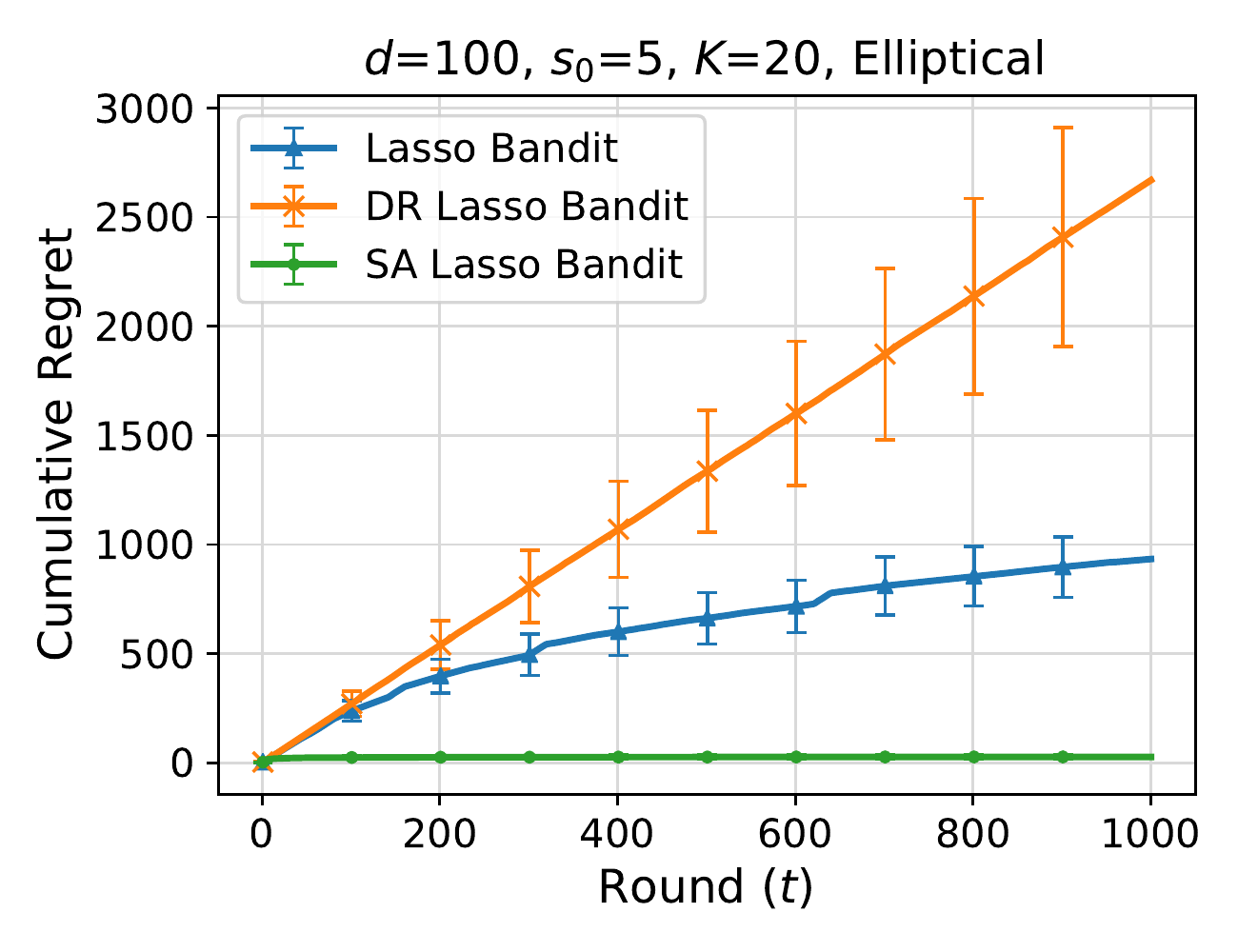}
    \end{subfigure}
    \begin{subfigure}[b]{0.245\textwidth}
        \includegraphics[width=\textwidth]{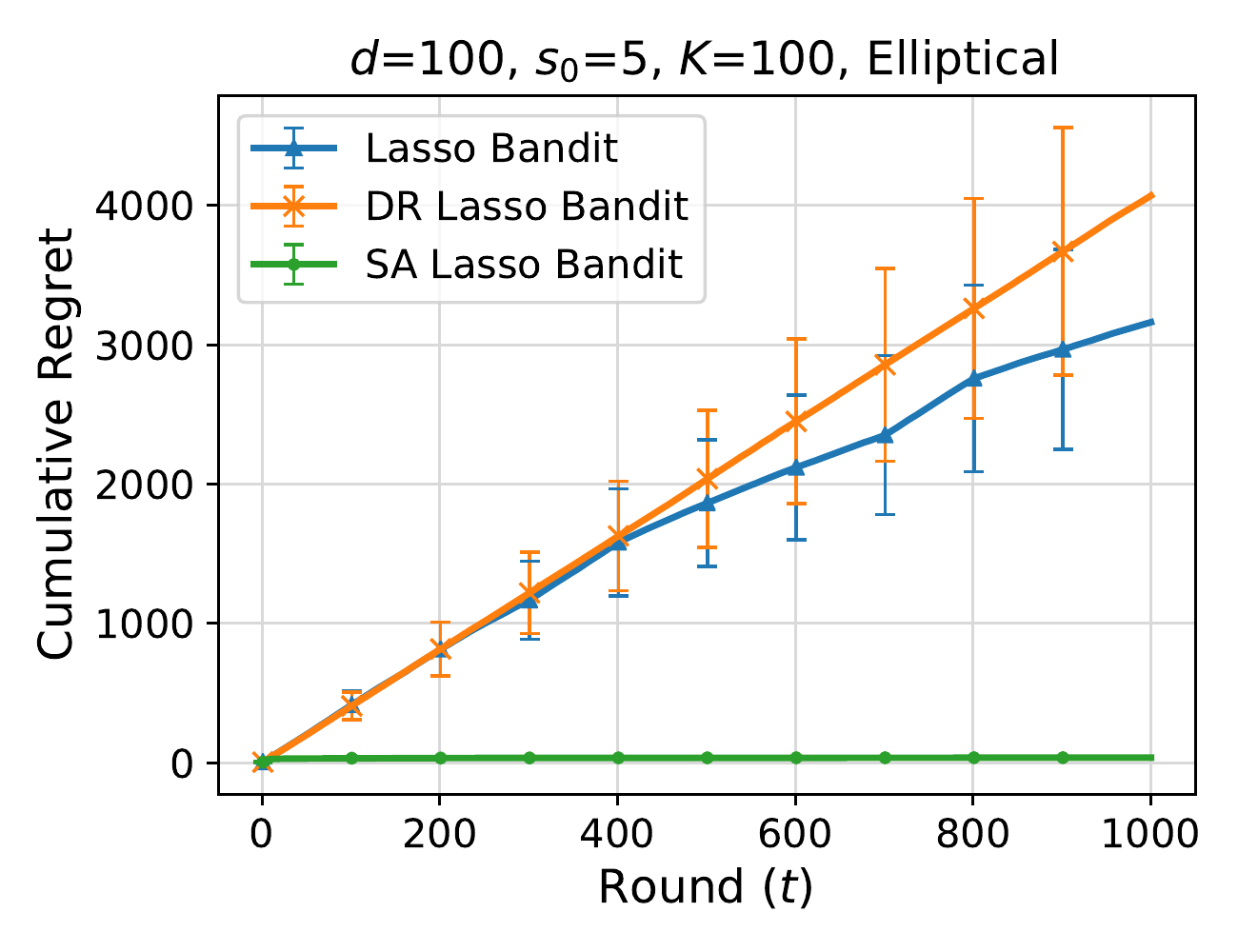}
    \end{subfigure}
    \begin{subfigure}[b]{0.245\textwidth}
        \includegraphics[width=\textwidth]{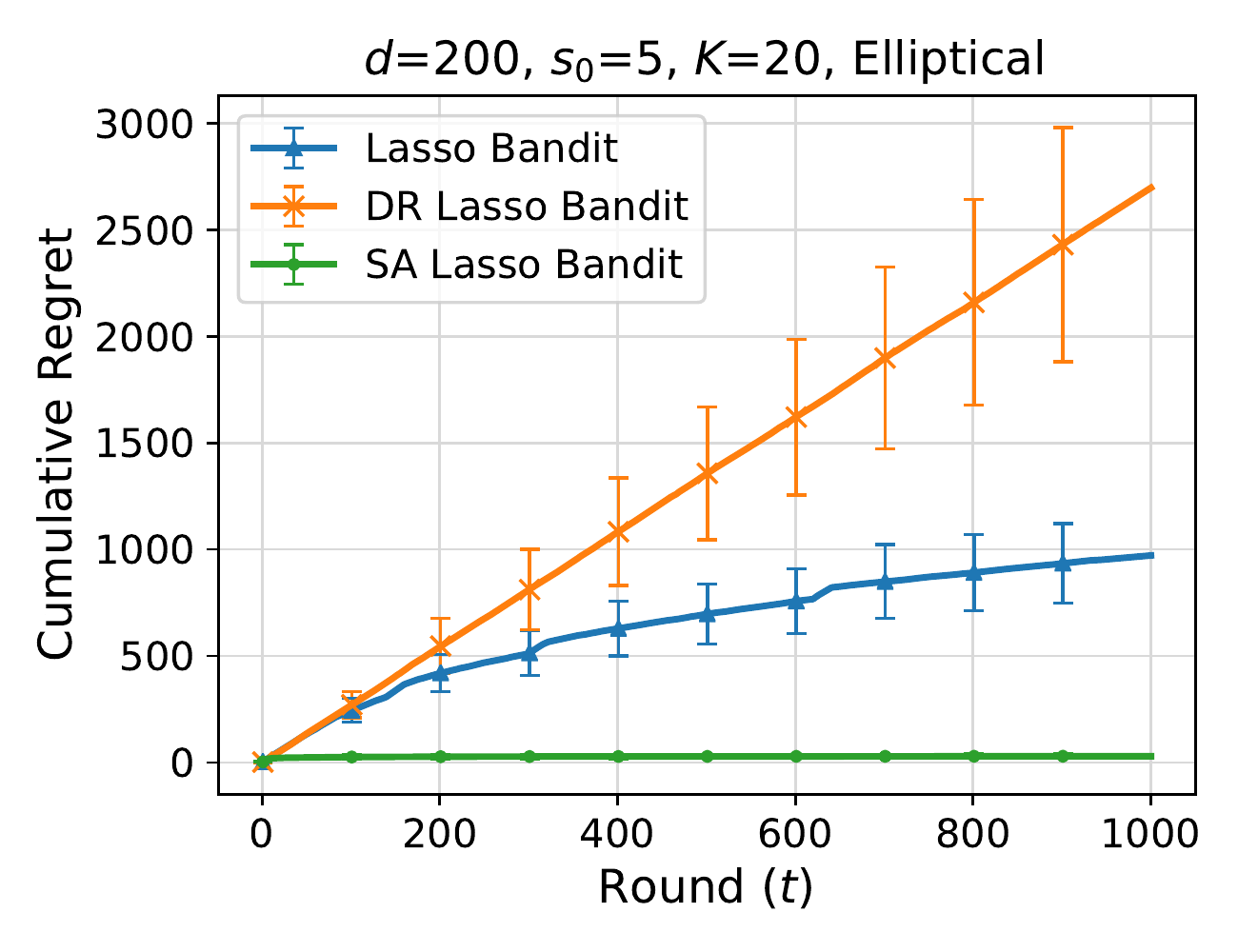}
    \end{subfigure}
        \begin{subfigure}[b]{0.245\textwidth}
        \includegraphics[width=\textwidth]{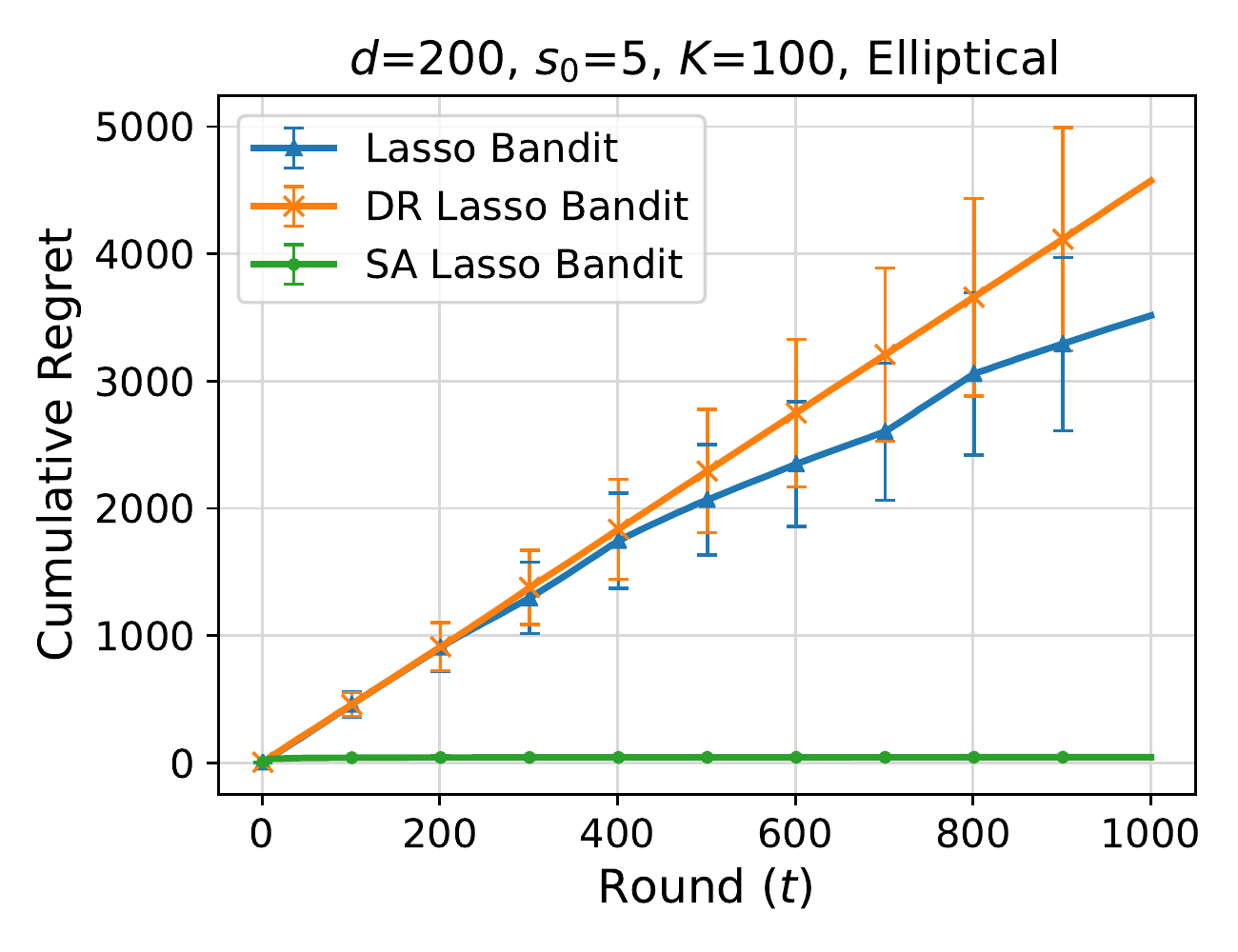}
    \end{subfigure}
    \caption{\small The plots show the $t$-round cumulative regret of \textsc{SA Lasso Bandit} (Algorithm~\ref{algo:SA_Lasso_bandit}), \textsc{DR Lasso Bandit} \citep{kim2019doubly}, and \textsc{Lasso Bandit} \citep{bastani2020online} with varying number of arms $K \in \{20, 100\}$, feature dimensions $d \in \{100, 200\}$, and different distributions. In the first two rows, features are drawn from a multivariate Gaussian distribution with weak and strong correlation levels. The third row shows evaluations with features drawn from the multi-dimensional uniform distribution. In the fourth row, features are drawn from a non-Gaussian elliptical distribution.} 
     \label{fig:K_arm}
\end{figure*}

Figure~\ref{fig:K_arm} shows results of the numerical evaluations (averaged over 20 independent runs per problem instance);
additional results are presented in the appendix.
The results provide convincing evidence that the performance of our proposed algorithm is superior to the existing sparse bandit methods that we compare with. 
Again, \textsc{SA Lasso Bandit} outperforms the existing sparse bandit algorithms by significant margins, even though the correct sparsity index $s_0$ is revealed to these algorithms and kept hidden from \textsc{SA Lasso Bandit}.
Furthermore,
\textsc{SA Lasso Bandit} is much more practical and simple to implement with a minimal number of a hyperparameter. 

In the experiments with Gaussian distributions shown in the first and second rows in Figure~\ref{fig:K_arm}, we again observe that algorithms generally perform better under strong correlation compared to weak correlation instances.
This is expected since strongly (positively) correlated arms imply a smaller discrepancy between expected payoffs of the arms. 
A strong correlation between the arms also implies a smaller $C_\mathcal{X}$, hence leading to a lower regret, as briefly discussed earlier when we introduce the balanced covariance condition. 
Thus, the balanced covariance condition appears to capture the essence of positive correlation between arms.
It is important to note that there are two different notions of correlation: correlation between the arms and correlation between the features of an arm. 
A higher correlation between the features
potentially decreases the value of compatibility constant. Thus, the regret may increase with an increase in correlation of the features as far as the compatibility condition is concerned.
The plots in the third and fourth rows in Figure~\ref{fig:K_arm} show that
when the feature vectors are drawn i.i.d. according to the uniform
distribution and non-Gaussian elliptical distributions, 
the performance of existing algorithms (e.g., \textsc{DR Lasso Bandit} from \citet{kim2019doubly}) deteriorates significantly; 
\textsc{SA Lasso Bandit} still exhibits superior performances. Thus, our
proposed algorithm is very robust to the changes in the distribution of
the feature vectors.

\section{Concluding Remarks}\label{sec:conclusion}
In this paper, we study a high-dimensional contextual bandit problem with
sparse structure. Previously known 
learning algorithms for this problem require  a priori knowledge of the
sparsity index $s_0$ of the unknown
parameter. Our goal in this paper is to remove this restriction. 
 We propose and analyze an algorithm that 
does not require this information.
The proposed algorithm achieves a tight regret upper bound which depends
on a logarithmic function of the feature dimension which matches the
scaling of the offline Lasso convergence results. The algorithm attains
this sharp result without knowing the sparsity of the unknown parameter,
overcoming a weakness of existing algorithms.  
We demonstrate that our proposed algorithm significantly outperforms the
benchmark, supporting the theoretical claims.  We conclude by outlining
some future directions.
\smallskip

\noindent \textbf{Minimax Regret in Sparse Bandits.}
Minimax regret in sparse bandits is more subtle to define than in
(non-sparse) linear or GLM bandits. 
Consider the following setting.  
Suppose nature is allowed to freely choose $s_0 \in [d]$, it can force the
regret for any sparse bandit algorithm to be polynomial in $d$ by choosing
$s_0 = d$.   
On the other hand, if we limit nature to choose $s_0 \in [1,s_{\max}]$, it
will choose $s_0 = s_{\max}$, and therefore, sparse bandit algorithms can
assume that the sparsity index $s_0$ is known, and  equal to  $s_{\max}$. 
Thus, it is not clear how to define a minimax criterion in a manner that
does not reveal the dominating choice for nature,  
and therefore, forces learning algorithm to play a strategy which hedges against
a range of values of the sparsity index. 
\smallskip

\noindent  \textbf{Reinforcement Learning with High-Dimensional Covariates.}
Another compelling direction is to extend our analysis and proposed approach to reinforcement learning with high-dimensional context or with high-dimensional function approximation. A main challenge in this direction appears to be the need for an algorithm to be optimistic. To our knowledge, almost all reinforcement learning algorithms with provable efficiency rely on the principle of optimism \citep{jaksch2010near,jin2018q}. But, as we have discussed in this paper, in order to be optimistic in the tightest sense under sparse structure, the knowledge on sparsity is generally needed. 


\vskip 0.2in
\bibliography{ref}

\newpage
\appendix

\section{Proofs of Lemmas for Theorem~\ref{thm:GLM-Lasso-regret-bound_CC}}
\label{app:Theorem1}

\subsection{Proof of Lemma~\ref{lemma:oracle_ineq_GLM_L1}}

The proof follows from modifying the proof of the standard Lasso oracle inequality \citep{buhlmann2011statistics} using martingale theory.
Recall from \eqref{eq:Lasso_glm} that the negative log-likelihood of the GLM is 
\begin{equation*}
    \ell_t(\beta) =   -\frac{1}{t}\sum_{\tau=1}^t \left[ Y_\tau X_\tau^\top \beta - m(X_\tau^\top \beta) \right]
\end{equation*}
where $m$ is a normalizing function with its gradient $\dot{m}(X^\top \beta) = \mu(X^\top \beta)$.
Now, we denote the expectation of $\ell_t(\beta)$ over $Y$ by $\Bar{\ell}_t(\beta)$:
\begin{equation*}
    \Bar{\ell}_t(\beta) := \mathbb{E}_Y[\ell_t(\beta)] = -\frac{1}{t}\sum_{\tau=1}^t \left[ \mu(X_\tau^\top \beta^*) X_\tau^\top \beta - m(X_\tau^\top \beta) \right] .
\end{equation*}
Note that $\nabla_\beta \Bar{\ell}_t(\beta) = -\frac{1}{t}\sum_{\tau=1}^t \left[ \mu(X_\tau^\top \beta^*)  -  \mu(X_\tau^\top \beta)   \right] X_\tau$. Hence, we have $\nabla_\beta \Bar{\ell}_t(\beta^*) = \vec{0}_d$ which implies that $\beta^* = \argmin_{\beta}  \Bar{\ell}_t(\beta)$ given the fact that $m$ is convex in the GLM.
Hence, for any parameter $\beta \in \mathbb{R}^d$, the excess risk is defined as
\begin{equation*}
    \mathcal{E}(\beta) := \Bar{\ell}_t(\beta) - \Bar{\ell}_t(\beta^*).
\end{equation*}
Note that by definition, $\mathcal{E}(\beta) \geq 0$, for all $\beta \in \mathbb{R}^d$ (with $\mathcal{E}(\beta^*) = 0$).
The Lasso estimate $\hat{\beta}_t$ for the GLM is given by the minimization of the penalized negative log-likelihood
\begin{align*}
    \hat{\beta}_t := \argmin_{\beta} \big\{ \ell_t(\beta) + \lambda_t\| \beta \|_1 \big\}
\end{align*}
where $\lambda$ is the penalty parameter whose value needs to be chosen to control the noise of the model. Now, we define the empirical process of the problem as
\begin{equation*}
    v_t(\beta) := \ell_t(\beta) - \Bar{\ell}_t(\beta).
\end{equation*}
Note that the randomness in $\{Y_\tau\}$ still plays a role on $\ell_t(\beta)$ and hence on $v_t(\beta)$.
Then by the definition of $\hat{\beta}_t$, we have
\begin{align*}
    \ell_t(\hat{\beta}_t) + \lambda_t\| \hat{\beta}_t \|_1  \leq \ell_t(\beta^*) + \lambda_t\| \beta^* \|_1.
\end{align*}
Adding and subtracting terms, we have
\begin{align*}
        \ell_t(\hat{\beta}_t) - \Bar{\ell}_t(\hat{\beta}_t) + \Bar{\ell}_t(\hat{\beta}_t) - \Bar{\ell}_t(\beta^*)+ \lambda_t\| \hat{\beta}_t \|_1  \leq \ell_t(\beta^*) - \Bar{\ell}_t(\beta^*) + \lambda_t\| \beta^* \|_1 \,.
\end{align*}
Rearranging terms gives the following ``basic inequality'' for the GLM
\begin{equation*}\label{eq:basic_ineq_MNL}
    \mathcal{E}(\hat{\beta}_t) + \lambda_t\| \hat{\beta}_t \|_1 \leq -[v_t(\hat{\beta}_t) - v_t(\beta^*)] + \lambda_t\| \beta^* \|_1 \,.
\end{equation*}
The basic inequality implies that in order to provide an upper-bound for the penalized excess risk, we need to control the deviation of the empirical process $[v_t(\hat{\beta}_t) - v_t(\beta^*)]$ \citep{buhlmann2011statistics}. And we bound this deviation of the empirical process in terms of the parameter estimation error $\| \hat{\beta}_t - \beta^*\|_1$. 
Essentially, $[v_t(\hat{\beta}_t) - v_t(\beta^*)]$ is where the random noise plays a role, and with large enough penalization (suitably large $\lambda$) we can control such randomness in the empirical process.
We define the event of the empirical process being controlled by the penalization.
\begin{equation}\label{eq:event_T}
    \mathcal{T} := \{ | v_t(\hat{\beta}_t) - v_t(\beta^*) | \leq \lambda \| \hat{\beta}_t - \beta^*\|_1 \} \,.
\end{equation}
Lemma~\ref{lemma:GLM_event_T_adaptive} ensures that we can control this empirical process deviation with high probability. Hence, in the rest of the proof, we restrict ourselves to the case where the empirical process behaves well, i.e., event $\mathcal{T}$ in \eqref{eq:event_T} holds.

\begin{lemma}\label{lemma:GLM_event_T_adaptive}
Assume $X_t$ satisfies $\|X_t\|_2 \leq x_{\max}$ for all $t$. If $\lambda = \sigma x_{\max} \sqrt{\frac{2[\log(2/\delta) + \log d]}{t}}$, then with probability at least $1 - \delta$ we have
\begin{equation*}
    | v_t(\hat{\beta}_t) - v_t(\beta^*) | \leq \lambda \| \hat{\beta}_t - \beta^*\|_1 \,.
\end{equation*}
\end{lemma}
On event $\mathcal{T}$, for $\lambda_t\geq 2\lambda$, we have
\begin{align}\label{eq:ub_for_basic_ineq}
        2\mathcal{E}(\hat{\beta}_t) + 2\lambda_t\| \hat{\beta}_t \|_1 \leq \lambda_t\| \hat{\beta}_t - \beta^*\|_1 + 2\lambda_t\| \beta^* \|_1 \,.
\end{align}
Let $\hat{\beta} := \hat{\beta}_t$ for brevity. 
Using the active set $S_0$, we can define the following:
\begin{align*}
    \beta_{j,S_0} := \beta_j \mathbbm{1}\{ j \in S_0\} \qquad \beta_{j,S^c_0} := \beta_j \mathbbm{1}\{ j \notin S_0\}
\end{align*}
so that $\beta_{S_0} = [\beta_{1,S_0}, ..., \beta_{d,S_0}]^\top$ has zero elements outside the set $S_0$ and the elements of $\beta_{S_0^c}$ can only be non-zero in the complement of $S_0$. 
We can then lower-bound  $\|\hat{\beta}\|_1$ using the triangle inequality,
\begin{align*}
    \|\hat{\beta}\|_1 &= \|\hat{\beta}_{S_0}\|_1 + \|\hat{\beta}_{S_0^c}\|_1\\
    &\geq \| \beta^*_{S_0} \|_1 - \| \hat{\beta}_{S_0} - \beta^*_{S_0}\|_1 + \|\hat{\beta}_{S_0^c}\|_1 \,.
\end{align*}
Also, we can rewrite 
\begin{align*}
    \|\hat{\beta} - \beta^*\|_1 &= \|\hat{\beta}_{S_0} - \beta^*_{S_0}\|_1 + \|\hat{\beta}_{S_0^c} - \beta^*_{S_0^c}\|_1\\
    &= \|\hat{\beta}_{S_0} - \beta^*_{S_0}\|_1 + \|\hat{\beta}_{S_0^c} \|_1 \,.
\end{align*}
Then we continue from \eqref{eq:ub_for_basic_ineq}
\begin{align*}
    2\mathcal{E}(\hat{\beta}) + 2\lambda_t\| \beta^*_{S_0} \|_1 - 2\lambda_t \| \hat{\beta}_{S_0} - \beta^*_{S_0}\| + 2\lambda_t \|\hat{\beta}_{S_0^c}\|_1 &\leq \lambda_t\|\hat{\beta}_{S_0} - \beta^*_{S_0}\|_1 + \lambda_t\|\hat{\beta}_{S_0^c} \|_1 + 2\lambda_t\| \beta^* \|_1\\
    &= \lambda_t\|\hat{\beta}_{S_0} - \beta^*_{S_0}\|_1 + \lambda_t\|\hat{\beta}_{S_0^c} \|_1 + 2\lambda_t\| \beta^*_{S_0} \|_1 \,.
\end{align*}
Therefore, we have
\begin{align}
       0\leq 2\mathcal{E}(\hat{\beta})
       &\leq  3\lambda_t\|\hat{\beta}_{S_0} - \beta^*_{S_0}\|_1 - \lambda_t\|\hat{\beta}_{S_0^c}\|_1 \label{eq:intermed_l1_active_bound}\\
       &= \lambda_t\left( 3\|\hat{\beta}_{S_0} - \beta^*_{S_0}\|_1 -  \|\hat{\beta}_{S_0^c} -  \beta^*_{S_0^c}\|_1 \right) \notag
\end{align}
Then the compatibility condition can be applied to the vector $\hat{\beta} - \beta^*$ which gives
\begin{align}\label{eq:apply_RE}
    \| \hat{\beta}_{S_0} - \beta^*_{S_0} \|^2_1 \leq s_0 (\hat{\beta} - \beta^*)^\top \hat{\Sigma} (\hat{\beta} - \beta^*) /\phi^2_t \,.
\end{align}
From \eqref{eq:intermed_l1_active_bound}, we have 
\begin{align*}
    2\mathcal{E}(\hat{\beta}) + \lambda_t\|\hat{\beta}_{S_0^c}\|_1
    &\leq  3\lambda_t\|\hat{\beta}_{S_0} - \beta^*_{S_0}\|_1  \,.
\end{align*}
Therefore, we have
\begin{align*}
    2\mathcal{E}(\hat{\beta})  + \lambda_t\| \hat{\beta} - \beta^* \|_1
    &= 2\mathcal{E}(\hat{\beta}) + \lambda_t\|\hat{\beta}_{S_0^c}\|_1 + \lambda_t\| \hat{\beta}_{S_0} - \beta^*_{S_0} \|_1\\
    &\leq 3 \lambda_t\|\hat{\beta}_{S_0} - \beta^*_{S_0}\|_1 + \lambda_t\| \hat{\beta}_{S_0} - \beta^*_{S_0} \|_1\\
    &= 4 \lambda_t\|\hat{\beta}_{S_0} - \beta^*_{S_0}\|_1\\
    &\leq 4 \lambda_t\sqrt{s_0 (\hat{\beta} - \beta^*)^\top \hat{\Sigma} (\hat{\beta} - \beta^*) }/\phi_t  \\
    &\leq \kappa_0(\hat{\beta} - \beta^*)^\top \hat{\Sigma} (\hat{\beta} - \beta^*) + \frac{4\lambda_t^2 s_0}{\kappa_0 \phi^2_t}\\
    &\leq 2\mathcal{E}(\hat{\beta})  + \frac{4\lambda^2 s_0}{\kappa_0 \phi^2_t}
\end{align*}
where the second inequality is from applying the compatibility condition \eqref{eq:apply_RE} and
the third inequality is by using $4uv \leq u^2 + 4v^2$ with $u = \sqrt{\kappa_0(\hat{\beta} - \beta^*)^\top \hat{\Sigma} (\hat{\beta} - \beta^*) }$ and $v = \frac{\lambda_t \sqrt{s_0}}{\phi_t \sqrt{\kappa_0}}$. The last inequality is from Lemma~\ref{lemma:margin_cond}.
Hence, rearranging gives
\begin{align*}
     \| \hat{\beta} - \beta^* \|_1 \leq \frac{ 4  s_0 \lambda_t}{\kappa_0\phi^2_t} \,.
\end{align*}
This completes the proof.
\endproof

\subsection{Proof of Lemma~\ref{lemma:GLM_event_T_adaptive}}
\proof{}
By the definitions of the negative log-likelihood $\ell_t(\beta)$ and its expectation $\Bar{\ell}_t(\beta)$, we can rewrite the empirical process $v_t(\beta)$ as
\begin{align*}
    v_t(\beta) &= \ell_t(\beta) - \Bar{\ell}_t(\beta)\\
        &=  -\frac{1}{t}\sum_{\tau=1}^t \left[ Y_\tau X_\tau^\top \beta - m(X_\tau^\top \beta) \right] +  \frac{1}{t}\sum_{\tau=1}^t \left[ \mu(X_\tau^\top \beta^*) X_\tau^\top \beta - m(X_\tau^\top \beta) \right]\\
        &=  -\frac{1}{t}\sum_{\tau=1}^t \left[ Y_\tau X_\tau^\top \beta - \mu(X_\tau^\top \beta^*) X_\tau^\top \beta \right]\\
        &=-\frac{1}{t} \sum_{\tau=1}^t  \epsilon_{\tau} X_{\tau}^\top \beta
\end{align*}
where the last equality uses the definition of $\epsilon_{\tau}$.
Then, the empirical process deviation is
\begin{align*}
    v_t(\hat{\beta}_t) - v_n(\beta^*) = -\frac{1}{t} \sum_{\tau=1}^t \epsilon_{\tau} X_{\tau}^\top (\hat{\beta}_t - \beta^*).
\end{align*}
Applying Hölder's inequality, we have
\begin{align*}
    | v_t(\hat{\beta}_t) - v_t(\beta^*) | &\leq \frac{1}{t} \left\|  \sum_{\tau=1}^t \epsilon_{\tau} X_{\tau} \right\|_{\infty} \|\hat{\beta}_t - \beta^*\|_1 .
\end{align*}
Then controlling the empirical process reduces to controlling $\frac{1}{t} \left\|  \sum_{\tau=1}^t  \epsilon_{\tau} X_{\tau} \right\|_{\infty}$. 
Then, using the union bound, it follows that
\begin{align*}
    \mathbb{P}\left(\frac{1}{t} \left\|  \sum_{\tau=1}^t  \epsilon_{\tau} X_{\tau} \right\|_{\infty} \leq \lambda \right) 
    &= 1 - \mathbb{P}\left(\frac{1}{t} \left\|  \sum_{\tau=1}^t  \epsilon_{\tau} X_{\tau} \right\|_{\infty} > \lambda \right)\\
    &\geq 1 - \sum_{j=1}^d \mathbb{P}\left(\frac{1}{t} \left|   \sum_{\tau=1}^t \epsilon_{\tau} X_{\tau}^{(j)} \right| > \lambda \right)
\end{align*}
where $X_{\tau}^{(j)}$ is the $j$-th element of $X_{\tau}$.
For each $j \in [d]$, and $\tau \in [t]$, we let $Z_{\tau}^{(j)} := \epsilon_{\tau} X_{\tau}^{(j)}$. 
Let $\tilde{\mathcal{F}}_{t-1}$ denote the 
sigma-field that contains all observed information
prior to taking an action in round $t$, i.e., $\tilde{\mathcal{F}}_{t-1}$ is generated by random variables of previously chosen actions $\{a_1, ..., a_{t-1}\}$, their features $\{X_1, ..., X_{t-1}\}$, the corresponding rewards $\{Y_1, ..., Y_{t-1}\}$
and the set of feature vectors $\mathcal{X}_t = \{X_{t,1}, ..., X_{t,K}\}$ in round~$t$. 

Then, each $\{Z^{(j)}_{\tau}\}_{\tau = 1}^t$ for $j \in [d]$ is a martingale difference sequence adapted to the filtration $\tilde{\mathcal{F}}_1 \subset ... \subset \tilde{\mathcal{F}}_\tau$ since $\mathbb{E}[\epsilon_\tau X_{\tau}^{(j)} | \tilde{\mathcal{F}}_{\tau-1}] = X_{\tau}^{(j)}\mathbb{E}[\epsilon_\tau  | \tilde{\mathcal{F}}_{\tau-1}] = 0$ for each $j$. 
Note that each $X_\tau^{(j)}$ is a bounded random variable with $| X_\tau^{(j)} | \leq \| X_\tau \|_\infty \leq \| X_\tau \|_2 \leq x_{\max}$. Then from the fact that  $\epsilon_\tau$ is $\sigma^2$-sub-Gaussian, it follows that $Z_{\tau}^{(j)}$ is also $\sigma^2$-sub-Gaussian. That is,
\begin{align*}
    \mathbb{E}\left[\exp(\alpha Z^{(j)}_{\tau}) \mid \tilde{\mathcal{F}}_{\tau-1}\right] &= \mathbb{E}\left[\exp\left\{\left(\alpha X_{\tau}^{(j)}\right) \epsilon_{\tau} \right\} \mid \tilde{\mathcal{F}}_{\tau-1} \right]\\
    &\leq  \mathbb{E}\left[\exp\!\left(\alpha x_{\max} \epsilon_{\tau} \right) \mid \tilde{\mathcal{F}}_{\tau-1} \right]\\
    &\leq \exp\!\left( \frac{\alpha^2 x_{\max}^2 \sigma^2}{2} \right)
\end{align*}
for any $\alpha \in \mathbb{R}$.
Then, using the concentration result in Lemma~\ref{lemma:bernstein}, we have
\begin{align*}
    \mathbb{P}\left( \left|   \sum_{\tau=1}^t  \epsilon_{\tau} X_{\tau}^{(j)} \right| > t\lambda \right)
    \leq 2 \exp\!\left( -\frac{t^2 \lambda^2}{2 t \sigma^2 x_{\max}^2 } \right) \leq 2 \exp\!\left( -\frac{t \lambda^2}{2 \sigma^2 x_{\max}^2  } \right) \,.
\end{align*}
So, with $\lambda = \sigma x_{\max} \sqrt{\frac{2[\log(2/\delta) + \log d]}{t}}$, we have
\begin{align*}
    \mathbb{P}\left(\frac{1}{t} \left\|  \sum_{\tau=1}^t  \epsilon_{\tau} X_{\tau} \right\|_{\infty} \leq \lambda \right) 
    \geq 1 - 2d\exp\!\left(\log\frac{\delta}{2} - \log d\right) 
    = 1 - \delta \,.
\end{align*}
\endproof

\newpage

\begin{lemma}\label{lemma:margin_cond}
The excess risk is lower-bounded by 
\begin{equation*}
    \mathcal{E}(\hat{\beta}_t) \geq \frac{\kappa_0}{2} (\hat{\beta}_t - \beta^*)^\top \hat{\Sigma}(\hat{\beta}_t - \beta^*) \,.
\end{equation*}
\end{lemma}

\proof{}
By the definition of the excess risk $\mathcal{E}(\beta)$, we have
\begin{align*}
    \mathcal{E}(\beta) 
    &= \Bar{\ell}_t(\beta) - \Bar{\ell}_t(\beta^*)\\
    &=  -\frac{1}{t}\sum_{\tau=1}^t \left[ \mu(X_\tau^\top \beta^*) X_\tau^\top \beta - m(X_\tau^\top \beta) \right]
    +  \frac{1}{t}\sum_{\tau=1}^t \left[ \mu(X_\tau^\top \beta^*) X_\tau^\top \beta^* - m(X_\tau^\top \beta^*) \right] \,.
\end{align*}
Since $\dot{m}(\cdot) = \mu(\cdot)$, we have $\nabla_\beta \Bar{\ell}_t(\beta^*) = \vec{0}_d$. Hence, the gradient of the  excess risk $\nabla_\beta  \mathcal{E}(\beta)$ and the Hessian are given as
\begin{align*}
    \nabla_\beta  \mathcal{E}(\beta) &=  -\frac{1}{t}\sum_{\tau=1}^t \left[ \mu(X_\tau^\top \beta^*) X_\tau - \mu(X_\tau^\top \beta) X_\tau \right] , \\
    H_{\mathcal{E}}(\beta) &:= \nabla^2_\beta  \mathcal{E}(\beta) = \frac{1}{t}\sum_{\tau=1}^t  \dot{\mu}(X_\tau^\top \beta) X_\tau X_\tau^\top \,.
\end{align*}
Using the Taylor expansion, with $\Bar{\beta} = c\beta^* + (1-c)\hat{\beta}$ for some $c \in (0,1)$
\begin{align}\label{eq:risk_taylor_exp}
    \mathcal{E}(\hat{\beta}_t) =  \mathcal{E}(\beta^*) + \nabla_\beta \mathcal{E}(\beta^*)^\top (\hat{\beta}_t - \beta^*) + \frac{1}{2}(\hat{\beta}_t - \beta^*)^\top H_{\mathcal{E}}(\Bar{\beta})(\hat{\beta}_t - \beta^*) \,.
\end{align}
Note that by the definition of $\beta^*$, we have $\mathcal{E}(\beta^*) = 0$ and $\nabla_\beta \mathcal{E}(\beta^*) = \nabla_\beta \ell(\beta^*) = \vec{0}_d$. Hence, combining with the definition of the Hessian, we have
\begin{align*}
    \mathcal{E}(\hat{\beta}_t) &= \frac{1}{2} (\hat{\beta}_t - \beta^*)^\top \left[ \frac{1}{t}\sum_{\tau=1}^t  \dot{\mu}(X_\tau^\top \Bar{\beta}) X_\tau X_\tau^\top \right](\hat{\beta}_t - \beta^*)\\
    & \geq \frac{\kappa_0}{2 } (\hat{\beta}_t - \beta^*)^\top \hat{\Sigma}(\hat{\beta}_t - \beta^*)
\end{align*}
 where the last inequality is from Assumption~\ref{assum:link_func_bounds} and $\hat{\Sigma} = \frac{1}{t} \sum_{\tau=1}^t   X_\tau X_\tau^\top$.
\endproof

\subsection{Proof of Lemma~\ref{lemma:cov_matrix_lowerbound_asymmetry}}
\label{app:lemma:cov_matrix_lowerbound_asymmetry}
\proof{}
Consider $\mathcal{X} = \{X_1, X_2\}$. Let the joint density function of $x_1, x_2$ as $p_\mathcal{X} (x_1, x_2)$. 
Then we have
\begin{align*}
    \mathbb{E}[\mathbf{X}^\top \mathbf{X}] 
    &= \int (x_1 x_1^\top + x_2 x_2^\top) p_\mathcal{X}(x_1, x_2)   d x_1, x_2\\
    &= \int  x_1 x_1^\top \left[\mathbbm{1}\left\{(x_1 - x_2)^\top \beta \geq 0 \right\} + \mathbbm{1}\left\{(x_1 - x_2)^\top \beta \leq 0 \right\}\right] p_\mathcal{X}(x_1,x_2)   d x_1, x_2\\
        &\quad + \int  x_2 x_2^\top \left[\mathbbm{1}\left\{(x_1 - x_2)^\top \beta \geq 0 \right\} + \mathbbm{1}\left\{(x_1 - x_2)^\top \beta \leq 0 \right\}\right] p_\mathcal{X}(x_1,x_2)   d x_1, x_2
\end{align*}
Let's first look at the first integral.
\begin{align*}
    &\int  x_1 x_1^\top \left[\mathbbm{1}\left\{(x_1 - x_2)^\top \beta \geq 0 \right\} + \mathbbm{1}\left\{(x_1 - x_2)^\top \beta \leq 0 \right\}\right] p_\mathcal{X}(x_1,x_2)   d x_1, x_2\\
    &= \int  x_1 x_1^\top \left[\mathbbm{1}\left\{(x_1 - x_2)^\top \beta \geq 0 \right\}p_\mathcal{X}(x_1,x_2) + \mathbbm{1}\left\{-(x_1 - x_2)^\top \beta \geq 0 \right\}p_\mathcal{X}(x_1,x_2) \right]    d x_1, x_2\\
    &\preccurlyeq \int  x_1 x_1^\top \mathbbm{1}\left\{(x_1 - x_2)^\top \beta \geq 0 \right\}p_\mathcal{X}(x_1,x_2)     d x_1, x_2 \\
    &\quad + \nu \int  x_1 x_1^\top  \mathbbm{1}\left\{-(x_1 - x_2)^\top \beta \geq 0 \right\}p_\mathcal{X}(-x_1,-x_2)     d x_1, x_2\\
    &= \int  x_1 x_1^\top \mathbbm{1}\left\{(x_1 - x_2)^\top \beta \geq 0 \right\}p_\mathcal{X}(x_1,x_2)     d x_1, x_2 \\
    &\quad + \nu \int  x_1 x_1^\top  \mathbbm{1}\left\{(x_1 - x_2)^\top \beta \geq 0 \right\}p_\mathcal{X}(x_1,x_2)     d x_1, x_2\\
    &= (1 + \nu)\int  x_1 x_1^\top \mathbbm{1}\left\{(x_1 - x_2)^\top \beta \geq 0 \right\}p_\mathcal{X}(x_1,x_2)     d x_1, x_2 \\
    &= (1+\nu)\mathbb{E}\left[X_1 X_1^\top \mathbbm{1}\{X_1 = \argmax_{X \in \mathcal{X}} X^\top \beta\} \right]
\end{align*}
where the inequality follows from Assumption~\ref{assum:rho}.
Likewise, we can show for the second integral that
\begin{align*}
    &\int  x_2 x_2^\top \left[\mathbbm{1}\left\{(x_1 - x_2)^\top \beta \geq 0 \right\} + \mathbbm{1}\left\{(x_1 - x_2)^\top \beta \leq 0 \right\}\right] p_\mathcal{X}(x_1,x_2)   d x_1, x_2\\
    &= (1+\nu)\mathbb{E}\left[X_2 X_2^\top \mathbbm{1}\{X_1 = \argmax_{X \in \mathcal{X}} X^\top \beta\} \right] \,.
\end{align*}
Hence,
\begin{align*}
     \mathbb{E}[\mathbf{X}^\top \mathbf{X}] = (1+\nu)\left( \mathbb{E}\left[X_1 X_1^\top \mathbbm{1}\{X_1 = \argmax_{X \in \mathcal{X}} X^\top \beta\} \right] + \mathbb{E}\left[X_2 X_2^\top \mathbbm{1}\{X_2 = \argmax_{X \in \mathcal{X}} X^\top \beta\} \right] \right) \,.
\end{align*}
Therefore, with the fact that $\nu \geq 1$, we have
\begin{align*}
\sum_{i=1}^2 \mathbb{E}\left[X_i X_i^\top \mathbbm{1}\{X_i = \argmax_{X \in \mathcal{X}} X^\top \beta\}\right]  \succcurlyeq  \frac{2}{1+\nu} \cdot \frac{1}{2}\mathbb{E}[\mathbf{X}^\top \mathbf{X}] \succcurlyeq  \nu^{-1} \Sigma\,.
\end{align*}
\endproof

\subsection{Bernstein-type Inequality for Adapted Samples}
In this section, we derive a Bernstein-type inequality for adapted samples which is shown in Lemma \ref{lemma:maximal_bernstein_ineq}. We first define the following function of a random variable $X_t$ which is used throughout this section.

\begin{definition}
For all $i,j$ with $1 \leq i \leq j \leq d$, we define $\gamma^{ij}_t(X_t)$ to be a real-value function which take random variable $X_t \in \mathbb{R}^d$ as input:
\begin{align}\label{eq:gamma_definition}
    \gamma^{ij}_t(X_t) := \frac{1}{2 x_{\max}^2}\left( X^{(i)}_t X^{(j)}_t - \mathbb{E}[X^{(i)}_t X^{(j)}_t \mid \mathcal{F}_{t-1}] \right)
\end{align}
where $X^{(i)}_t$ is the $i$-th element of $X_t$.
\end{definition}
It is easy to see that $\mathbb{E}\big[ \gamma^{ij}_t(X_t)  \mid \mathcal{F}_{t-1} \big] = 0$ and $\mathbb{E}\big[| \gamma^{ij}_t(X_t) |^m \mid \mathcal{F}_{t-1}\big] \leq 1$ for all integer $m \geq 2$. 
While we introduce this specific function $\gamma^{ij}_t(X_t)$ in order to connect to the matrix concentration $\|  \Sigma_\tau - \hat{\Sigma}_\tau \|_{\infty}$, Lemma~\ref{lemma:mean_expo_bound} and Lemma~\ref{lemma:maximal_bernstein_ineq} can be applied to any function $\gamma^{ij}_t(X_t)$ that satisfies the zero mean and the bounded $m$-th moment conditions.

\begin{lemma}[\citet{buhlmann2011statistics}, Lemma~14.1]\label{lemma:expo_mg_bound}
Let $Z_t \in \mathbb{R}$ be a random variable with $\mathbb{E}[Z_t \mid \mathcal{F}_{t-1}] = 0$. Then it holds that
\begin{align*}
    \log \mathbb{E} \left[e^{Z_t}  \mid \mathcal{F}_{t-1}\right] \leq \mathbb{E}\left[ e^{|Z_t|} \mid \mathcal{F}_{t-1} \right]-1-\mathbb{E}\left[ |Z| \mid \mathcal{F}_{t-1} \right] \,.
\end{align*}
\end{lemma}

\proof{}
The proof follows directly from the proof of Lemma~14.1 in \citet{buhlmann2011statistics}, applying their result to a conditional expectation. For any $c > 0$,
\begin{align*}
    \exp(Z_t - c) - 1 &\leq \frac{\exp(Z_t)}{1+c} - 1 \\
    &= \frac{e^{Z_t} -1 - Z_t + Z_t - c}{1+c} \\
    &\leq \frac{e^{|Z_t|} -1 - |Z_t| + Z_t - c}{1+c} \,.
\end{align*}
Let $c = \mathbb{E}\left[ e^{|Z_t|} \mid \mathcal{F}_{t-1}\right] -1 - \mathbb{E}\left[ |Z| \mid \mathcal{F}_{t-1} \right]$. Hence, since $\mathbb{E}[Z_t \mid \mathcal{F}_{t-1}] = 0$,
\begin{align*}
    \mathbb{E}\left[\exp(Z_t - c) \mid \mathcal{F}_{t-1} \right] - 1  
    &\leq \frac{\mathbb{E}\left[ e^{|Z_t|}  \mid \mathcal{F}_{t-1} \right] -1 - \mathbb{E}\left[|Z_t|  \mid \mathcal{F}_{t-1} \right]  - c}{1+c} = 0 \,.
\end{align*}
\endproof

\begin{lemma}\label{lemma:mean_expo_bound}
Suppose $\mathbb{E}\big[ \gamma^{ij}_t(X_t)  \mid \mathcal{F}_{t-1} \big] = 0$ and $\mathbb{E}\big[| \gamma^{ij}_t(X_t) |^m \mid \mathcal{F}_{t-1}\big] \leq m!$ for all integer $m \geq 2$, all $t \geq 1$ and all $1 \leq i \leq j \leq d$. Then, for $L > 1$ we have
\begin{align*}
    \mathbb{E}\left[ \exp\!\left( \frac{1}{L}\sum_{t=1}^\tau \gamma^{ij}_t(X_t) \right) \right] \leq \exp\!\left( \frac{\tau}{L(L-1)} \right) .
\end{align*}
\end{lemma}

\proof{}
\begin{align*}
    \mathbb{E}\left[ \exp\!\left( \frac{1}{L}\sum_{t=1}^\tau \gamma^{ij}_t(X_t) \right) \right] 
    &= \mathbb{E}\left[ \mathbb{E}\bigg[ \exp\!\bigg( \frac{1}{L}\sum_{t=1}^\tau \gamma^{ij}_t(X_t) \bigg) \mid \mathcal{F}_{\tau-1}\bigg] \right]\\
    &= \mathbb{E}\left[ \mathbb{E}\bigg[ \exp\!\bigg( \frac{\gamma^{ij}_\tau (X_\tau)}{L} \bigg) \mid \mathcal{F}_{\tau-1}\bigg] \exp\!\bigg( \frac{1}{L}\sum_{t=1}^{\tau-1} \gamma^{ij}_t(X_t) \bigg) \right]\\
    &\leq e^{\frac{1}{L(L-1)}} \mathbb{E}\left[ \exp\!\bigg( \frac{1}{L}\sum_{t=1}^{\tau-1} \gamma^{ij}_t(X_t) \bigg) \right]
\end{align*}
where the inequality is from Lemma~\ref{lemma:expo_mg_bound} and noting that
\begin{align*}
    \log \mathbb{E}\bigg[ \exp\!\bigg(  \frac{\gamma^{ij}_\tau (X_\tau)}{L}  \bigg) \mid \mathcal{F}_{\tau-1}\bigg]
    &\leq \mathbb{E}\left[ e^{ \left|\gamma^{ij}_\tau (X_\tau) \right|/\tau }  - 1 -  \frac{\left|\gamma^{ij}_\tau (X_\tau) \right|}{L}  \mid \mathcal{F}_{\tau-1} \right]\\
    &= \mathbb{E}\left[ \sum_{m=2}^\infty \frac{ \left|\gamma^{ij}_\tau (X_\tau) \right|^m  }{L^m m!} \mid \mathcal{F}_{\tau-1} \right]\\
    &= \sum_{m=2}^\infty \frac{ \mathbb{E}\left[ \left|\gamma^{ij}_\tau (X_\tau) \right|^m \mid \mathcal{F}_{\tau-1} \right] }{L^m m!} \\
    &\leq \frac{1}{L(L-1)} \,.
\end{align*}
Then, repeatedly applying this to the rest of the sum $\frac{1}{L} \sum_{t=1}^{\tau-1} \gamma^{ij}_t(X_t)$, we have
\begin{align*}
    \mathbb{E}\left[ \exp\!\left( \frac{1}{L}\sum_{t=1}^\tau \gamma^{ij}_t(X_t) \right) \right] \leq \exp\!\left( \frac{\tau}{L(L-1)} \right) \,.
\end{align*}
\endproof

\begin{lemma}[Bernstein-type inequality for adapted samples]\label{lemma:maximal_bernstein_ineq}
Suppose $\mathbb{E}\big[ \gamma^{ij}_t(X_t)  \mid \mathcal{F}_{t-1} \big] = 0$ and $\mathbb{E}\big[| \gamma^{ij}_t(X_t) |^m \mid \mathcal{F}_{t-1}\big] \leq m!$ for all integer $m \geq 2$, all $t \geq 1$ and all $1 \leq i \leq j \leq d$. Then for all $w > 0$, we have
\begin{align*}
\mathbb{P}\left( \max_{1\leq i\leq j \leq d} \left|\frac{1}{\tau}\sum_{t=1}^\tau \gamma^{ij}_t(X_t) \right| \geq w + \sqrt{2w} + \sqrt{\frac{4\log(2d^2)}{\tau}} + \frac{2\log(2d^2)}{\tau} \right) \leq \exp\left(-\frac{\tau w}{2} \right) \,.
\end{align*}
\end{lemma}

\proof{}
Using the Chernoff bound and Lemma~\ref{lemma:mean_expo_bound}, for any $L > 1$ we have
\begin{align*}
    \mathbb{P}\left(  \sum_{t=1}^\tau \gamma^{ij}_t(X_t) \geq a \right)
    &= \mathbb{P}\left(  \exp\!\bigg( \frac{1}{L} \sum_{t=1}^\tau \gamma^{ij}_t(X_t) \bigg) \geq \exp\!\left(\frac{a}{L}\right) \right)\\
    &\leq \frac{\mathbb{E}\left[ \exp\!\left(  \frac{1}{L} \sum_{t=1}^\tau \gamma^{ij}_t(X_t) \right) \right] }{\exp\!\left( \frac{a}{L} \right)}\\
    &\leq \exp\!\left(-\frac{a}{L}\right) \exp\!\left( \frac{\tau}{L(L-1)} \right)\\
    &= \exp\!\left(-\frac{a}{L} + \frac{\tau}{L(L-1)} \right) \,.
\end{align*}
Here, $L = \frac{\tau + a + \sqrt{\tau^2 + \tau a}}{a}$ minimizes the right hand side above for $L > 1$. Therefore,
\begin{align*}
    \mathbb{P}\left(  \sum_{t=1}^\tau \gamma^{ij}_t(X_t) \geq a \right)
    &\leq \exp\left\{-\frac{a^2}{\tau + a + \sqrt{\tau^2 + \tau a}} + \frac{\tau a^2}{(\tau + a + \sqrt{\tau^2 + \tau a})(\tau + \sqrt{\tau^2 + \tau a})} \right\}\\
    &= \exp\left\{-\left( \frac{\sqrt{1 + a/\tau}}{1 + \sqrt{1 + a/\tau}}\right) \frac{a^2}{\tau + a + \sqrt{\tau^2 + \tau a}}  \right\}\\
    &\leq \exp\left\{- \frac{a^2}{2\left(\tau + a + \sqrt{\tau^2 + \tau a}\right)}  \right\}\\
    &\leq \exp\left\{- \frac{a^2}{2\left(\tau + a + \sqrt{\tau^2 + 2\tau a}\right)}  \right\} \,.
\end{align*}
Choosing $a = \tau \left(w + \sqrt{2 w} \right)$ gives
\begin{align}\label{eq:bernstein_ineq}
    \mathbb{P}\left(  \frac{1}{\tau}\sum_{t=1}^\tau \gamma^{ij}_t(X_t) \geq w + \sqrt{2w} \right) \leq \exp\!\left( -\frac{\tau w}{2} \right) \,.
\end{align}
Then for the maximal inequality, we first apply the union bound to \eqref{eq:bernstein_ineq}.
\begin{align*}
    \mathbb{P}\left( \max_{1\leq i\leq j \leq d} \frac{1}{\tau} \left|\sum_{t=1}^\tau \gamma^{ij}_t(X_t) \right| \geq w + \sqrt{2w} \right) 
    &\leq  \sum_{1\leq i\leq j \leq d}  2 \mathbb{P}\left(  \frac{1}{\tau}\sum_{t=1}^\tau \gamma^{ij}_t(X_t) \geq w + \sqrt{2w} \right)\\
    &\leq 2 d^2 \exp\!\left( -\frac{\tau w}{2} \right)\\
    &= \exp\!\left( -\frac{\tau w}{2} + \log(2d^2)\right) \,.
\end{align*}
Then, 
\begin{align*}
    &\mathbb{P}\left( \max_{1\leq i\leq j \leq d} \frac{1}{\tau}\left|\sum_{t=1}^\tau \gamma^{ij}_t(X_t) \right| \geq w + \sqrt{2w} + \sqrt{\frac{4\log(2d^2)}{\tau}} + \frac{2\log(2d^2)}{\tau} \right)\\
    &\leq  \mathbb{P}\left( \max_{1\leq i\leq j \leq d} \frac{1}{\tau} \left|\sum_{t=1}^\tau \gamma^{ij}_t(X_t) \right| \geq \left(w + \frac{2\log(2d^2)}{\tau} \right) +  \sqrt{2\left(w + \frac{2\log(2d^2)}{\tau} \right) }   \right)\\
    &= \mathbb{P}\left( \max_{1\leq i\leq j \leq d} \frac{1}{\tau}\left|\sum_{t=1}^\tau \gamma^{ij}_t(X_t) \right| \geq w' + \sqrt{2w'} \right)\\
    &\leq \exp\!\left( -\frac{\tau w'}{2} + \log(2d^2)\right)\\
    &= \exp\!\left( -\frac{\tau w}{2} \right)
\end{align*}
where $w' = w + \frac{2\log(2d^2)}{\tau}$.
\endproof

\subsection{Proof of Lemma~\ref{lemma:Sigma_hat_concentration}}

\proof{}
Notice the difference between the unconditional theoretical Gram matrix $\Sigma$ and its adapted version $\mathbb{E}[X_t X_t^\top | \mathcal{F}_{t-1}]$ which is a conditional covariance matrix conditioned on the history $\mathcal{F}_{t-1}$.
Recall that from Algorithm \ref{algo:SA_Lasso_bandit}, in each round $t$ we choose $X_t$ given the history $\mathcal{F}_{t-1}$. More precisely, we compute $\beta_t$ based on $\mathcal{F}_{t-1}$ and choose $X_t$ which maximizes the product $X_t^\top \hat{\beta}_t$, i.e.,  $\argmax_{X \in \mathcal{X}_t} X^\top \hat{\beta}_t$ where $\mathcal{X}_t = \{X_{t,1}, X_{t,2} \}$. Hence, we can write  $\mathbb{E}[X_t X_t^\top | \mathcal{F}_{t-1}]$ as the following:
\begin{align*}
    \mathbb{E}[X_t X_t^\top | \mathcal{F}_{t-1}] 
    = \sum_{i = 1}^2\mathbb{E}_{\mathcal{X}_t}\Big[X_{t, i} X_{t, i}^\top  \mathbbm{1}\{X_{t i} = \argmax_{X \in \mathcal{X}_t} X^\top \hat{\beta}_t\} \mid \hat{\beta}_t \Big] \,.
\end{align*}
From Lemma~\ref{lemma:cov_matrix_lowerbound_asymmetry}, it follows that
\begin{align*}
    \mathbb{E}[X_t X_t^\top | \mathcal{F}_{t-1}] \succcurlyeq \nu^{-1} \Sigma \,.
\end{align*}
Now, taking an average over $t$ gives,
\begin{align*}
    \Sigma_\tau =\frac{1}{\tau}\sum_{t=1}^\tau \mathbb{E}[X_t X_t^\top | \mathcal{F}_{t-1}] \succcurlyeq \nu^{-1} \Sigma \,.
\end{align*}
Then, we define $\tilde{\beta}$ corresponding to compatibility constant $\phi^2(\Sigma_\tau,S_0)$, that is,
\begin{align*}
    \tilde{\beta} := \argmin_{\beta} \left\{ \frac{ \beta^\top \Sigma_\tau \beta}{\|\beta_{S_0}\|^2_1} : \|\beta_{S^c_0}\|_1 \leq 3\|\beta_{S_0}\|_1 \neq 0 \right\} \,.
\end{align*}
Therefore, it follows that
\begin{align}\label{eq:comp_constant_Simga_t}
      \frac{\tilde{\beta}^\top \Sigma_\tau \tilde{\beta}}{\|\tilde{\beta}_{S_0}\|^2_1} 
      \geq \frac{\tilde{\beta}^\top \Sigma \tilde{\beta} }{\nu  \|\tilde{\beta}_{S_0}\|^2_1 } 
      \geq \frac{ \phi^2_0}{\nu  }
\end{align}
where the second inequality is by the compatibility condition on $\Sigma$. Thus, $\Sigma_\tau$ satisfies the compatibility condition with compatibility constant $\phi^2(\Sigma_\tau,S_0) = \frac{ \phi^2_0}{\nu }$.

\noindent Now, noting that $\frac{1}{2 x_{\max}^2} \|  \Sigma_\tau - \hat{\Sigma}_\tau \|_{\infty} = \max_{1\leq i\leq j \leq d} \frac{1}{\tau} \left|\sum_{t=1}^\tau \gamma^{ij}_t(X_t) \right|$ for $\gamma^{ij}_t(\cdot)$ defined in \eqref{eq:gamma_definition}, we can use a Bernstein-type inequality
for adapted samples in Lemma~\ref{lemma:maximal_bernstein_ineq} to get
\begin{align*}
    \mathbb{P}\left( \frac{\| \Sigma_\tau - \hat{\Sigma}_\tau \|_{\infty}}{2 x_{\max}^2} \geq w + \sqrt{2w} + \sqrt{\frac{4 \log(2d^2)}{\tau}} + \frac{2\log(2d^2)}{\tau} \right) \leq \exp\!\left(-\frac{\tau w}{2}\right) \,.
\end{align*}
For $\tau \geq \frac{2\log(2d^2)}{C_0(s_0)^2}$ where $C_0(s_0) = \min\!\Big(\frac{1}{2}, \frac{\phi_0^2}{256 s_0 \nu x_{\max}^2} \Big)$, letting $w = C_0(s_0)^2$ gives
\begin{align*}
     w + \sqrt{2w} + \sqrt{\frac{4 \log(2d^2)}{\tau}} + \frac{2\log(2d^2)}{\tau}
    &\leq 2 \left( C_0(s_0)^2 + \sqrt{2} C_0(s_0)\right)\\
    &\leq 4 C_0(s_0)\\
    &\leq \frac{ \phi_0^2}{64 s_0 \nu x_{\max}^2}\\
    &= \frac{\phi^2(\Sigma_\tau,S_0)}{64 s_0 x_{\max}^2}\,.
\end{align*}
Hence,
\begin{align*}
    \mathbb{P}\left( \frac{\| \Sigma_\tau - \hat{\Sigma}_\tau \|_{\infty}}{2 x_{\max}^2 } \geq  \frac{\phi^2(\Sigma_\tau,S_0)}{64 s_0 x_{\max}^2} \right) 
    &\leq \mathbb{P}\left( \frac{\| \Sigma_\tau - \hat{\Sigma}_\tau \|_{\infty}}{2 x_{\max}^2} \geq w + \sqrt{2w} + \sqrt{\frac{4 \log(2d^2)}{\tau}} + \frac{2\log(2d^2)}{\tau} \right)\\
    &\leq \exp\!\left(- \frac{\tau w}{2}\right)\\
    &= \exp\!\left(- \frac{\tau C_0(s_0)^2}{2} \right) \,.
\end{align*}
\endproof

\begin{corollary}\label{cor:compatibility_sigma_hat}
For $t \geq \frac{2\log(2d^2)}{C_0(s_0)^2}$ where $C_0(s_0) = \min\!\Big(\frac{1}{2}, \frac{\phi_0^2}{256 s_0 \nu x_{\max}^2} \Big)$, 
the empirical Gram matrix $\hat{\Sigma}_t$ satisfies the compatibility condition with compatibility constant $\phi_t \geq \frac{\phi^2_0}{2 \nu } > 0$ with probability at least $1 - \exp\left\{- t C_0(s_0)^2/2 \right\}$.
\end{corollary}

\proof{}
We can use Corollary~\ref{cor:gram_mat_infinity_norm_compatibility} (\citet{buhlmann2011statistics}, Corollary~6.8) to show that the empirical Gram matrix $\hat{\Sigma}_\tau$ satisfies the compatibility condition as long as $\Sigma_\tau$ satisfies the compatibility condition. 
From \eqref{eq:comp_constant_Simga_t}, we know $\Sigma_\tau$ satisfies the compatibility condition with compatibility constant $\frac{ \phi^2_0}{\nu  }$. 
Then, combining Lemma~\ref{lemma:Sigma_hat_concentration} and Corollary~\ref{cor:gram_mat_infinity_norm_compatibility}, it follows that given $\| \Sigma_t - \hat{\Sigma}_t \|_{\infty} \leq  \frac{ \phi^2_0}{32  s_0\nu }$
for $t \geq \lceil T_0 \rceil$, we have
\begin{align*}
\phi^2(\hat{\Sigma}_t, S_0) \geq \frac{\phi^2(\Sigma_t, S_0)}{2} \geq \frac{\phi^2_0}{2 \nu } > 0\,.
\end{align*}
That is, $\hat{\Sigma}_\tau$ satisfies the compatibility condition with compatibility constant which is at least $\frac{ \phi^2_0}{ 2 \nu  } > 0$.
\endproof

\section{Proof of Theorem~\ref{thm:GLM-Lasso-regret-bound_CC}}
\label{app:Theorem1main}
\proof{}
First, let $T_0 := \frac{2\log(2d^2)}{C_0(s_0)^2}$ where $C_0(s_0) = \min\!\Big(\frac{1}{2}, \frac{\phi_0^2}{256 s_0 \nu x_{\max}^2} \Big)$.
Also, we define the high probability event $\mathcal{E}_t$:
\begin{align*}
 \mathcal{E}_t := 
 \left\{ \| \Sigma_t - \hat{\Sigma}_t \|_{\infty} \geq  \frac{ \phi^2_0}{32 s_0\nu }  \right\} \,.
\end{align*}
Hence, on this event
$\mathcal{E}_t$, if $t \geq T_0$, then from Corollary~\ref{cor:compatibility_sigma_hat} we have $\phi^2_t \geq \frac{
  \phi^2_0}{ 2 \nu }$, i.e., the compatibility condition
holds in round $t$. Slightly overloading the subscript for brevity, let
$X_t := X_{t,a_t}$ be a feature of the arm chosen in round $t$ and
$X_{a_t^*} := X_{t,a^*_t}$ be the feature of the optimal arm in round $t$.
First, we look at the (non-expected) immediate regret $\text{Reg}(t)$ with $\mathcal{R}(t) = \mathbb{E}[\text{Reg}(t)]$ in round $t$. Notice that by Assumptions~\ref{assum:feature_param_bounds}~and~\ref{assum:link_func_bounds} and by the mean value theorem, $\text{Reg}(t)$ is bounded by
\begin{align*}
    \text{Reg}(t) \leq \kappa_1 \big( X_{a_t^*}^\top \beta^* - X_t^\top \beta^* \big) \leq \kappa_1 \| X_{a_t^*} - X_t \|_2 \| \beta^*\|_2 \leq 2\kappa_1 x_{\max} b
\end{align*}
Then we can decompose the immediate regret as follows.
\begin{align*}
    \text{Reg}(t) 
    &= \text{Reg}(t)\mathbbm{1}(t \leq T_0) + \text{Reg}(t)\mathbbm{1}(t > T_0, \mathcal{E}_t) + \text{Reg}(t)\mathbbm{1}(t > T_0, \mathcal{E}_t^c)\\
    &\leq 2\kappa_1 x_{\max} b \mathbbm{1}(t \leq T_0) + \text{Reg}(t)\mathbbm{1}(t > T_0, \mathcal{E}_t) + 2\kappa_1 x_{\max} b \mathbbm{1}(t > T_0, \mathcal{E}_t^c)\\
    &= 2\kappa_1 x_{\max} b \mathbbm{1}(t \leq T_0) + \text{Reg}(t)\mathbbm{1}\left( \mu(X_t^\top \hat{\beta}_t) \geq \mu(X_{a_t^*}^\top  \hat{\beta}_t),  t > T_0, \mathcal{E}_t \right)\\ 
    &\quad + 2\kappa_1 x_{\max} b \mathbbm{1}(t > T_0, \mathcal{E}_t^c)
\end{align*}
where the last equality follows from the optimality of $X_t$ with respect to parameter $\hat{\beta}_t$, i.e., $X_t = \argmax_{X \in \mathcal{X}_t} \mu(X^\top \hat{\beta}_t)$. 
For the second term, we have
\begin{align*}
    \mathbb{P}\left( \mu(X_t^\top \hat{\beta}_t) \geq \mu(X_{a_t^*}^\top  \hat{\beta}_t)\right)
    &= \mathbb{P}\left( \mu(X_t^\top \hat{\beta}_t) - \mu(X_{a_t^*}^\top  \hat{\beta}_t) + \text{Reg}(t)\geq \text{Reg}(t) \right)\\
    &= \mathbb{P}\left( (\mu(X_t^\top \hat{\beta}_t) - \mu(X_t^\top \beta^*)) - (\mu(X_{a_t^*}^\top  \hat{\beta}_t) - \mu(X_{a^*_t}^\top \beta^*))\geq \text{Reg}(t) \right)\\
    &\leq \mathbb{P}\left( | \mu(X_t^\top \hat{\beta}_t) - \mu(X_t^\top \beta^*)| + |\mu(X_{a_t^*}^\top  \hat{\beta}_t) - \mu(X_{a^*_t}^\top \beta^*)| \geq \text{Reg}(t) \right)\\
    &\leq \mathbb{P}\left( \kappa_1 \| \hat{\beta}_t - \beta^*\|_1 \| X_t\|_{\infty} + \kappa_1 \| \hat{\beta}_t - \beta^*\|_1 \| X_{a_t^*}\|_{\infty} \geq \text{Reg}(t) \right)\\
    &\leq \mathbb{P}\left( 2 \kappa_1 \| \hat{\beta}_t - \beta^*\|_1\geq \text{Reg}(t) \right)
\end{align*}
where the last inequality is from the fact that each $X_{t,i}$ is bounded.
For an arbitrary constant $g_t > 0$, we continue with expected regret $\mathcal{R}(t) = \mathbb{E}[\text{Reg}(t)]$ for $t > T_0$.
\begin{align*}
    \mathcal{R}(t) &\leq 
     \mathbb{E}\left[\text{Reg}(t)\mathbbm{1}\left( 2 \kappa_1 \| \hat{\beta}_t - \beta^*\|_1 \geq \text{Reg}(t),  \mathcal{E}_t \right) \right] + 2\kappa_1 x_{\max} b\mathbb{P}(\mathcal{E}_t^c)\\
    &=  \mathbb{E}\left[\text{Reg}(t)\mathbbm{1}\left( 2 \kappa_1 \| \hat{\beta}_t - \beta^*\|_1\geq \text{Reg}(t),  \text{Reg}(t) \leq  \kappa_1 g_t, \mathcal{E}_t \right) \right]\\
    &\quad +  \mathbb{E}\left[\text{Reg}(t)\mathbbm{1}\left( 2 \kappa_1 \| \hat{\beta}_t - \beta^*\|_1\geq \text{Reg}(t), \text{Reg}(t) > \kappa_1 g_t, \mathcal{E}_t,  \right) \right] + 2\kappa_1 x_{\max} b \mathbb{P}(\mathcal{E}_t^c)\\
    &\leq   \kappa_1 g_t +  \kappa_1 \mathbb{P}\left( 2\| \hat{\beta}_t - \beta^*\|_1 \geq g_t,  \mathcal{E}_t\right) + 2\kappa_1 x_{\max} b \mathbb{P}(\mathcal{E}_t^c) \,.
\end{align*}
Summing over all rounds after the initial $T_0$ rounds, we have
\begin{align}\label{eq:regret_decompose}
   \sum_{t=\lceil T_0 \rceil}^T \mathcal{R}(t) &\leq
    \underbrace{\kappa_1 \sum_{\mathclap{t= \lceil T_0 \rceil}}^T g_t}_{(a)}
   + \underbrace{\kappa_1 \sum_{\mathclap{t= \lceil T_0 \rceil}}^T\mathbb{P}\left( 2\| \hat{\beta}_t - \beta^*\|_1 \geq g_t, \mathcal{E}_t\right)}_{(b)}
   + \underbrace{2\kappa_1 x_{\max} b \sum_{\mathclap{t= \lceil T_0 \rceil}}^T\mathbb{P}(\mathcal{E}_t^c)}_{(c)} \,.
\end{align}
We first bound the term $(b)$ in \eqref{eq:regret_decompose}.
We choose $g_t := \frac{2 s_0 \lambda_t}{\kappa_0\phi^2_t} = \frac{4 \sigma x_{\max} s_0}{\kappa_0\phi^2_t} \sqrt{  \frac{4\log t + 2\log d}{t} }$. Then using Lemma~\ref{lemma:oracle_ineq_GLM_L1}, we have
\begin{align*}
    \mathbb{P}\left( 2\| \hat{\beta}_t - \beta^*\|_1 \geq g_t,  \mathcal{E}_t\right) \leq \frac{2}{t^2} \,.
\end{align*}
for all $t \geq T_0$. Therefore, it follows that
\begin{align*}
    \sum_{t=\lceil T_0 \rceil}^T\mathbb{P}\left( 2\| \hat{\beta}_t - \beta^*\|_1 \geq g_t,   \mathcal{E}_t\right)
    \leq \sum_{t=\lceil T_0 \rceil}^T \frac{2}{t^2}
    \leq \sum_{t=1}^{\infty} \frac{2}{t^2}
    \leq \frac{\pi^2}{3} < 4 \,.
\end{align*}
For the term (a) in \eqref{eq:regret_decompose}, we have $\phi^2_t \geq \frac{ \phi^2_0}{ 2 \nu }$ provided that event $\mathcal{E}_t$ holds. Hence, we have
\begin{align*}
    \sum_{t=\lceil T_0 \rceil}^T g_t &= \sum_{t=\lceil T_0 \rceil}^T\frac{4 \sigma x_{\max} s_0 }{\kappa_0\phi^2_t} \sqrt{ \frac{ 4\log t + 2\log d}{t} } \\
    &\leq \sum_{t=\lceil T_0 \rceil}^T\frac{8 \nu \sigma x_{\max} s_0 }{\kappa_0\phi^2_0 } \sqrt{ \frac{ 4\log t + 2\log d}{t} } \\
    &\leq \frac{8 \nu \sigma x_{\max} s_0 \sqrt{  4\log T + 2\log d }}{\kappa_0\phi^2_0}\sum_{t= \lceil T_0 \rceil}^{T}\frac{1}{\sqrt{t}} \\
    &\leq \frac{8 \nu \sigma x_{\max} s_0 \sqrt{  4\log T + 2\log d }}{\kappa_0\phi^2_0}\sum_{t= 1}^{T}\frac{1}{\sqrt{t}} \\
    &\leq \frac{16 \nu \sigma x_{\max} s_0 \sqrt{  4\log T + 2\log d }}{\kappa_0\phi^2_0} \sqrt{T}
\end{align*}
where  the last inequality is from the fact that $\sum_{t= 1}^{T}\frac{1}{\sqrt{t}} \leq \int_{t= 0}^{T}\frac{1}{\sqrt{t}} = 2\sqrt{T}$.

\noindent Finally, for the term (c) in \eqref{eq:regret_decompose},
we have from Lemma~\ref{lemma:Sigma_hat_concentration}:
\begin{align*}
    \sum_{t= \lceil T_0 \rceil}^T\mathbb{P}(\mathcal{E}_t^c) &\leq \sum_{t=  \lceil T_0 \rceil}^T \mathbb{P}\left( \| \Sigma_t - \hat{\Sigma}_t \|_{\infty} \geq  \frac{ \phi^2_0}{32  s_0\nu }  \right)\\
    &\leq  \sum_{t= \lceil T_0 \rceil}^T \exp\left(- \frac{t C_0(s_0)^2}{2} \right)\\
    &\leq  \sum_{t= 1}^\infty \exp\left(- \frac{t C_0(s_0)^2}{2} \right) \\
    &\leq \frac{2}{C_0(s_0)^2} \,.
\end{align*}
\endproof

\section{Proof of Theorem~\ref{thm:GLM-Lasso-regret-bound_re}}

The proof follows similar arguments as the proof of Theorem~\ref{thm:GLM-Lasso-regret-bound_CC}. The key difference is that the RE condition involves $\ell_2$ norm and therefore the analysis requires the Lasso oracle inequality of the GLM in $\ell_2$ norm, which we provide as an extension of Lemma \ref{lemma:oracle_ineq_GLM_L1}.

\begin{corollary}\label{cor:oracle_ineq_GLM_L2}
Assume that the RE condition holds for $\hat{\Sigma}_t$ with active set $S_0$ and restricted eigenvalue $\phi_t$. For some $\delta \in (0,1)$, let the regularization parameter $\lambda_t$ be
\begin{equation*}
    \lambda_t := 2\sigma x_{\max} \sqrt{\frac{2[\log(2/\delta) + \log d]}{t}} \,.
\end{equation*}
Then with probability at least $1-\delta$, we have
\begin{align*}
        \| \hat{\beta}_t - \beta^* \|_2 \leq \frac{3 \sqrt{s_0} \lambda_t }{\kappa_0\phi^2_t} \,.
\end{align*}
\end{corollary}

\proof{}
Continuing from \eqref{eq:intermed_l1_active_bound} in Lemma~\ref{lemma:oracle_ineq_GLM_L1}, the RE condition can be applied to the vector $\hat{\beta} - \beta^*$ which gives
\begin{align}\label{eq:apply_compatibility}
     \| \hat{\beta} - \beta^* \|^2_2 \leq \frac{ (\hat{\beta} - \beta^*)^\top \hat{\Sigma}_t (\hat{\beta} - \beta^*)}{\phi^2_t} \,.
\end{align}
Again from \eqref{eq:intermed_l1_active_bound}, we can use the margin condition in Lemma~\ref{lemma:margin_cond}
\begin{align*}
       3\lambda_t\|\hat{\beta}_{S_0} - \beta^*_{S_0}\|_1 &\geq 2\mathcal{E}(\hat{\beta}_n) \notag\\
       &\geq \kappa_0 (\hat{\beta} - \beta^*)^\top \hat{\Sigma}_t (\hat{\beta} - \beta^*) \notag\\
       &\geq \kappa_0\phi^2_t \| \hat{\beta} - \beta^* \|^2_2
\end{align*}
where the last inequality is from \eqref{eq:apply_compatibility} applying the RE condition.
Then, it follows that
\begin{align*}
   \kappa_0\phi^2_t \| \hat{\beta} - \beta^* \|^2_2 &\leq  3 \lambda_t \|\hat{\beta}_{S_0} - \beta^*_{S_0}\|_1\\
    &\leq 3 \lambda_t \sqrt{s_0}\|\hat{\beta}_{S_0} - \beta^*_{S_0}\|_2 \\
    &\leq 3 \lambda_t \sqrt{s_0}\|\hat{\beta} - \beta^*\|_2 \,.
\end{align*}
Hence, dividing the both sides by $ \| \hat{\beta} - \beta^* \|_2$ and rearranging gives
\begin{align*}
     \| \hat{\beta} - \beta^* \|_2 \leq \frac{ 3 \sqrt{s_0} \lambda_t}{\kappa_0\phi^2_t} \,.
\end{align*}
This complete the proof.
\endproof

\subsection{Ensuring the RE Condition for the Empirical Gram Matrix}

To distinguish from the compatibility constant, we introduce the definition of a generic restricted eigenvalue of matrix $M$ over active set $S_0$.
\begin{definition}
The restricted eigenvalue of $M$ over $S_0$ is
\begin{align*}
    \phi^2_{\text{RE}}(M,S_0) := \min_\beta \left\{ \frac{ \beta^\top M \beta}{\|\beta\|^2_2} : \|\beta_{S_0^c}\|_1 \leq 3\|\beta_{S_0}\|_1 \neq 0 \right\}.
\end{align*}
\end{definition}

Note that Assumption~\ref{assum:re_condition} only provides the RE condition for the theoretical Gram matrix~$\Sigma$.
Then, we follow the same arguments as in the analysis under the compatibility condition to show  that $\phi^2_{\text{RE}}(\Sigma_t, S_0) \geq \frac{\phi^2_{\text{RE}}(\Sigma, S_0)}{\nu } > 0$, i.e., $\Sigma_t$ satisfies the RE condition.
Then using Lemma \ref{lemma:Sigma_hat_concentration},
we can show that $\hat{\Sigma}_t$ concentrates to $\Sigma_t$ with high probability. 
The following lemma (similar to Corollary~\ref{cor:gram_mat_infinity_norm_compatibility}) ensures the RE condition of $\hat{\Sigma}_t$ conditioned on the matrix concentration of the empirical Gram matrix $\hat{\Sigma}_t$.

\begin{lemma}\label{lemma:gram_mat_infinity_norm_RE}
Suppose that the RE condition holds for $\Sigma_0$  and the index set $S$ with cardinality $s = |S|$, with restricted eigenvalue $\phi^2_{\text{RE}}(\Sigma_0, S) > 0$, and that $\| \Sigma_1 - \Sigma_0 \|_\infty \leq \Delta$, where $32s \Delta \leq \phi^2_{\text{RE}}(\Sigma_0,S)$.
Then, for the set $S$, the RE condition holds as well for $\Sigma_1$, with $\phi^2_{\text{RE}}(\Sigma_1, S) \geq \phi^2_{\text{RE}}(\Sigma_0, S)/2$.
\end{lemma}

\proof{}
The proof is an adaptation of Lemma 6.17 in \cite{buhlmann2011statistics} to the RE condition.
\begin{align*}
\left| \beta^\top \Sigma_1 \beta - \beta^\top \Sigma_0 \beta \right| 
&= \left| \beta^\top (\Sigma_1 - \Sigma_0) \beta \right| \\
&\leq \| \Sigma_1 - \Sigma_0 \|_{\infty} \| \beta\|_1^2\\
&\leq \Delta \| \beta\|_1^2
\end{align*}
For $\beta$ such that $\| \beta_{S^c} \| \leq 3 \| \beta_S \|$, we have the RE condition satisfied for $\Sigma_0$. Hence, we have
\begin{align*}
    \| \beta \|_1 \leq 4 \| \beta_S \|_1 \leq 4 \sqrt{s}\| \beta_S \|_2 \leq 4 \sqrt{s}\| \beta \|_2 \leq \frac{4\sqrt{s_0 \beta^\top \Sigma_0 \beta}}{\phi_{\text{RE}}(\Sigma_0, S) } \,.
\end{align*}
Therefore, it follows that
\begin{align*}
    \left| \beta^\top \Sigma_1 \beta - \beta^\top \Sigma_0 \beta \right| 
    \leq \frac{16 s \Delta \beta^\top \Sigma_0 \beta}{\phi_{\text{RE}}^2(\Sigma_0, S) } \,.
\end{align*}
Since $\beta^\top \Sigma_0 \beta > 0$, dividing the both sides by $\beta^\top \Sigma_0 \beta$ gives
\begin{align*}
    \left| \frac{\beta^\top \Sigma_1 \beta}{\beta^\top \Sigma_0 \beta} - 1 \right| 
    \leq \frac{16 s \Delta  }{\phi_{\text{RE}}^2(\Sigma_0, S) } 
\end{align*}
Now, since  $32s \Delta \leq \phi^2_{\text{RE}}(\Sigma_0,S)$, it follows that
\begin{align*}
     \frac{1}{2}\cdot \frac{\beta^\top \Sigma_0 \beta}{\|\beta\|^2_2} 
     \leq \frac{\beta^\top \Sigma_1 \beta}{\|\beta\|^2_2} 
     \leq \frac{3}{2}\cdot \frac{\beta^\top \Sigma_0 \beta}{\|\beta\|^2_2}\,.
\end{align*}
Hence, 
\begin{align*}
    \phi^2_{\text{RE}}(\Sigma_1, S) \geq \frac{\phi^2_{\text{RE}}(\Sigma_0, S)}{2} \,.
\end{align*}

\endproof

\subsection{Proof of Theorem~\ref{thm:GLM-Lasso-regret-bound_re}}

\proof{}
The proof of Theorem~\ref{thm:GLM-Lasso-regret-bound_re} follows the similar arguments as the proof of Theorem~\ref{thm:GLM-Lasso-regret-bound_CC}. The only difference is that we use $\ell_2$ error bound $\| \hat{\beta}_t - \beta^*\|_2$ instead of $\| \hat{\beta}_t - \beta^*\|_1$.
First, note that
\begin{align*}
    \mathbb{P}\left( \mu(X_t^\top \hat{\beta}_t) \geq \mu(X_{a_t^*}^\top  \hat{\beta}_t)\right)
    &\leq \mathbb{P}\left( | \mu(X_t^\top \hat{\beta}_t) - \mu(X_t^\top \beta^*)| + |\mu(X_{a_t^*}^\top  \hat{\beta}_t) - \mu(X_{a^*_t}^\top \beta^*)| \geq \text{Reg}(t) \right)\\
    &\leq \mathbb{P}\left( \kappa_1 \| \hat{\beta}_t - \beta^*\|_2 \| X_t\|_2 + \kappa_1 \| \hat{\beta}_t - \beta^*\|_2 \| X_t^*\|_2 \geq \text{Reg}(t) \right)\\
    &\leq \mathbb{P}\left( 2 \kappa_1 \| \hat{\beta}_t - \beta^*\|_2\geq \text{Reg}(t) \right).
\end{align*}
For an arbitrary constant $g_t > 0$, we continue with expected regret $\mathbb{E}[\text{Reg}(t)]$ for $t > T_0$.
\begin{align*}
    \mathcal{R}(t) 
    &\leq   \kappa_1 g_t +  \kappa_1 \mathbb{P}\left( 2\| \hat{\beta}_t - \beta^*\|_2 \geq g_t,  \mathcal{E}_t\right) + 2\kappa_1 x_{\max} b\mathbb{P}(\mathcal{E}_t^c)\,.
\end{align*}
Hence, the cumulative regret is bounded by
 \begin{align*}
       \sum_{t=1}^T \mathcal{R}(t) &\leq 2\kappa_1 x_{\max} b T_0 + \kappa_1 \sum_{\mathclap{t= \lceil T_0 \rceil}}^T g_t + \kappa_1 \sum_{\mathclap{t= \lceil T_0 \rceil}}^T\mathbb{P}\left( 2\| \hat{\beta}_t - \beta^*\|_2 \geq g_t, \mathcal{E}_t\right) + 2\kappa_1 x_{\max} b\sum_{\mathclap{t= \lceil T_0 \rceil}}^T\mathbb{P}(\mathcal{E}_t^c)\,.
 \end{align*}
Let $g_t := \frac{3 \sqrt{s_0} \lambda_t}{2 \kappa_0\phi^2_t} = \frac{6 \sigma x_{\max}}{\kappa_0\phi^2_t} \sqrt{  \frac{s_0 (4\log t + 2\log d)}{t} }$. From Lemma~\ref{lemma:oracle_ineq_GLM_L1}, we have
\begin{align*}
    \mathbb{P}\left( 2\| \hat{\beta}_t - \beta^*\|_2 \geq g_t,  \mathcal{E}_t\right) \leq \frac{2}{t^2}
\end{align*}
for all $t$. Therefore, it follows that
\begin{align*}
    \sum_{t=\lceil T_0 \rceil}^T\mathbb{P}\left( 2\| \hat{\beta}_t - \beta^*\|_2 \geq g_t,   \mathcal{E}_t\right)
    \leq \sum_{t=1}^T\mathbb{P}\left( 2\| \hat{\beta}_t - \beta^*\|_2 \geq g_t,   \mathcal{E}_t\right)
    \leq \frac{\pi^2}{3} < 4 \,.
\end{align*}
For $t \geq T_0$, we have $\phi^2_t \geq \frac{ \phi^2_1}{2\nu }$ provided that event $\mathcal{E}_t$ holds. Hence, we have
\begin{align*}
    \sum_{t=\lceil T_0 \rceil}^T g_t &= \sum_{t=\lceil T_0 \rceil}^T\frac{6 \sigma x_{\max} }{\kappa_0\phi^2_t} \sqrt{\frac{ s_0 (4\log t + 2\log d) }{t} } \\
    &\leq \sum_{t=\lceil T_0 \rceil}^T\frac{12 \nu \sigma x_{\max}  }{\kappa_0\phi^2_1 } \sqrt{ \frac{ s_0 (4\log t + 2\log d) }{t} } \\
    &\leq \frac{12 \nu \sigma x_{\max} \sqrt{ s_0 (4\log T + 2\log d) }}{\kappa_0\phi^2_1}\sum_{t= 1}^{T}\frac{1}{\sqrt{t}} \\
    &\leq \frac{24 \nu \sigma x_{\max} \sqrt{ s_0 (4\log T + 2\log d) }}{\kappa_0\phi^2_1} \sqrt{T}
\end{align*}
where  the last inequality is from the fact that $\sum_{t= 1}^{T}\frac{1}{\sqrt{t}} \leq \int_{t= 0}^{T}\frac{1}{\sqrt{t}} = 2\sqrt{T}$.
Combining all the results with the bounds on $T_0$ and $\sum_{t= \lceil T_0 \rceil}^T\mathbb{P}(\mathcal{E}_t^c)$ from the proof of Theorem~\ref{thm:GLM-Lasso-regret-bound_CC}, the expected regret under the RE condition is bounded by
\begin{align*}
    \mathcal{R}^\pi(T) &\leq  4 \kappa_1 + \frac{4\kappa_1 x_{\max} b(\log(2d^2)+1)}{C_2(\phi_1,s_0)^2} + \frac{48 \kappa_1\nu \sigma x_{\max} \sqrt{  s_0 T \log (dT) }}{\kappa_0\phi^2_1} 
\end{align*}
where $C_2(\phi_1, s_0) = \min\!\Big(\frac{1}{2}, \frac{\phi_1^2}{256s_0 \nu x_{\max}^2} \Big)$.
\endproof

\section{Regret Analysis for \textit{K}-Armed Case}\label{sec:K-arm-regret-analysis}

\subsection{Proof Outline of Theorem~\ref{thm:GLM-Lasso-regret-bound_CC_K-arm}}
\label{app:thm:GLM-Lasso-regret-bound_CC_K-arm}

As discussed in Section~\ref{sec:K-arms}, the analysis for the \textit{K}-armed bandit mostly follows the proof of the two-armed bandit analysis in Section~\ref{sec:regret_analysis}. Assuming the compatibility condition of the empirical Gram matrix $\hat{\Sigma}_t$, the Lasso oracle inequality for adapted samples in Lemma~\ref{lemma:oracle_ineq_GLM_L1} can be directly applied. Hence, what we have left is ensuring the compatibility condition of~$\hat{\Sigma}_t$. 
As before, for each $\mathbb{E}[X_\tau X_\tau^\top | \mathcal{F}_\tau]$ in $\Sigma_t$, the history $\mathcal{F}_\tau$ affects how feature vector $X_\tau$ is chosen.   
Similar to the two-armed bandit case, we rewrite $\Sigma_t$ as
\begin{align*}\label{eq:adapted_cov_matrix_K-arm}
    \Sigma_t
    = \frac{1}{t}\sum_{\tau=1}^t\sum_{i = 1}^K \mathbb{E}_{\mathcal{X}_t}\big[X_{\tau, i} X_{\tau, i}^\top  \mathbbm{1}\{X_{\tau, i} = \argmax_{X \in \mathcal{X}_\tau}
  X^\top \hat{\beta}_{\tau}\} \mid \hat{\beta}_{\tau} \big]\,. 
\end{align*}
Recall that the compatibility condition is only assumed for the
theoretical Gram matrix~$\Sigma$ (Assumption~\ref{assum:comp_condition}).
Again, the adapted Gram matrix $\Sigma_t$ is used to bridge $\Sigma$ and $\hat{\Sigma}_t$ to ensure the compatibility of $\hat{\Sigma}_t$.
The key difference between the two-armed bandit analysis and the $K$-armed bandit analysis lies in how $\Sigma_t$ is controlled by $\Sigma$. In particular, under the balanced covariance condition in Assumption~\ref{assum:balanced_cov}, we show the following lemma which is a generalization of Lemma~\ref{lemma:cov_matrix_lowerbound_asymmetry}.

\begin{lemma}\label{lemma:cov_matrix_lowerbound_asymmetry_K-arm}
Suppose Assumption~\ref{assum:balanced_cov} holds. For a fixed vector $\beta \in \mathbb{R}^d$, we have
\begin{equation*}
     \sum_{i=1}^K \mathbb{E}_{\mathcal{X}_t}\Big[X_{t,i} X_{t,i}^\top \mathbbm{1}\{X_i =
     \argmax_{\mathclap{X \in \mathcal{X}_t}} X^\top \beta\}\Big]
     \succcurlyeq (2 \nu C_\mathcal{X})^{-1}\Sigma \,.
\end{equation*}
\end{lemma}

With this result, we can lower-bound the compatibility constant
$\phi^2(\Sigma_t,S_0)$ of the adapted Gram matrix in terms of the compatibility constant
$\phi^2(\Sigma,S_0)$ for the theoretical Gram matrix.
That is, we have $\Sigma_t \succcurlyeq (2 \nu C_\mathcal{X})^{-1} \Sigma$ which implies that
\begin{align*}
\phi^2(\Sigma_t, S_0) \geq \frac{\phi^2(\Sigma, S_0)}{2\nu C_\mathcal{X} } > 0 \,.    
\end{align*}
Hence, $\Sigma_t$ satisfies the compatibility condition.
Then, we can show that $\hat{\Sigma}_t$ concentrates to $\Sigma_t$ with high probability which directly follows from applying  Lemma~\ref{lemma:cov_matrix_lowerbound_asymmetry}, which is formally stated as follows.
\begin{corollary}
\label{lemma:Sigma_hat_concentration_K-arm}
For $t \geq \frac{2\log(2d^2)}{C_1(s_0)^2}$ where $C_1(s_0) = \min\!\Big(\frac{1}{2}, \frac{\phi_0^2}{256s_0 \nu C_\mathcal{X} x_{\max}^2} \Big)$, we have
\begin{align*}
    \mathbb{P}\left( \| \Sigma_t - \hat{\Sigma}_t \|_{\infty} \geq  \frac{ \phi^2_0}{32  s_0\nu C_\mathcal{X}} \right) 
    &\leq \exp\left\{-\frac{t C_1(s_0)^2}{2}\right\} \,.
\end{align*}
\end{corollary}
Now, we can invoke Corollary~\ref{cor:gram_mat_infinity_norm_compatibility} to connect this matrix concentration result to
guaranteeing the compatibility condition of $\hat{\Sigma}_t$. Therefore, $\hat{\Sigma}_t$ satisfies the compatibility condition with compatibility constant $\phi_t^2 = \frac{\phi_0^2}{4\nu C_\mathcal{X}} > 0$. The rest of the proof of Theorem~\ref{thm:GLM-Lasso-regret-bound_CC_K-arm} directly follows the proof of Theorem~\ref{thm:GLM-Lasso-regret-bound_CC} using this compatibility constant.

\subsection{Proof of Lemma~\ref{lemma:cov_matrix_lowerbound_asymmetry_K-arm}}
\proof{}
Since the distribution of $\mathcal{X}_t = \{X_{t,1}, ..., X_{t,K}\}$ is time-invariant, we suppress the subscript on $t$ and write $\mathcal{X} = \{X_1, ..., X_K\}$. Let joint distribution of $\mathcal{X}$ as $p_\mathcal{X} (x_1, ..., x_K) = p_\mathcal{X} (\mathbf{x})$ where we let $\mathbf{x} = (x_1, ..., x_K)$. All expectations in this proof is taken with respect to the tuple $\mathcal{X}$.  Then the theoretical Gram matrix is defined as
\begin{align*}
    \mathbb{E}[\mathbf{X}^\top \mathbf{X}] 
    &= \mathbb{E}\left[ \sum_{i=1}^K X_i X_i^\top \right] \\
    &= \int (x_1 x_1^\top + ... +  x_K x_K^\top) p_\mathcal{X}(\mathbf{x})   d \mathbf{x}
\end{align*}
Let's first focus on $\int x_1 x_1^\top p_\mathcal{X} (\mathbf{x})   d \mathbf{x}$. 
\begin{align*}
    \int x_1 x_1^\top  p_\mathcal{X} (\mathbf{x})   d \mathbf{x}
    &= \int x_1 x_1^\top \mathbbm{1}\Big\{ x_1 = \argmax_{x_i \in \mathcal{X} } x_i^\top \beta \Big\} p_\mathcal{X} (\mathbf{x})   d \mathbf{x}\\
    &\quad + \int x_1 x_1^\top \mathbbm{1}\Big\{ x_1 = \argmin_{x_i \in \mathcal{X} } x_i^\top \beta \Big\} p_\mathcal{X} (\mathbf{x})   d \mathbf{x}\\
    &\quad + \int x_1 x_1^\top \mathbbm{1}\Big\{ x_1 \neq \argmax_{x_i \in \mathcal{X} } x_i^\top \beta, x_1 \neq \argmin_{x_i \in \mathcal{X} } x_i^\top \beta \Big\} p_\mathcal{X} (\mathbf{x})   d \mathbf{x} \,.
\end{align*}
We define three disjoint sets of possible orderings for $\{1, ..., K\}$ as follows.
\begin{definition}\label{def:ordering_sets}
We define the following sets of permutations of $(1, ..., K)$.
\begin{align*}
    \mathcal{I}_1^{\max} &:= \{ \text{indices  $(i_1, ..., i_K) $ such that $i_K = 1$} \}\\
    \mathcal{I}_1^{\min} &:= \{ \text{indices $(i_1, ..., i_K)$ such that $i_1 = 1$} \}\\
    \mathcal{I}_1^{\text{mid}} &:= \{ \text{indices $(i_1, ..., i_K)$ such that $i_1 \neq 1$ and $i_K \neq 1$} \} .
\end{align*}
\end{definition}
Then, for $\int x_1 x_1^\top \mathbbm{1}\{ x_1 = \argmin_{x_i \in \mathcal{X} } x_i^\top \beta \} p_\mathcal{X} (\mathbf{x})   d \mathbf{x}$, we can write
\begin{align*}
    \int x_1 x_1^\top \mathbbm{1}\Big\{ x_1 = \argmin_{x_i \in \mathcal{X} } x_i^\top \beta \Big\} p_\mathcal{X} (\mathbf{x})   d \mathbf{x} 
    &= \sum_{(i_1, ..., i_K) \in \mathcal{I}_1^{\min}} \int x_1 x_1^\top \mathbbm{1}\Big\{  x_{i_1}^\top \beta \leq ... \leq x_{i_K}^\top \beta \Big\} p_\mathcal{X} (\mathbf{x})   d \mathbf{x}
\end{align*}
Then for any $(i_1, ..., i_K) \in \mathcal{I}_1^{\min}$, 
\begin{align*}
    \int x_1 x_1^\top \mathbbm{1}\Big\{  x_{i_1}^\top \beta \leq ... \leq x_{i_K}^\top \beta \Big\} p_\mathcal{X} (\mathbf{x})   d \mathbf{x} 
    &= \int x_1 x_1^\top \mathbbm{1}\Big\{  -x_{i_1}^\top \beta \geq ... \geq -x_{i_K}^\top \beta \Big\} p_\mathcal{X} (\mathbf{x})   d \mathbf{x}\\
    &\preccurlyeq \nu\int x_1 x_1^\top \mathbbm{1}\Big\{  -x_{i_1}^\top \beta \geq ... \geq -x_{i_K}^\top \beta \Big\} p_\mathcal{X} (-\mathbf{x})   d \mathbf{x}\\
    &= \nu\int x_1 x_1^\top \mathbbm{1}\Big\{  x_{i_1}^\top \beta \geq ... \geq x_{i_K}^\top \beta \Big\} p_\mathcal{X} (\mathbf{x})   d \mathbf{x}
\end{align*}
where  the inequality is again from Assumption~\ref{assum:rho}.
Since the elements in $\mathcal{I}_1^{\min}$ can be considered as reversed orderings of elements in $\mathcal{I}_1^{\max}$ (and obviously $| \mathcal{I}_1^{\min} | = | \mathcal{I}_1^{\max} |$),
\begin{align*}
\mathbb{E}\left[X_1 X_1^\top \mathbbm{1}\{X_1 = \argmin_{X \in \mathcal{X}} X^\top \beta\} \right]
    &= \int x_1 x_1^\top \mathbbm{1}\Big\{ x_1 = \argmin_{x_i \in \mathcal{X} } x_i^\top \beta \Big\} p_\mathcal{X} (\mathbf{x})   d \mathbf{x} \\
    &=   \sum_{(i_1, ..., i_K) \in \mathcal{I}_1^{\min}}  \int x_1 x_1^\top \mathbbm{1}\Big\{  x_{i_1}^\top \beta \leq ... \leq x_{i_K}^\top \beta \Big\} p_\mathcal{X} (\mathbf{x})   d \mathbf{x} \\
    &\preccurlyeq \sum_{(i_1, ..., i_K) \in \mathcal{I}_1^{\min}} \nu  \int x_1 x_1^\top \mathbbm{1}\Big\{  x_{i_1}^\top \beta \geq ... \geq x_{i_K}^\top \beta \Big\} p_\mathcal{X} (\mathbf{x})   d \mathbf{x}\\
    &= \nu \int x_1 x_1^\top \mathbbm{1}\Big\{ x_1 = \argmax_{x_i \in \mathcal{X} } x_i^\top \beta \Big\} p_\mathcal{X} (\mathbf{x})   d \mathbf{x} \\
    &= \nu  \mathbb{E}\left[X_1 X_1^\top \mathbbm{1}\{X_1 = \argmax_{X \in \mathcal{X}} X^\top \beta\} \right] \,.
\end{align*}
Also, using the definitions of $\mathcal{I}_1^{\min}, \mathcal{I}_1^{\text{mid}}$ and $\mathcal{I}_1^{\max}$, we can rewrite $\mathbb{E}\left[ X_1 X^\top_1 \right]$.
\begin{align*}
    \mathbb{E}\left[ X_1 X^\top_1 \right]
    &= \mathbb{E}\left[X_1 X_1^\top \mathbbm{1}\{X_1 = \argmin_{X \in \mathcal{X}} X^\top \beta\} \right] + \mathbb{E}\left[X_1 X_1^\top \mathbbm{1}\{X_1 = \argmax_{X \in \mathcal{X}} X^\top \beta\} \right]\\
    &\quad + \mathbb{E}\left[X_1 X_1^\top \mathbbm{1}\{X_1 \neq \argmin_{X \in \mathcal{X}} X^\top \beta, X_1 \neq \argmax_{X \in \mathcal{X}} X^\top \beta\} \right]\\
    &= \sum_{(i_1,...,i_K) \in \mathcal{I}_1^{\text{min}}} \mathbb{E}\left[ X_1 X^\top_1  \mathbbm{1}\{ X_{i_1}^\top \beta < \cdots < X_{i_K}^\top \beta \} \right]\\
    &\quad + \sum_{(i_1,...,i_K) \in \mathcal{I}_1^{\text{max}}} \mathbb{E}\left[ X_1 X^\top_1  \mathbbm{1}\{ X_{i_1}^\top \beta < \cdots < X_{i_K}^\top \beta \} \right]\\
    &\quad + \sum_{(i_1,...,i_K) \in \mathcal{I}_1^{\text{mid}}} \mathbb{E}\left[ X_1 X^\top_1  \mathbbm{1}\{ X_{i_1}^\top \beta < \cdots < X_{i_K}^\top \beta \} \right]\\
    &= \sum_{(i_1,...,i_K) \in \mathcal{I}_1^{\text{min}}} \mathbb{E}\left[ X_{i_1} X^\top_{i_1}  \mathbbm{1}\{ X_{i_1}^\top \beta < \cdots < X_{i_K}^\top \beta \} \right]\\
     &\quad + \sum_{(i_1,...,i_K) \in \mathcal{I}_1^{\text{max}}} \mathbb{E}\left[ X_{i_K} X^\top_{i_K}  \mathbbm{1}\{ X_{i_1}^\top \beta < \cdots < X_{i_K}^\top \beta \} \right]\\
    &\quad + \sum_{(i_1,...,i_K) \in \mathcal{I}_1^{\text{mid}}} \mathbb{E}\left[ X_1 X^\top_1  \mathbbm{1}\{ X_{i_1}^\top \beta < \cdots < X_{i_K}^\top \beta \} \right] \,.
\end{align*}

s\noindent From Assumption~\ref{assum:balanced_cov}, we have
\begin{align*}
    \mathbb{E}\left[ X_1 X^\top_1  \mathbbm{1}\{ X_{i_1}^\top \beta < \cdots < X_{i_K}^\top \beta \} \right]
    \preccurlyeq C_\mathcal{X}  \mathbb{E}\left[ ( X_{i_1} X^\top_{i_1} + X_{i_K} X^\top_{i_K}) \mathbbm{1}\{ X_{i_1}^\top \beta < \cdots < X_{i_K}^\top \beta \} \right] \,.
\end{align*}
Then it follows that
\begin{align*}
    \mathbb{E}\left[ X_1 X^\top_1 \right]
    &\preccurlyeq  \sum_{(i_1,...,i_K) \in \mathcal{I}_1^{\text{min}}} \mathbb{E}\left[ X_{i_1} X^\top_{i_1}  \mathbbm{1}\{ X_{i_1}^\top \beta < \cdots < X_{i_K}^\top \beta \} \right]\\
    &\quad + \sum_{(i_1,...,i_K) \in \mathcal{I}_1^{\text{max}}} \mathbb{E}\left[ X_{i_K} X^\top_{i_K}  \mathbbm{1}\{ X_{i_1}^\top \beta < \cdots < X_{i_K}^\top \beta \} \right]\\
    &\quad +  \sum_{(i_1,...,i_K) \in \mathcal{I}_1^{\text{mid}}} C_\mathcal{X}  \mathbb{E}\left[  \big( X_{i_1} X^\top_{i_1} + X_{i_K} X^\top_{i_K} \big) \mathbbm{1}\{ X_{i_1}^\top \beta < \cdots < X_{i_K}^\top \beta \} \right]\\
    &\preccurlyeq 
     \sum_{(i_1,...,i_K) \in \mathcal{I}_1^{\text{min}}} C_\mathcal{X} \mathbb{E}\left[ \big(X_{i_1} X^\top_{i_1} + X_{i_K} X^\top_{i_K} \big)  \mathbbm{1}\{ X_{i_1}^\top \beta < \cdots < X_{i_K}^\top \beta \} \right]\\
    &\quad  + \sum_{(i_1,...,i_K) \in \mathcal{I}_1^{\text{max}}} C_\mathcal{X} \mathbb{E}\left[ \big(X_{i_1} X^\top_{i_1} + X_{i_K} X^\top_{i_K} \big) \mathbbm{1}\{ X_{i_1}^\top \beta < \cdots < X_{i_K}^\top \beta \} \right]\\
    &\quad + \sum_{(i_1,...,i_K) \in \mathcal{I}_1^{\text{mid}}} C_\mathcal{X}  \mathbb{E}\left[  \big( X_{i_1} X^\top_{i_1} + X_{i_K} X^\top_{i_K} \big) \mathbbm{1}\{ X_{i_1}^\top \beta < \cdots < X_{i_K}^\top \beta \} \right] \,.
\end{align*}
Since $\mathcal{I}_1^{\min}, \mathcal{I}_1^{\text{mid}}$ and $\mathcal{I}_1^{\max}$ are disjoint sets, we can write
\begin{align*}
    \mathbb{E}\left[X_i X_i^\top \mathbbm{1}\{X_i = \argmin_{X \in \mathcal{X}} X^\top \beta\} \right]
    &= \sum_{(i_1,...,i_K) \in \mathcal{I}_1^{\text{min}}} \mathbb{E}\left[ X_{i_1} X^\top_{i_1} \mathbbm{1}\{ X_{i_1}^\top \beta < \cdots < X_{i_K}^\top \beta \} \right]\\
    &\quad  + \sum_{(i_1,...,i_K) \in \mathcal{I}_1^{\text{max}}} \mathbb{E}\left[ X_{i_1} X^\top_{i_1} \mathbbm{1}\{ X_{i_1}^\top \beta < \cdots < X_{i_K}^\top \beta \} \right]\\
    &\quad + \sum_{(i_1,...,i_K) \in \mathcal{I}_1^{\text{mid}}} \mathbb{E}\left[ X_{i_1} X^\top_{i_1} \mathbbm{1}\{ X_{i_1}^\top \beta < \cdots < X_{i_K}^\top \beta \} \right] \,.
\end{align*}
We can also express $\mathbb{E}\left[X_i X_i^\top \mathbbm{1}\{X_i = \argmax_{X \in \mathcal{X}} X^\top \beta\} \right]$ similarly.
Therefore, we have
\begin{align*}
    \mathbb{E}\left[ X_1 X^\top_1 \right]
    &\preccurlyeq C_\mathcal{X} \sum_{i=1}^K \left(  \mathbb{E}\left[X_i X_i^\top \mathbbm{1}\{X_i = \argmin_{X \in \mathcal{X}} X^\top \beta\} \right] +   \mathbb{E}\left[X_i X_i^\top \mathbbm{1}\{X_i = \argmax_{X \in \mathcal{X}} X^\top \beta\} \right] \right)\\
    &\preccurlyeq C_\mathcal{X} (1+\nu) \sum_{i=1}^K \mathbb{E}\left[X_i X_i^\top \mathbbm{1}\{X_i = \argmax_{X \in \mathcal{X}} X^\top \beta\} \right] \,.
\end{align*}
Then, summing $\mathbb{E}\left[ X_j X^\top_j \right]$ over all $j = 1, ..., K$ gives
\begin{align*}
    \mathbb{E}[\mathbf{X}^\top \mathbf{X}] &= \sum_{j=1}^K \mathbb{E}\left[ X_j X^\top_j \right]
    \preccurlyeq K C_\mathcal{X} (1+\nu) \sum_{i=1}^K \mathbb{E}\left[X_i X_i^\top \mathbbm{1}\{X_i = \argmax_{X \in \mathcal{X}} X^\top \beta\} \right] \,.
\end{align*}
Hence,
\begin{align*}
    \sum_{i=1}^K \mathbb{E}\left[X_i X_i^\top \mathbbm{1}\{X_i = \argmax_{X \in \mathcal{X}} X^\top \beta\} \right] \succcurlyeq \frac{1}{C_\mathcal{X} (1+\nu)} \cdot \frac{1}{K}\mathbb{E}[\mathbf{X}^\top \mathbf{X}] \succcurlyeq (2 C_\mathcal{X} \nu)^{-1}\Sigma \,.
\end{align*}
\endproof

\subsection{Proposition \ref{prop:c_x_bound}}
\label{app:prop:c_x_bound}

\begin{proposition}\label{prop:c_x_bound}
In the case of independent arms, both a multivariate Gaussian distribution and a uniform distribution on a unit sphere satisfy Assumption~\ref{assum:balanced_cov} with  $C_\mathcal{X} = \mathcal{O}(1)$. For an arbitrary distribution, it holds with $C_{\mathcal{X}} = \binom{K-1}{K_0}$  where $K_0 = \lceil (K-1)/2 \rceil$.
\end{proposition}

\noindent The proof of Proposition \ref{prop:c_x_bound} involves the following few technical lemmas.

\begin{lemma}\label{lemma:N_var_middle_case_bound_gaussian}
Suppose each $X_i \in \mathbb{R}^d$ is i.i.d. Gaussian with  mean $\mu$ and covariance matrix $\Gamma$.
For any permutation $(i_1,...,i_K)$ of $(1, ..., K)$, any integer $k \in \{2, ..., K-1\}$ and fixed $\beta$,
\begin{align*}
 \mathbb{E}\Big[X_{i_k} X_{i_k}^\top \mathbbm{1}\{ X_{i_1}^\top \beta < ... < X_{i_K}^\top \beta \} \Big] &\preccurlyeq \mathbb{E}\Big[X_{i_1} X_{i_1}^\top \mathbbm{1}\{ X_{i_1}^\top \beta < ... < X_{i_K}^\top \beta \} \Big]\\
 &\quad + \mathbb{E}\Big[X_{i_K} X_{i_K}^\top \mathbbm{1}\{ X_{i_1}^\top \beta < ... < X_{i_K}^\top \beta \} \Big] \,.
\end{align*}
\end{lemma}

\proof{}
It suffices to show that for any $y \in \mathbb{R}^d$
\begin{align*}
    &\mathbb{E}\left[ (X^\top_{i_k} y)^2 \mathbbm{1}\{ X_{i_1}^\top \beta < ... < X_{i_K}^\top \beta \} \right]\\
    &\leq  \mathbb{E}\left[ (X^\top_{i_1} y)^2 \mathbbm{1}\{ X_{i_1}^\top \beta < ... < X_{i_K}^\top \beta \} \right]
    + \mathbb{E}\left[ (X^\top_{i_K} y)^2 \mathbbm{1}\{ X_{i_1}^\top \beta < ... < X_{i_K}^\top \beta \} \right] \,.
\end{align*}
Now, we can write
\begin{align*}
    y = \tilde{\beta}( \tilde{\beta}^\top y) + \sum_{j=1}^{d-1} g_j g_j^\top y
    := \tilde{\beta} w_0  + \sum_{j=1}^{d-1} g_j g_j^\top y \,.
\end{align*}
where $w_0 = \tilde{\beta}^\top y$ and $\tilde{\beta} = \frac{\beta}{\| \beta \|}$ and $\left[\tilde{\beta}, g_i, ..., g_{d-1}\right]$ form an orthonormal basis. For $i \in [N]$, we can write
\begin{align*}
    X_i^\top y &= (X_i^\top \tilde{\beta}) w_0 + X_i^\top \left( \sum_{j=1}^{d-1} g_j g_j^\top \right) y \\
    &= (X_i^\top \tilde{\beta}) w_0 + \left[\left( \sum_{j=1}^{d-1} g_j g_j^\top \right) X_i \right]^\top y \,.
\end{align*}
Then we define the following two random variables
\begin{align*}
    U_i := X^\top_i \tilde{\beta}, \qquad
    V_i :=  G X_i
\end{align*}
where $G = \sum_{j=1}^{d-1} g_j g_j^\top$. Then we have
\begin{align*}
\begin{bmatrix}
U_i\\
V_i
\end{bmatrix} &\sim  
\mathcal{N}
\begin{pmatrix}
\begin{bmatrix}
\mu^\top \tilde{\beta}\\
G \mu
\end{bmatrix}\!\!,&
\begin{bmatrix}
A_{11} & A_{12} \\
A_{21} & A_{22}
\end{bmatrix}
\end{pmatrix}
\end{align*}
where
\begin{align*}
    A_{11} &= \tilde{\beta}^\top \Gamma \tilde{\beta} \in \mathbb{R}\\
    A_{12} &= A_{21}^\top = \tilde{\beta}^\top \Gamma G^\top \in \mathbb{R}^{1 \times d} \\
    A_{22} &= G \Gamma G^\top \in \mathbb{R}^{d \times d} \,.
\end{align*}
Then, we know from Lemma~\ref{lemma:conditional_gaussian} that the conditional distribution $V_i \mid U_i$ of a multivariate normal distribution is also a multivariate normal distribution. In particular,
\begin{align*}
    V_i \mid U_i = u_i \sim \mathcal{N}\left(G\mu + A_{21} A_{11}^{-1} (u_i - \mu^\top \tilde{\beta}), B \right)
\end{align*}
where $B = A_{22} - A_{21} A_{11}^{-1} A_{12}$. Therefore, given $U_{i_k} = u_{i_k}$, we can write
\begin{align*}
    X_{i_k}^\top y  &= u_{i_k} w_0 + V_{i_k}^\top y\\
    &= u_{i_k} w_0 + \left(G\mu + A_{21} A_{11}^{-1} (u_{i_k} - \mu^\top \tilde{\beta}) + B^{1/2} Z\right)^\top y \,.
\end{align*}
where $Z \sim \mathcal{N}(0, I_{d})$ and $Z \indep U_{i_k}$. Rearranging gives
\begin{align*}
    X_{i_k}^\top y = u_{i_k} \left( w_0 +  A_{11}^{-1} A_{12} y \right) + \left( G\mu -  A_{21} A_{11}^{-1}  \mu^\top \tilde{\beta} \right)^\top y +  Z^\top B^{1/2} y \,.
\end{align*}
Hence, $X_{i_k}^\top y$ is a linear function of $u_{i_k}$. Then it follows that
\begin{align*}
    \left( X^\top_{i_k} y \right)^2 
    &= \left[ u_{i_k} \left( w_0 +  A_{11}^{-1} A_{12} y \right) + \left( G\mu -  A_{21} A_{11}^{-1}  \mu^\top \tilde{\beta} \right)^\top y +  Z^\top B^{1/2} y \right]^2\\
    &\leq \max \Bigg\{ \left[ u_{i_1} \left( w_0 +  A_{11}^{-1} A_{12} y \right) + \left( G\mu -  A_{21} A_{11}^{-1}  \mu^\top \tilde{\beta} \right)^\top y +  Z^\top B^{1/2} y \right]^2 ,\\
    &\qquad \qquad \left[ u_{i_K} \left( w_0 +  A_{11}^{-1} A_{12} y \right) + \left( G\mu -  A_{21} A_{11}^{-1}  \mu^\top \tilde{\beta} \right)^\top y +  Z^\top B^{1/2} y \right]^2
    \Bigg\}\\
    &\leq  \left[ u_{i_1} \left( w_0 +  A_{11}^{-1} A_{12} y \right) + \left( G\mu -  A_{21} A_{11}^{-1}  \mu^\top \tilde{\beta} \right)^\top y +  Z^\top B^{1/2} y \right]^2 \\
    &\quad + \left[ u_{i_K} \left( w_0 +  A_{11}^{-1} A_{12} y \right) + \left( G\mu -  A_{21} A_{11}^{-1}  \mu^\top \tilde{\beta} \right)^\top y +  Z^\top B^{1/2} y \right]^2 .
\end{align*}
Therefore, it follows that
\begin{align*}
    &\mathbb{E}\left[ (X^\top_{i_k} y)^2 \mathbbm{1}\{ X_{i_1}^\top \beta < ... < X_{i_K}^\top \beta \} \right]\\
    &\leq  \mathbb{E}\left[ (X^\top_{i_1} y)^2 \mathbbm{1}\{ X_{i_1}^\top \beta < ... < X_{i_K}^\top \beta \} \right]
    + \mathbb{E}\left[ (X^\top_{i_K} y)^2 \mathbbm{1}\{ X_{i_1}^\top \beta < ... < X_{i_K}^\top \beta \} \right] \,.
\end{align*}
Hence,
\begin{align*}
    \mathbb{E}\left[ X_{i_k} X^\top_{i_k}  \mathbbm{1}\{ X_{i_1}^\top \beta < ... < X_{i_K}^\top \beta \} \right]
    \preccurlyeq  \mathbb{E}\left[ ( X_{i_1} X^\top_{i_1} + X_{i_K} X^\top_{i_K}) \mathbbm{1}\{ X_{i_1}^\top \beta < ... < X_{i_K}^\top \beta \} \right] \,.
\end{align*}
\endproof

\begin{lemma}\label{lemma:N_var_middle_case_bound_uniform}
Suppose $X \in \mathbb{R}^d$ is uniformly distributed on the unit sphere $\mathcal{S}^{d-1}$ and $K = o(d)$. For fixed vector $\beta \in \mathbb{R}^d$ and a given integer $k \in \{2,..., K-1\}$,
\begin{align*}
    \mathbb{E}\left[ X_{i_k} X^\top_{i_k}  \mathbbm{1}\{ X_{i_1}^\top \beta < ... < X_{i_K}^\top \beta \} \right]
    \preccurlyeq  C_\mathcal{X} \mathbb{E}\left[ ( X_{i_1} X^\top_{i_1} + X_{i_K} X^\top_{i_K}) \mathbbm{1}\{ X_{i_1}^\top \beta < ... < X_{i_K}^\top \beta \} \right] .
\end{align*}
where $C_\mathcal{X} = \mathcal{O}(1)$.
\end{lemma}

\proof{}
Here, we instead show directly
\begin{align*}
    \mathbb{E}[X X^\top ]
    \preccurlyeq C \left( \mathbb{E}\Big[X X^\top \mathbbm{1}\{X = \argmax_{\mathclap{X_i \in \{X_1, ..., X_K\}}} X_i^\top \tilde{\beta}\}\Big] + \mathbb{E}\Big[X X^\top \mathbbm{1}\{X = \argmin_{\mathclap{X_i \in \{X_1, ..., X_K\}}} X_i^\top \tilde{\beta}\}\Big] \right)
\end{align*}
for some constant $C$. It can be shown that if $C = \mathcal{O}(1)$, then the claim holds with $C_\mathcal{X} =  \mathcal{O}(1)$.
Suppose $X \in \mathbb{R}^d$ is uniformly distributed on the unit sphere $\mathcal{S}^{d-1} := \{s \in \mathbb{R}^d: \| s \|_2 = 1\}$. Then by Lemma~2 in \cite{cambanis1981theory}, we can write for each $X_i$,
\begin{align*}
    X_i \sim \left( B_i U_{i,1}, (1-B_i^2)^{1/2} U_{i,2} \right)
\end{align*}
where $B_i \sim \text{beta}\left(\frac{1}{2}, \frac{d-1}{2} \right)$, $U_{i,1} = \pm 1$ with probability $\frac{1}{2}$, $U_{i,2} \sim \text{unif}(\mathcal{S}^{d-2})$. $U_{i,1}$, $U_{i,2}$ and $B_i$ are independent of each other. Similar to the analysis of the Gaussian case, we can normalize $\beta$ so that $\tilde{\beta} = \frac{\beta}{\| \beta \|}$.
Without loss of generality, assume that $\tilde{\beta} = [1,0,...,0]^\top$. That is, only the first element is non-zero. We can do this since $X$ is spherical and rotation invariant. Then we can write
\begin{align*}
    \mathbb{E}\Big[X X^\top \mathbbm{1}\{X = \argmax_{\mathclap{X_i \in \{X_1, ..., X_K\}}} X_i^\top \tilde{\beta}\}\Big]
    = \mathbb{E}\Big[X X^\top \mathbbm{1}\{X = \argmax_{\mathclap{X_i \in \{X_1, ..., X_K\}}} X_i^{(1)}\}\Big]
\end{align*}
where $X_i^{(1)}$ is the first element of $ X_i$.
Similarly,
\begin{align*}
    \mathbb{E}\Big[X X^\top \mathbbm{1}\{X = \argmin_{\mathclap{X_i \in \{X_1, ..., X_K\}}} X_i^\top \tilde{\beta}\}\Big]
    = \mathbb{E}\Big[X X^\top \mathbbm{1}\{X = \argmin_{\mathclap{X_i \in \{X_1, ..., X_K\}}} X_i^{(1)}\}\Big] \,.
\end{align*}
Now, from the definition of $X$, for $B \sim \text{beta}\left(\frac{1}{2}, \frac{d-1}{2} \right)$ we have
\begin{align*}
    X_i X_i^\top = \begin{bmatrix}
B_i^2  & B_i\sqrt{1-B_i^2} U_{i,1} U_{i,2}^\top \\
B_i\sqrt{1-B_i^2} U_{i,1} U_{i,2} & (1-B_i^2) U_{i,2} U_{i,2}^\top
\end{bmatrix} .
\end{align*}
By the independence of $U_1, U_2$, and $B$, we have
\begin{align*}
    \mathbb{E}\left[ X X^\top \right] 
    = \mathbb{E}\begin{bmatrix}
\enskip B^2 & 0 \\
\enskip 0 & \frac{1}{d-1} (1-B^2) I_{d-1}
\end{bmatrix} .
\end{align*}
By the definitions of $B_i$ and $U_{i,1}$, it follows that
\begin{align*}
    \mathbb{E}\Big[X X^\top \mathbbm{1}\{B = \max_{\mathclap{B_i \in \{B_1, ..., B_K\}}} B_i\}\Big]
    \preccurlyeq \mathbb{E}\Big[X X^\top \mathbbm{1}\{X = \argmax_{\mathclap{X_i \in \{X_1, ..., X_K\}}} X_i^{(1)}\}\Big]
    + \mathbb{E}\Big[X X^\top \mathbbm{1}\{X = \argmin_{\mathclap{X_i \in \{X_1, ..., X_K\}}} X_i^{(1)}\}\Big].
\end{align*}
Since $\mathbb{E}[B^2]=\frac{(\alpha + 1) \alpha}{(\alpha + \beta + 1)(\alpha + \beta)}$ for $B \sim \text{beta}(\alpha, \beta)$, we have $\mathbb{E}[B^2] = \frac{3}{d(d+2)}$ and $\frac{1-\mathbb{E}[B^2]}{d-1} = \frac{d+3}{d(d+2)}$ using $\alpha = \frac{1}{2}$ and $\beta = \frac{d-1}{2}$.
Clearly, $\lambda_{\min}(\mathbb{E}\left[ X X^\top \right] ) = \frac{3}{d(d+2)}$. 
Similarly, for the matrix $\mathbb{E}\left[ X X^\top \mathbbm{1}\{B = \max_{i} B_i\} \right] $, we have
\begin{align*}
    \mathbb{E}\left[ X X^\top \mathbbm{1}\{B = \max_{i} B_i\} \right] 
    = \mathbb{E}\begin{bmatrix}
\enskip B^2 \mathbbm{1}\{B = \max_{i} B_i\} & 0 \\
\enskip 0 & \frac{1}{d-1} (1-B^2) \mathbbm{1}\{B = \max_{i} B_i\} I_{d-1}
\end{bmatrix} .
\end{align*}
Note that $\mathbb{E}[B^2 \mathbbm{1}\{B = \max_{i} B_i\}] = \sum_{j=1}^K \mathbb{E}[B_j^2 \mathbbm{1}\{B_j = \max_{i} B_i\}] \geq \mathbb{E}[B^2]$. Then, we need to show 
\begin{align*}
 C (1-\mathbb{E}[B^2 \mathbbm{1}\{B = \max_{i} B_i\}] ) \geq 1-\mathbb{E}[B^2]
\end{align*}
for some $C$. Note that $\mathbb{E}[B^2 \mathbbm{1}\{B = \max_{i} B_i\}] \leq N\mathbb{E}[B^2]$. Hence, we can show 
\begin{align*}
    C \geq \frac{1-\mathbb{E}[B^2]}{1-N\mathbb{E}[B^2]} = \frac{1 - \frac{3}{d(d + 2)}}{1 - \frac{3K}{d(d + 2)}} = \frac{d^2 +d - 3}{d^2 +d - 3K} \,.
\end{align*}
Since $K = o(d)$, we have $C = \mathcal{O}(1)$. Hence, 
\begin{align*}
    \mathbb{E}[X X^\top ]
    &\preccurlyeq C \mathbb{E}\Big[X X^\top \mathbbm{1}\{B = \max_{\mathclap{B_i \in \{B_1, ..., B_K\}}} B_i\}\Big]\\
    &\preccurlyeq C\left( \mathbb{E}\Big[X X^\top \mathbbm{1}\{X = \argmax_{\mathclap{X_i \in \{X_1, ..., X_K\}}} X_i^{(1)}\}\Big]
    + \mathbb{E}\Big[X X^\top \mathbbm{1}\{X = \argmin_{\mathclap{X_i \in \{X_1, ..., X_K\}}} X_i^{(1)}\}\Big] \right)\\
    &= C \left( \mathbb{E}\Big[X X^\top \mathbbm{1}\{X = \argmax_{\mathclap{X_i \in \{X_1, ..., X_K\}}} X_i^\top \tilde{\beta}\}\Big] + \mathbb{E}\Big[X X^\top \mathbbm{1}\{X = \argmin_{\mathclap{X_i \in \{X_1, ..., X_K\}}} X_i^\top \tilde{\beta}\}\Big] \right)
\end{align*}
which implies $C_\mathcal{X} = \mathcal{O}(1)$.
\endproof

\begin{lemma}
Consider i.i.d. arbitrary distribution $p_\mathcal{X}$.
Fix some vector $\beta \in \mathbb{R}^d$. For a given integer $k \in \{2,..., K-1\}$,
\begin{align*}
\mathbb{E}\left[ X_k X_k^\top  \mathbbm{1}\{ X_1^\top \beta < ... <  X_k^\top \beta <... < X_K^\top \beta \} \right]
\preccurlyeq  C_{K,k} \mathbb{E}\left[ ( X_1 X_1^\top + X_K X_K^\top) \mathbbm{1}\{ X_1^\top \beta < ... < X_K^\top \beta \} \right]
\end{align*}
where $C_{\mathcal{X}} = \binom{K-1}{(K-1)/2}$  assuming $K$ is odd --- if $K$ is even, we can use $\lceil (K-1)/2 \rceil$.
\end{lemma}

\proof{}
First notice that
\begin{align*}
&\mathbb{E}\left[ X_k X_k^\top  \mathbbm{1}\{ X_1^\top \beta < \cdots <  X_k^\top \beta <\cdots < X_K^\top \beta \} \right]\\
&= \mathbb{E}_V\left[ V V^\top  \mathbb{E}_{X_{1:K}/X_k}\left[ \mathbbm{1}\{ X_1^\top \beta < \cdots  < X_{k-1}^\top \beta  <  V^\top \beta < X_{k+1}^\top \beta < \cdots < X_K^\top \beta \} \mid V\right] \right]
\end{align*}
where $X_{1:K}/X_k$ denotes $X_1, ..., X_{k-1}, X_{k+1}, ..., X_K$.
Also, 
\begin{align*}
\mathbb{E}\left[ X_1 X_1^\top  \mathbbm{1}\{ X_1^\top \beta < \cdots < X_K^\top \beta \} \right]
&= \mathbb{E}_V\left[ V V^\top  \mathbb{E}_{X_{2:K}}\left[ \mathbbm{1}\{ V^\top \beta < X_2^\top \beta < \cdots < X_K^\top \beta \} \mid V\right] \right]\\
\mathbb{E}\left[ X_K X_K^\top  \mathbbm{1}\{ X_1^\top \beta < \cdots < X_K^\top \beta \} \right]
&= \mathbb{E}_V\left[ V V^\top  \mathbb{E}_{X_{1:K-1}}\left[ \mathbbm{1}\{ X_1^\top \beta < \cdots < X_{K-1}^\top \beta  < V^\top \beta \} \mid V\right] \right]
\end{align*}
Let $\psi(y) := \mathbb{P}(X^\top \beta \leq y)$ denote the CDF of $X^\top \beta$. Then
\begin{align*}
    &\mathbb{P}\left( X_1^\top \beta < \cdots  < X_{k-1}^\top \beta  <  V^\top \beta < X_{k+1}^\top \beta <\cdots < X_K^\top \beta \right)\\
    &= \prod_{i=1}^{k-1} \mathbb{P}\left( X_i^\top \beta \leq V^\top \beta \right) \frac{1}{(k-1)!} \prod_{i=k+1}^{N} \mathbb{P}\left( X_i^\top \beta \geq V^\top \beta \right) \frac{1}{(K-k)!}\\
    &= \frac{1}{(k-1)!(K-k)!} \psi(V^\top \beta)^{k-1} \left(1 - \psi(V^\top \beta) \right)^{K-k} .
\end{align*}
Likewise
\begin{align*}
    \mathbb{P}\left( V^\top \beta < X_2^\top \beta <  \cdots < X_K^\top \beta \right)
    &= \frac{1}{(K-1)!}  \left(1 - \psi(V^\top \beta) \right)^{K-1} ,\\
    \mathbb{P}\left(  X_1^\top \beta <  \cdots < X_{K-1}^\top \beta < V^\top \beta \right)
    &= \frac{1}{(K-1)!}  \psi(V^\top \beta)^{K-1} .
\end{align*}
Then, we need to show there exists $C_{K,k}$ such that
\begin{align*}
&\mathbb{P}\left( X_1^\top \beta < \cdots  < X_{k-1}^\top \beta  <  V^\top \beta < X_{k+1}^\top \beta <\cdots < X_K^\top \beta \right)\\
&\leq
    C_{K,k} \left[\mathbb{P}\left( V^\top \beta < X_2^\top \beta <  \cdots < X_K^\top \beta \right)
    + \mathbb{P}\left(  X_1^\top \beta <  \cdots < X_{K-1}^\top \beta < V^\top \beta \right)
    \right] \,.
\end{align*}
That is,
\begin{align*}
     \frac{ \psi(V^\top \beta)^{k-1} \left(1 - \psi(V^\top \beta) \right)^{K-k}}{(k-1)!(K-k)!} \leq  \frac{C_{K,k}}{(K-1)!} \left[  \left(1 - \psi(V^\top \beta) \right)^{K-1} + \psi(V^\top \beta)^{K-1}  \right] .
\end{align*}
Hence,
\begin{align*}
    C_{K,k} \geq
    \binom{K-1}{k-1} \frac{\psi(V^\top \beta)^{k-1} \left(1 - \psi(V^\top \beta) \right)^{K-k}}{ \left(1 - \psi(V^\top \beta) \right)^{K-1} + \psi(V^\top \beta)^{K-1} } \,.
\end{align*}
Since $\psi(V^\top \beta) \in [0, 1]$, we have
\begin{align*}
    \frac{\psi(V^\top \beta)^{k-1} \left(1 - \psi(V^\top \beta) \right)^{K-k}}{ \left(1 - \psi(V^\top \beta) \right)^{K-1} + \psi(V^\top \beta)^{K-1} } \leq 1
\end{align*}
for all $K$ and $k$.
Hence, for $C_{K,k} = \binom{K-1}{k-1}$,
\begin{align*}
&\mathbb{E}\left[ X_k X_k^\top  \mathbbm{1}\{ X_1^\top \beta < \cdots <  X_k^\top \beta <\cdots < X_K^\top \beta \} \right]\\
&\preccurlyeq  C_{K,k} \mathbb{E}\left[ ( X_1 X_1^\top + X_K X_K^\top) \mathbbm{1}\{ X_1^\top \beta < \cdots < X_K^\top \beta \} \right] .
\end{align*}
\endproof

\section{Other lemmas}
\begin{lemma}[\citet{wainwright2019high}, Theorem~2.19]\label{lemma:bernstein}
Let $\{Z_\tau, \mathcal{F}_\tau\}_\tau^\infty$ be a martingale difference sequence, and suppose that $Z_\tau$ is $\sigma^2$-sub-Gaussian in an adapted sense, i.e., for all $\alpha \in \mathbb{R}$, $\mathbb{E}[e^{\alpha Z_\tau} | \mathcal{F}_{\tau-1}] \leq e^{\alpha^2 \sigma^2/2}$ almost surely. Then for all $\gamma \geq 0$, $\mathbb{P}\left[ |\sum_{\tau=1}^n Z_\tau | \geq \gamma \right] \leq 2 \exp[-\gamma^2/(2n\sigma^2)]$.
\end{lemma}

Note that Lemma~\ref{lemma:conditional_gaussian} is a well-known result, but for the sake of completeness, we present its formal statment and proof.

\begin{lemma}\label{lemma:conditional_gaussian}
Let $X \in \mathbb{R}^d$ follow a multivariate Gaussian distribution with mean $\mu$ and covarance matrix $\Sigma$ and consider the partition of $X$ with
\begin{align*}
X =
\begin{bmatrix}
X_1\\
X_2
\end{bmatrix} &\sim  
\mathcal{N}
\begin{pmatrix}
\begin{bmatrix}
\mu_1\\
\mu_2
\end{bmatrix}\!\!,&
\begin{bmatrix}
\Sigma_{11} & \Sigma_{12} \\
\Sigma_{21} & \Sigma_{22} 
\end{bmatrix}
\end{pmatrix} .
\end{align*}
Then the conditional distribution of $X_1$ given $X_2$ is also a multivariate Gaussian distribution. In particular
\begin{align*}
    X_1 \mid X_2 = x_2 \sim \mathcal{N}\left( \mu_1 + \Sigma_{12} \Sigma^{-1}_{22} (x_2 - \mu_2), \Sigma_{11} - \Sigma_{12} \Sigma^{-1}_{22} \Sigma_{21} \right) .
\end{align*}
\end{lemma}

\proof{}
Define $Z = X_1 + {\bf A} X_2$ where ${\bf A} = -\Sigma_{12} \Sigma^{-1}_{22}$. Now we can write
\begin{align*} 
{\rm cov}(Z, X_2) &= {\rm cov}( X_{1}, X_2 ) + 
{\rm cov}({\bf A}X_2, X_2) \\
&= \Sigma_{12} + {\bf A} {\rm var}(X_2) \\
&= \Sigma_{12} - \Sigma_{12} \Sigma^{-1}_{22} \Sigma_{22} \\
&= 0
\end{align*}
Therefore $Z$ and $X_2$ are not correlated and, since they are jointly normal, they are independent\footnote{If a random vector has a multivariate normal distribution then any two or more of its components that are uncorrelated are independent.}. Now, clearly we have $\mathbb{E}(Z) = \mu_1 + {\bf A}\mu_2$. Then
\begin{align*}
\mathbb{E}[X_1 | X_2] &= \mathbb{E}[ Z - {\bf A} X_2 | X_2] \\
& = \mathbb{E}[Z|X_2] -  \mathbb{E}[{\bf A}X_2|X_2] \\
& = \mathbb{E}[Z] - {\bf A}X_2 \\
& = \mu_1 + {\bf A}  (\mu_2 - X_2) \\
& = \mu_1 + \Sigma_{12} \Sigma^{-1}_{22} (X_2- \mu_2) .
\end{align*}
For the covariance matrix, note that
\begin{align*}
{\rm var}(X_1|X_2) &= {\rm var}(Z - {\bf A} X_2 | X_2) \\
&= {\rm var}(Z|X_2) + {\rm var}({\bf A} X_2 | X_2) - {\bf A}{\rm cov}(Z, -X_2) - {\rm cov}(Z, -X_2) {\bf A}^\top \\
&= {\rm var}(Z|X_2) \\
&= {\rm var}(Z)
\end{align*}
Hence, it follows that
\begin{align*}
{\rm var}(X_1|X_2) 
&= {\rm var}( Z )\\
&= {\rm var}( X_1 + {\bf A} X_2 ) \\
&= {\rm var}( X_1 ) + {\bf A} {\rm var}( X_2 ) {\bf A}^\top
+ {\bf A} {\rm cov}(X_1,X_2) + {\rm cov}(X_2,X_1) {\bf A}^\top \\
&= \Sigma_{11} +\Sigma_{12} \Sigma^{-1}_{22} \Sigma_{22}\Sigma^{-1}_{22}\Sigma_{21}
- 2 \Sigma_{12} \Sigma_{22}^{-1} \Sigma_{21} \\
&= \Sigma_{11} +\Sigma_{12} \Sigma^{-1}_{22}\Sigma_{21}
- 2 \Sigma_{12} \Sigma_{22}^{-1} \Sigma_{21} \\
&= \Sigma_{11} -\Sigma_{12} \Sigma^{-1}_{22}\Sigma_{21}
\end{align*}
\endproof

\section{Additional Experiment Results}\label{sec:additional_numerical_experiments}

\subsection{Details on Experimental Setup}

For feature vectors drawn from the uniform distribution, we sample each feature vector $X$ independently from a $d$-dimensional hypercube $[-1,1]^d$. For elliptically distributed feature vectors, we construct each feature vector $X \in \mathbb{R}^d$ following the definition in Theorem~1 of \citet{cambanis1981theory}:
\begin{align*}
    X = \mu + R A U^{(k)}
\end{align*}
where $\mu \in \mathbb{R}^d$ is a mean vector, $U^{(k)} \in \mathbb{R}^k$ is uniformly distributed on the unit sphere in~$\mathbb{R}^k$, $R \in \mathbb{R}$ is a random variable independent of $U^{(k)}$, and $A$ is a $d \times k$-dimensional matrix with rank $k$. We sample $R$ from Gaussian distribution $\mathcal{N}(0,1)$, and sample each element of $A$ uniformly in $[0,1]$. We use zero mean $\mu = \vec{0}_d$.

\subsection{Additional Results for Two-Armed Bandits}

\begin{figure*}[ht]
    \begin{subfigure}[b]{0.32\textwidth}
        \includegraphics[width=\textwidth]{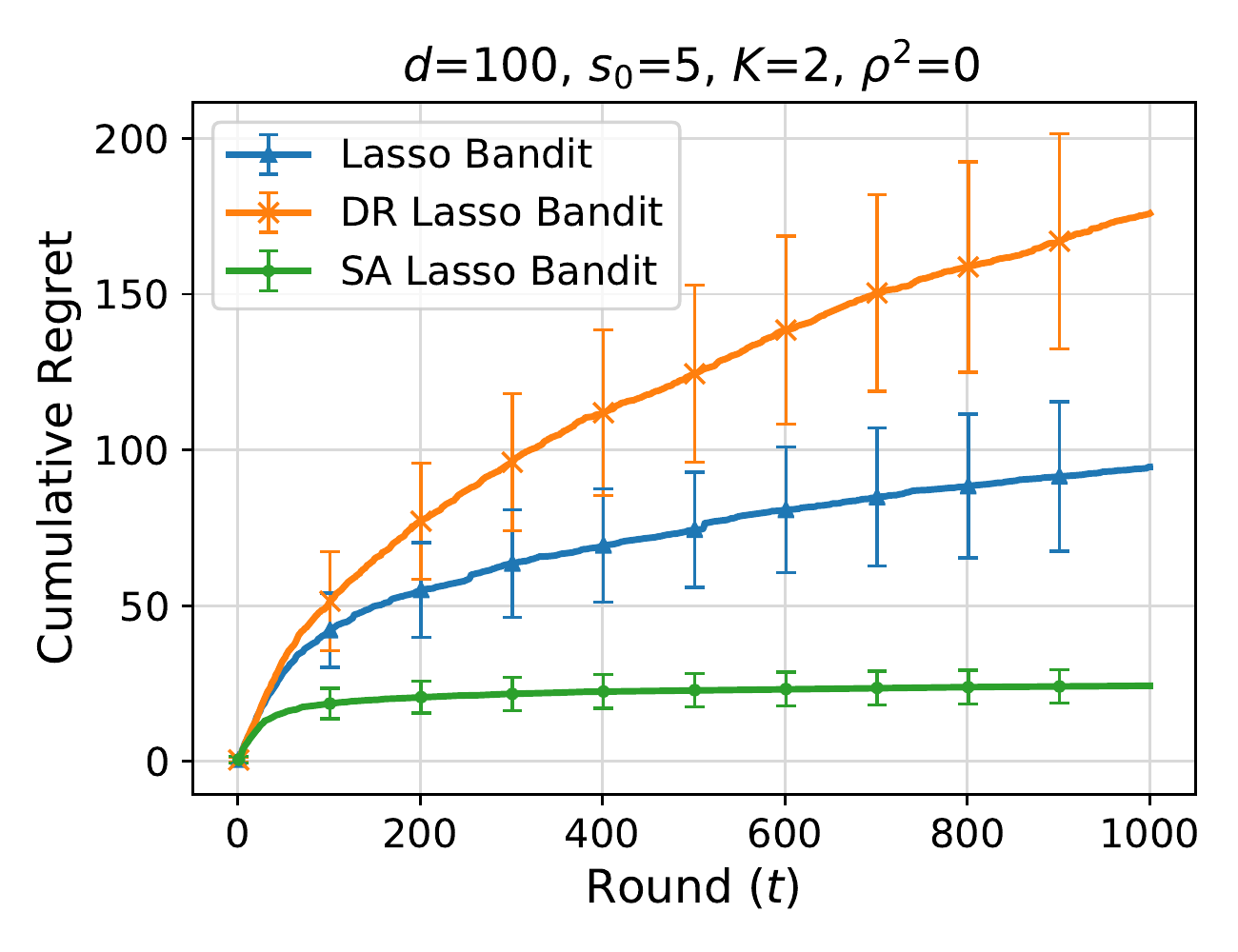}
    \end{subfigure}
    \begin{subfigure}[b]{0.32\textwidth}
        \includegraphics[width=\textwidth]{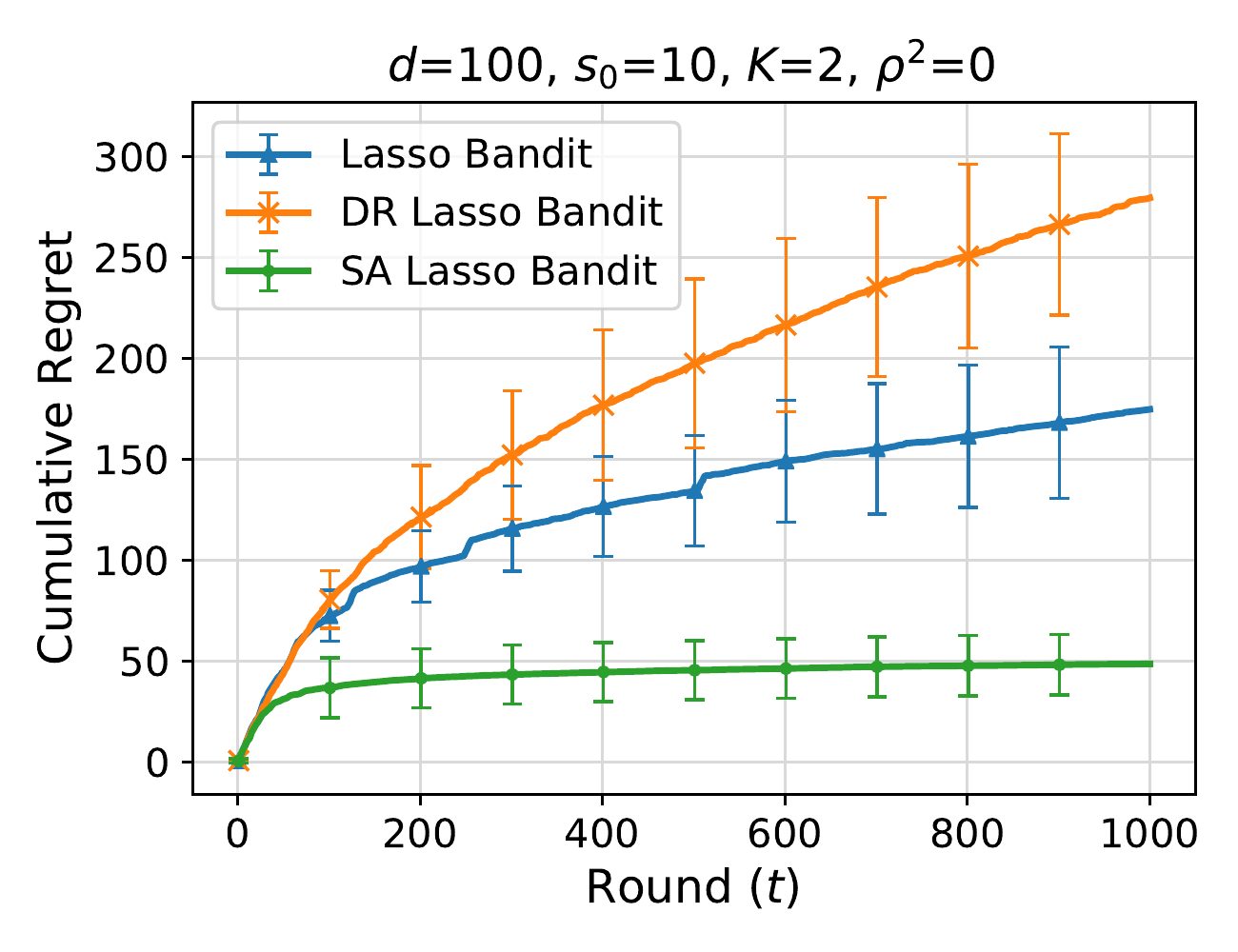}
    \end{subfigure}
    \begin{subfigure}[b]{0.32\textwidth}
        \includegraphics[width=\textwidth]{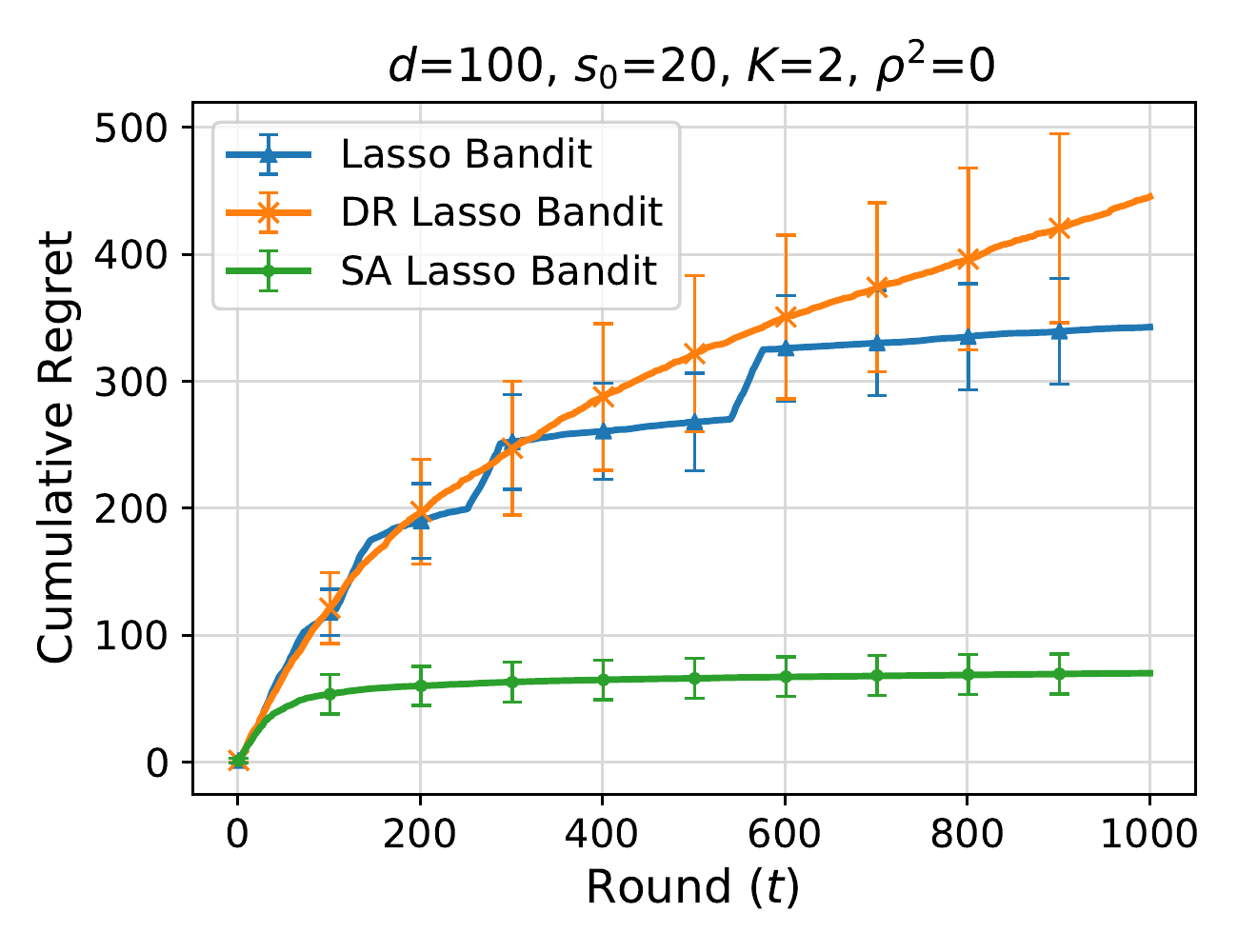}
    \end{subfigure}\\
    \begin{subfigure}[b]{0.32\textwidth}
        \includegraphics[width=\textwidth]{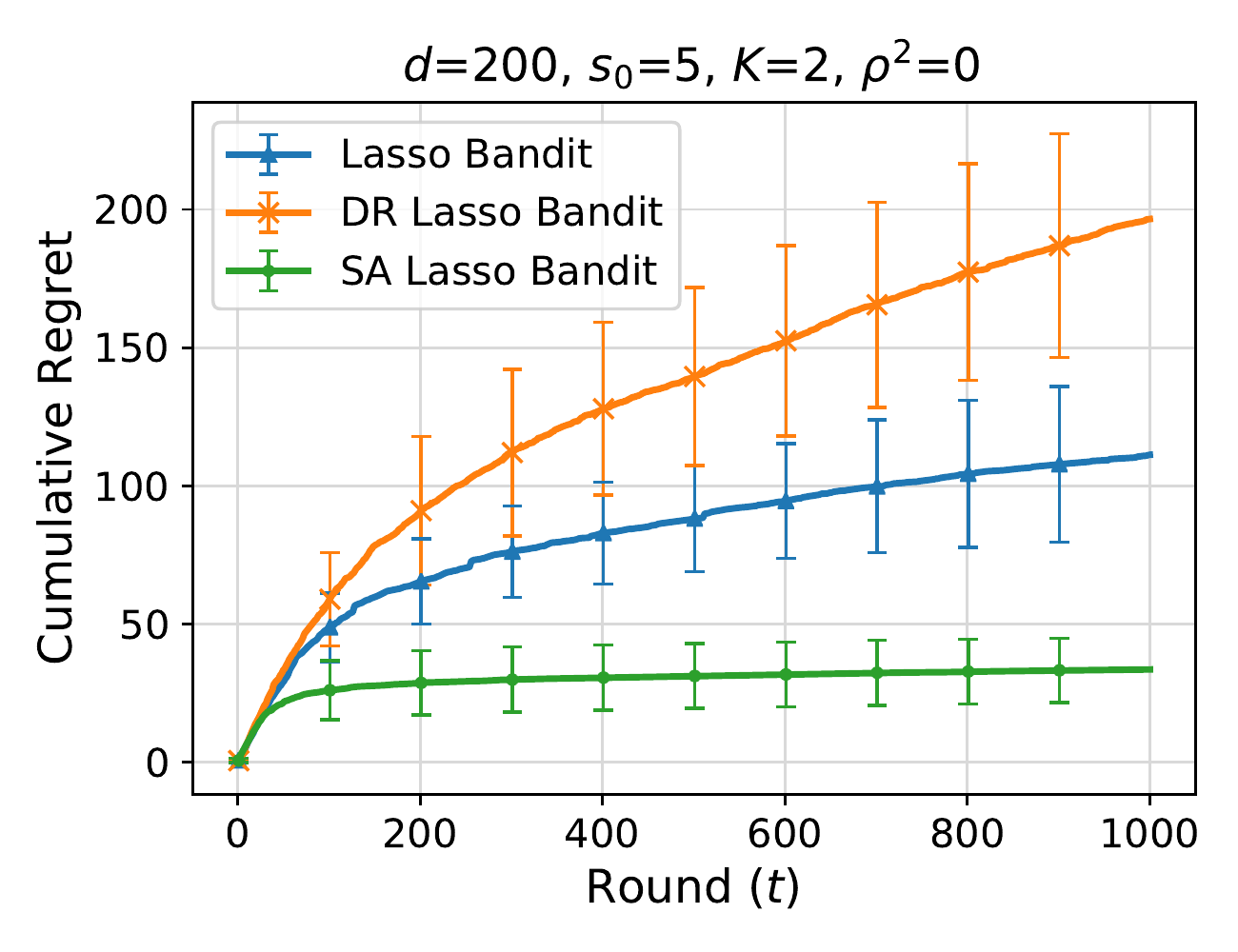}
    \end{subfigure}
    \begin{subfigure}[b]{0.32\textwidth}
        \includegraphics[width=\textwidth]{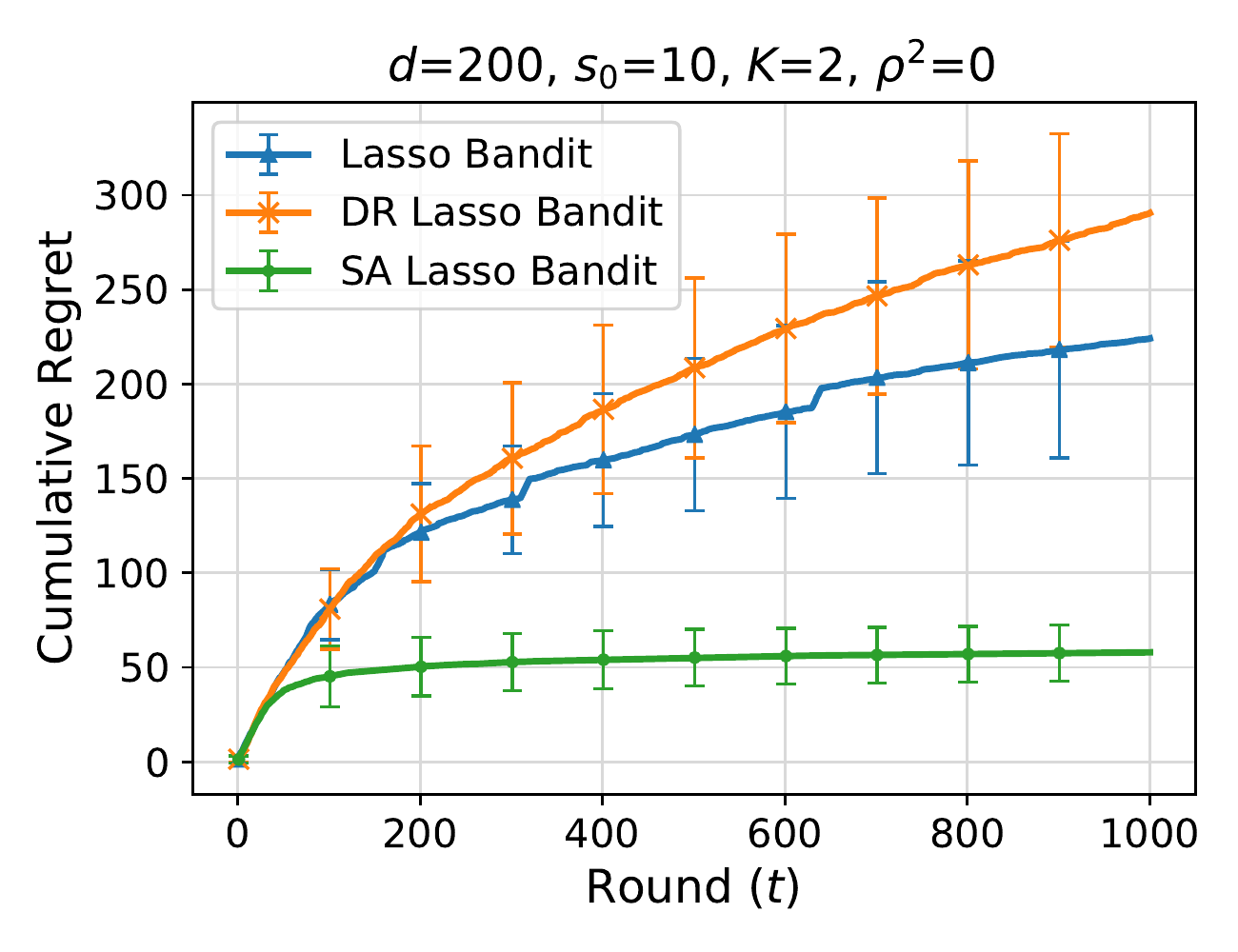}
    \end{subfigure}
    \begin{subfigure}[b]{0.32\textwidth}
        \includegraphics[width=\textwidth]{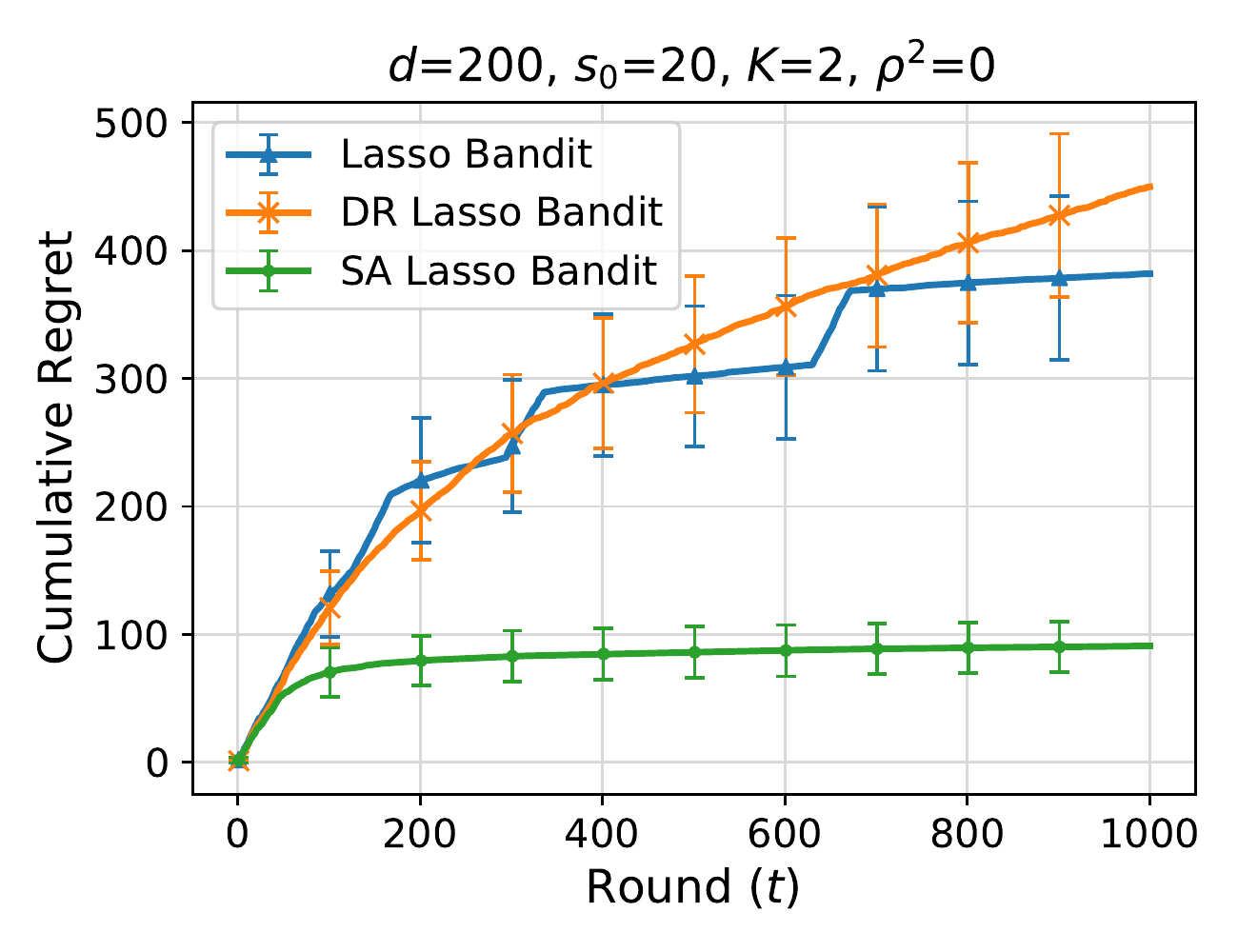}
    \end{subfigure}
    \caption{\small The plots show the $t$-round cumulative regret of \textsc{SA Lasso Bandit} (Algorithm~\ref{algo:SA_Lasso_bandit}), \textsc{DR Lasso Bandit} \citep{kim2019doubly}, and \textsc{Lasso Bandit} \citep{bastani2020online} for $K = 2$, $d \in \{100, 200\}$ and varying sparsity $s_0 \in \{5, 10, 20\}$ under no correlation between arms, $\rho^2 = 0$.} 
     \label{fig:two_arm_no_corr}
\end{figure*}

Figure~\ref{fig:two_arm_no_corr} shows the evaluations in two-armed bandits with independent arms whose features are drawn from a multivariate Gaussian distribution.
Comparing the numerical results in Figure~\ref{fig:two_arm_no_corr} with those in Figure~\ref{fig:two_arm_strong_corr} and Figure~\ref{fig:two_arm_weak_corr}, we observe that the performance of \textsc{DR Lasso Bandit} substantially deteriorates as correlation between arms decreases whereas the performances of \textsc{SA Lasso Bandit} and \textsc{Lasso Bandit} decrease more gracefully with a decrease in arm correlation. Throughout these experiments, our proposed algorithm, \textsc{SA Lasso Bandit}, consistently exhibits the fastest convergence to the optimal action  and robust performances under various instances.


\newpage

\subsection{Additional Results for K-Armed Bandits}

\begin{figure*}[ht]
    \begin{subfigure}[b]{0.24\textwidth}
        \includegraphics[width=\textwidth]{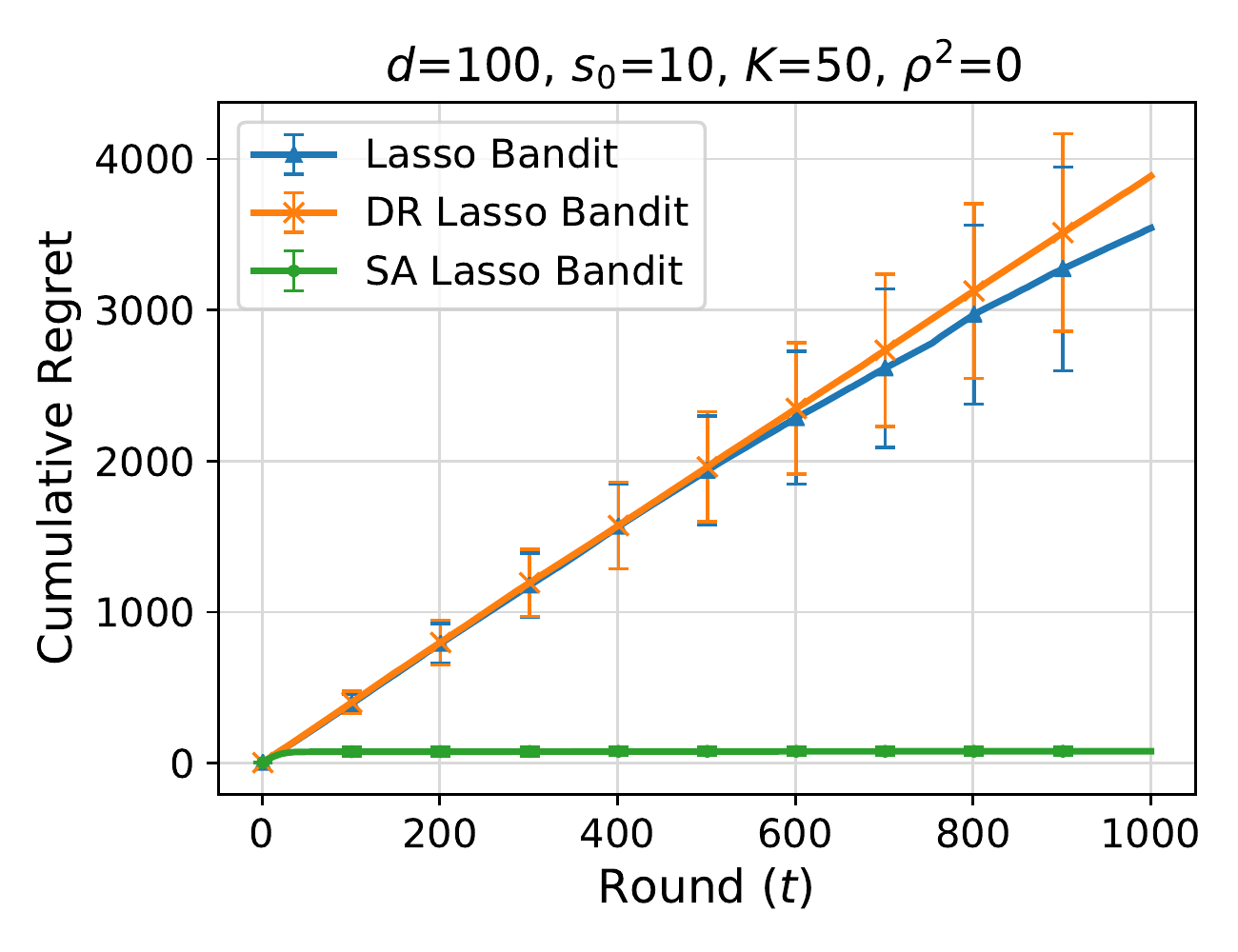}
    \end{subfigure}
    \begin{subfigure}[b]{0.24\textwidth}
        \includegraphics[width=\textwidth]{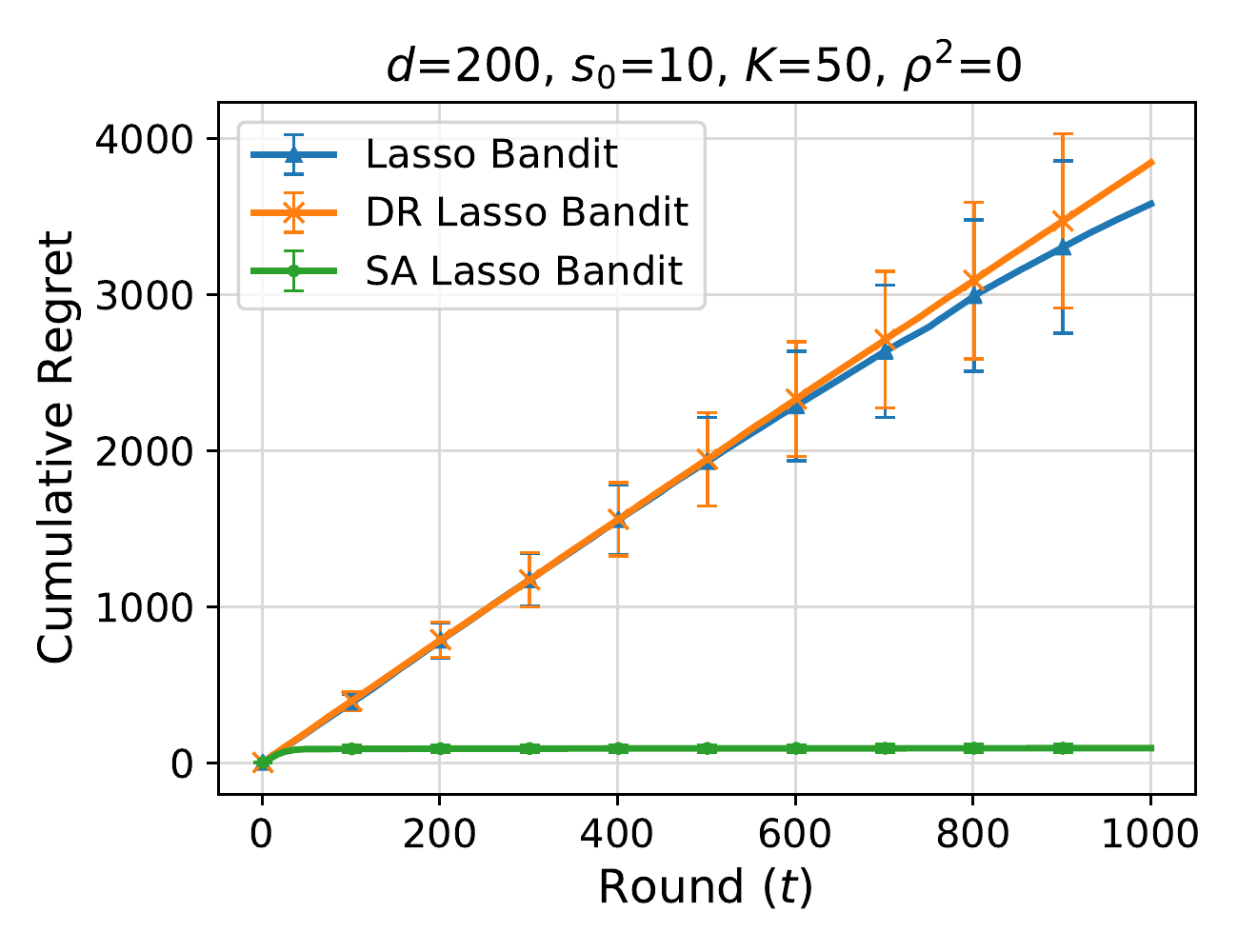}
    \end{subfigure}
    \begin{subfigure}[b]{0.24\textwidth}
        \includegraphics[width=\textwidth]{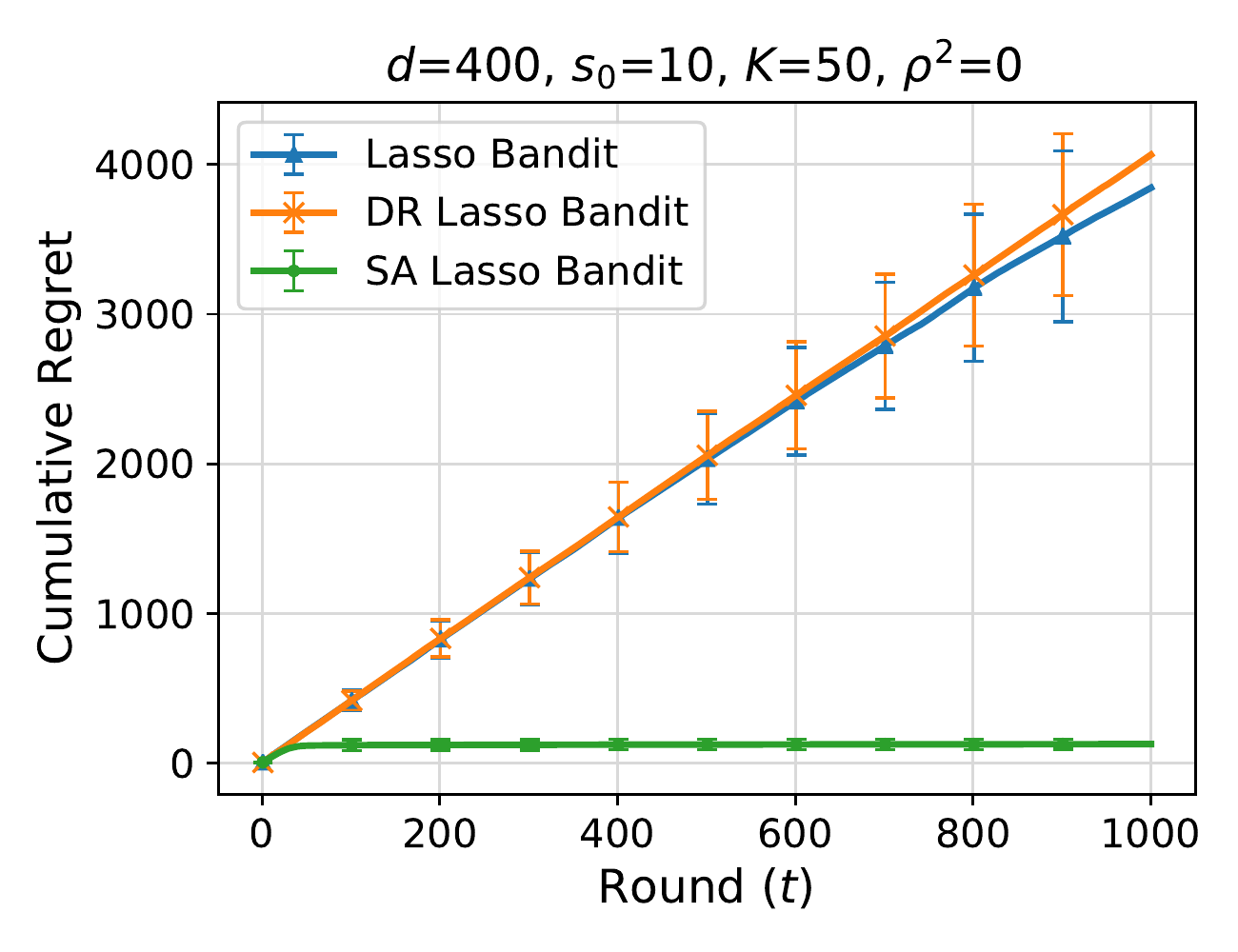}
    \end{subfigure}
    \begin{subfigure}[b]{0.24\textwidth}
        \includegraphics[width=\textwidth]{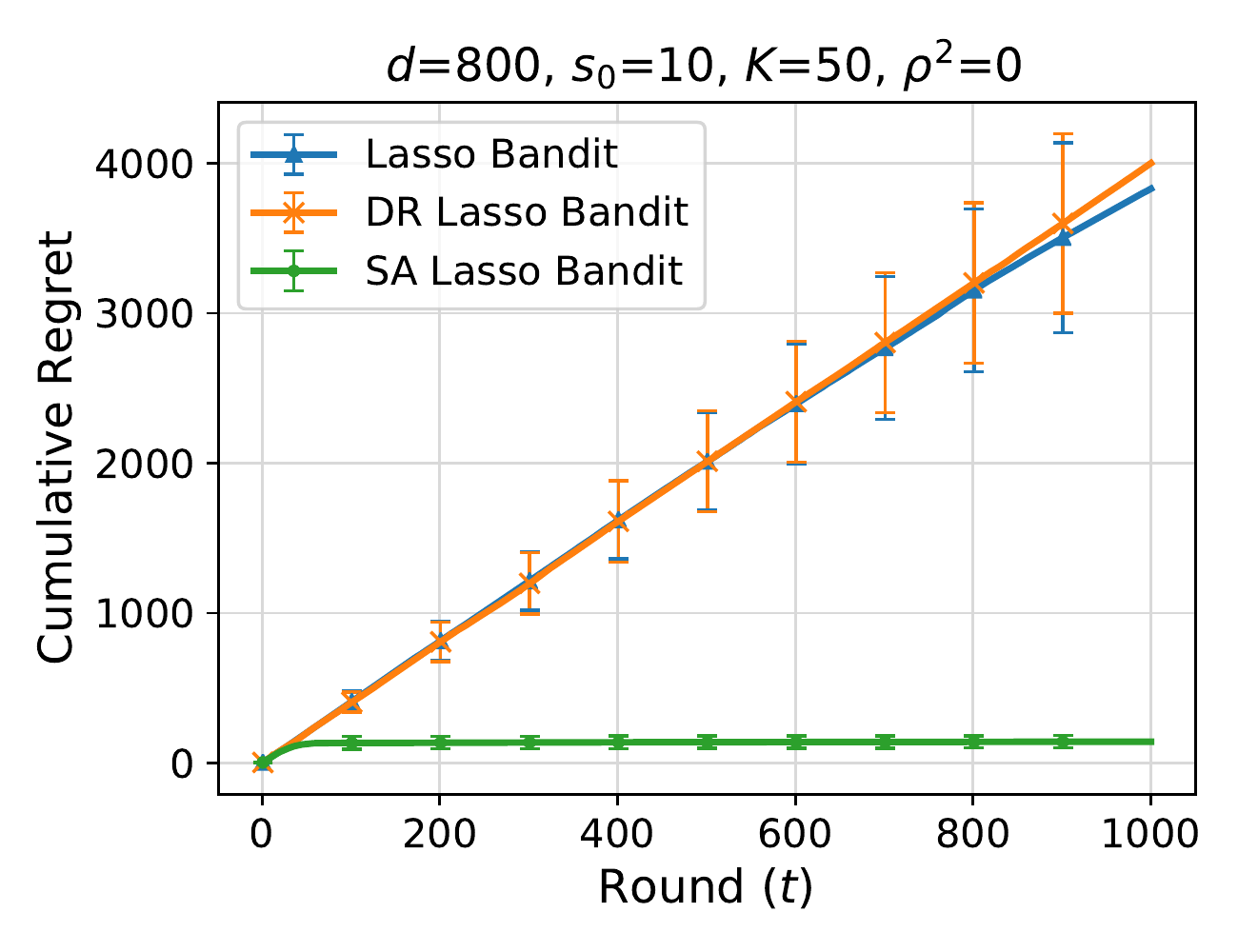}
    \end{subfigure}\\
    \begin{subfigure}[b]{0.24\textwidth}
        \includegraphics[width=\textwidth]{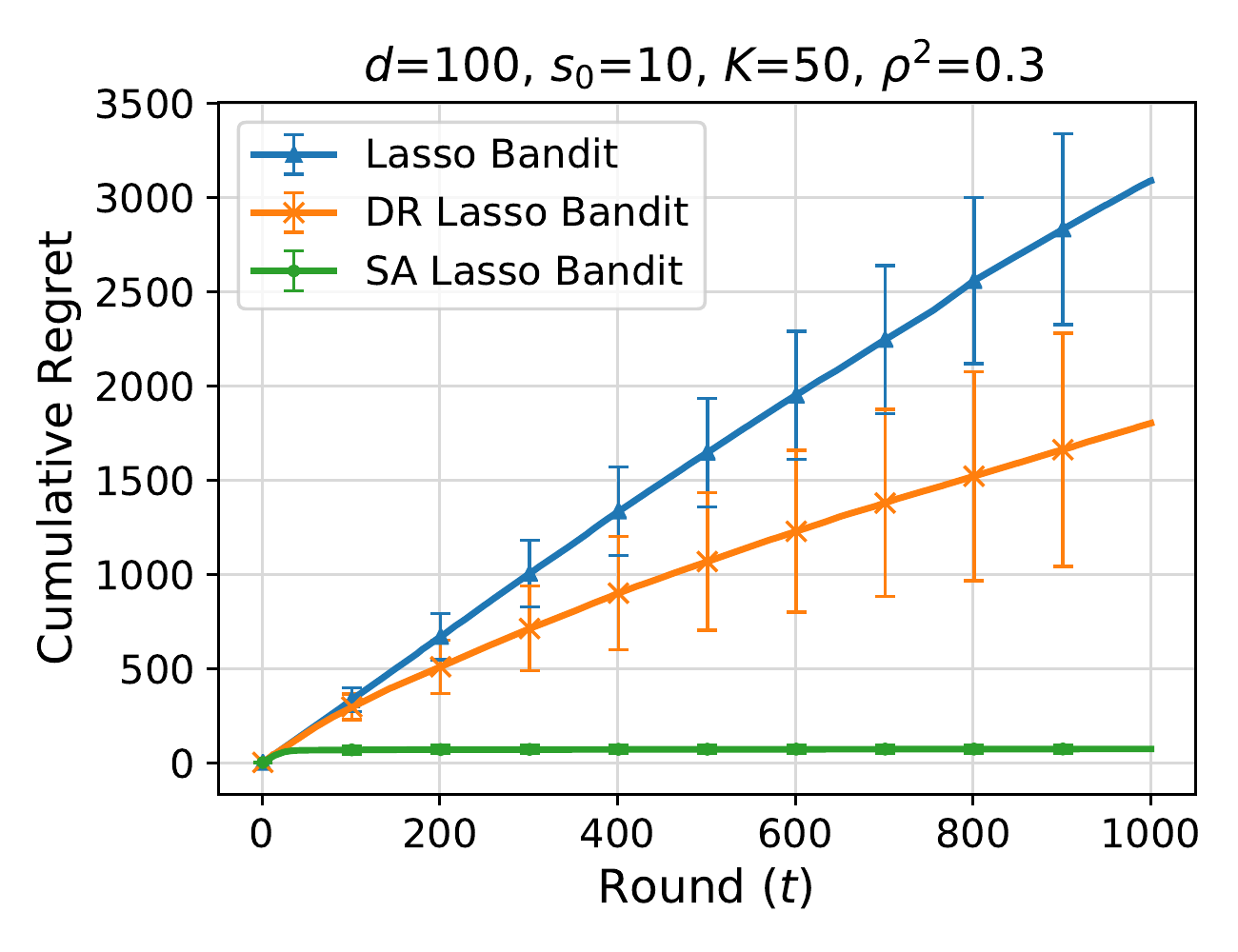}
    \end{subfigure}
    \begin{subfigure}[b]{0.24\textwidth}
        \includegraphics[width=\textwidth]{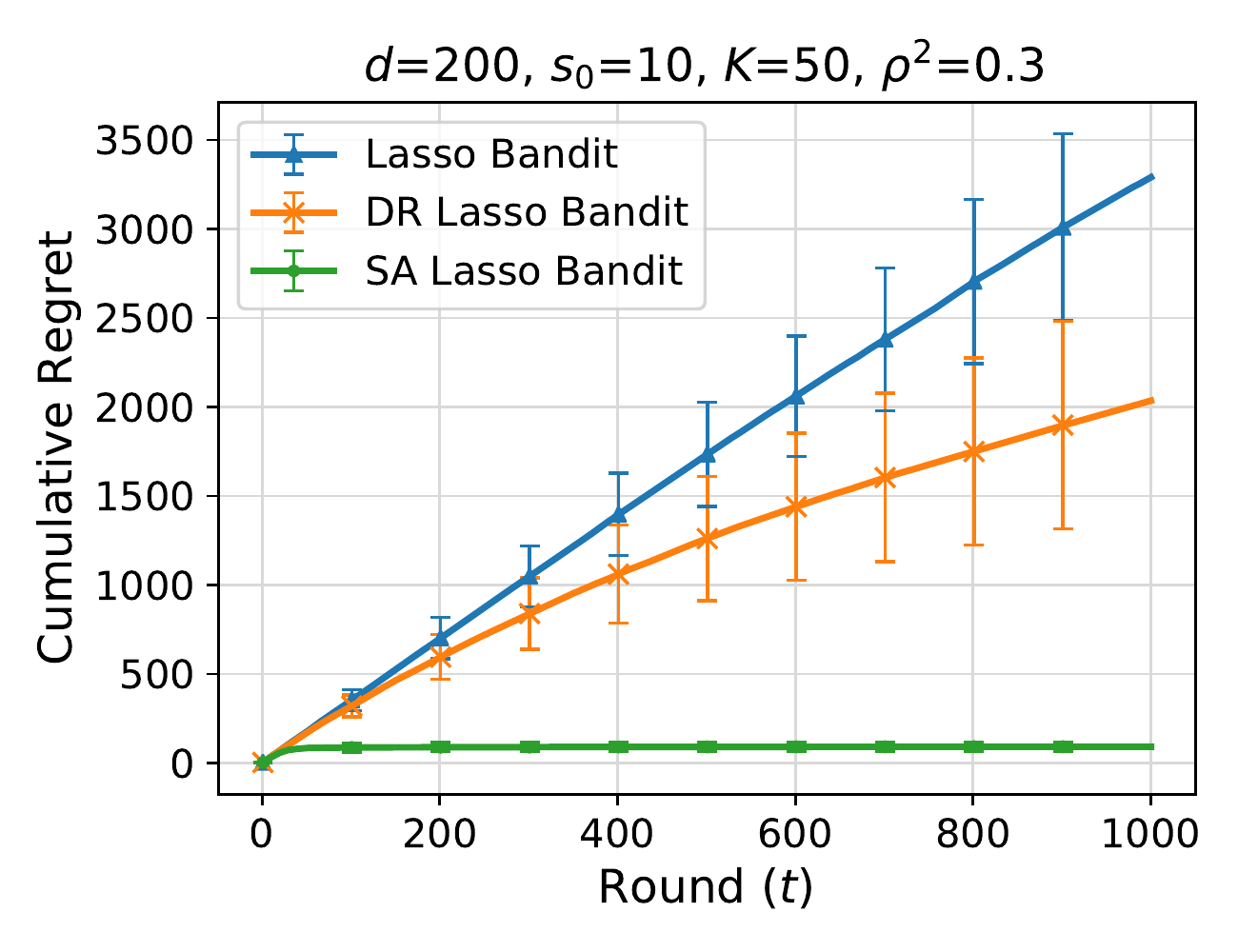}
    \end{subfigure}
    \begin{subfigure}[b]{0.24\textwidth}
        \includegraphics[width=\textwidth]{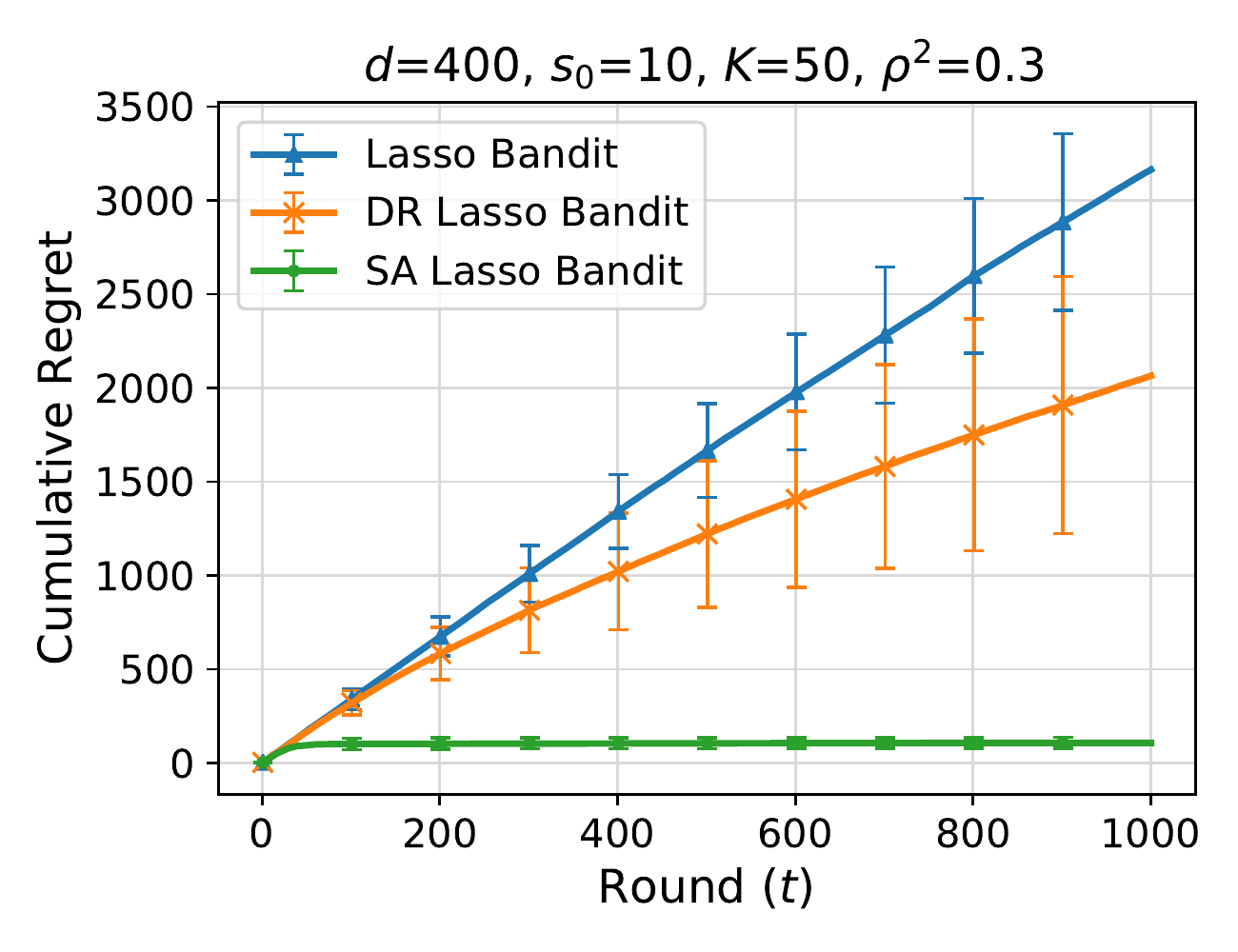}
    \end{subfigure}
    \begin{subfigure}[b]{0.24\textwidth}
        \includegraphics[width=\textwidth]{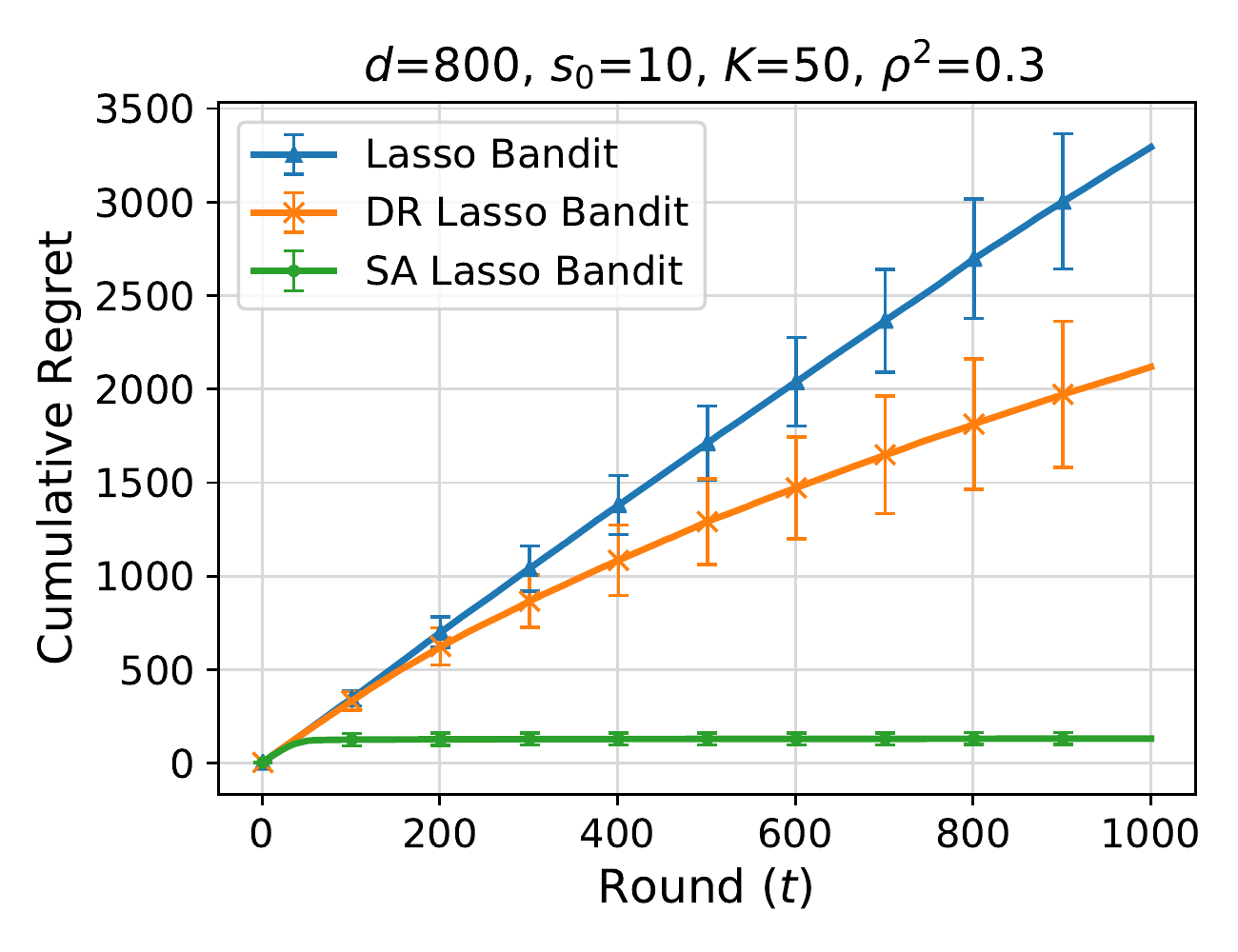}
    \end{subfigure}\\
    \begin{subfigure}[b]{0.24\textwidth}
        \includegraphics[width=\textwidth]{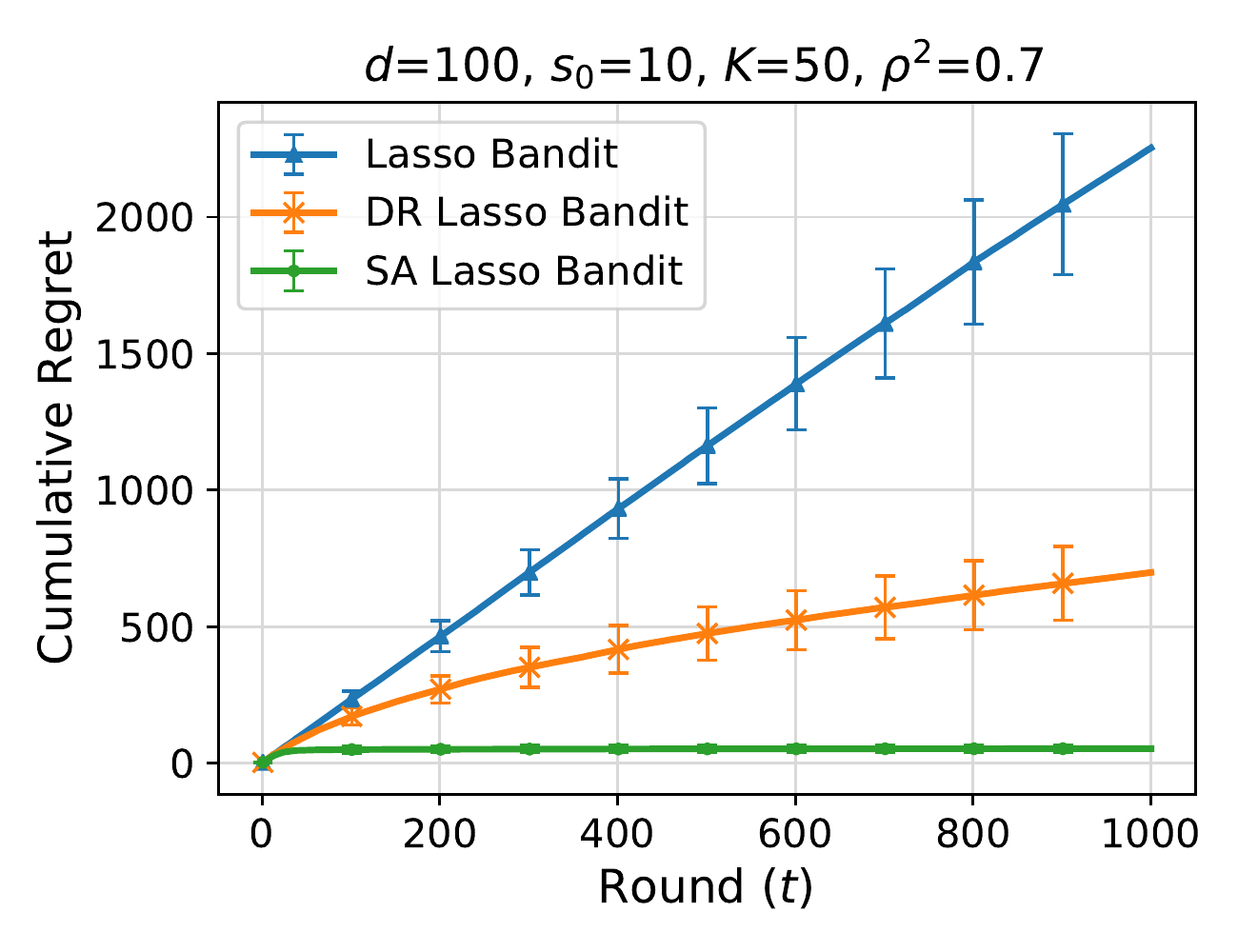}
    \end{subfigure}
    \begin{subfigure}[b]{0.24\textwidth}
        \includegraphics[width=\textwidth]{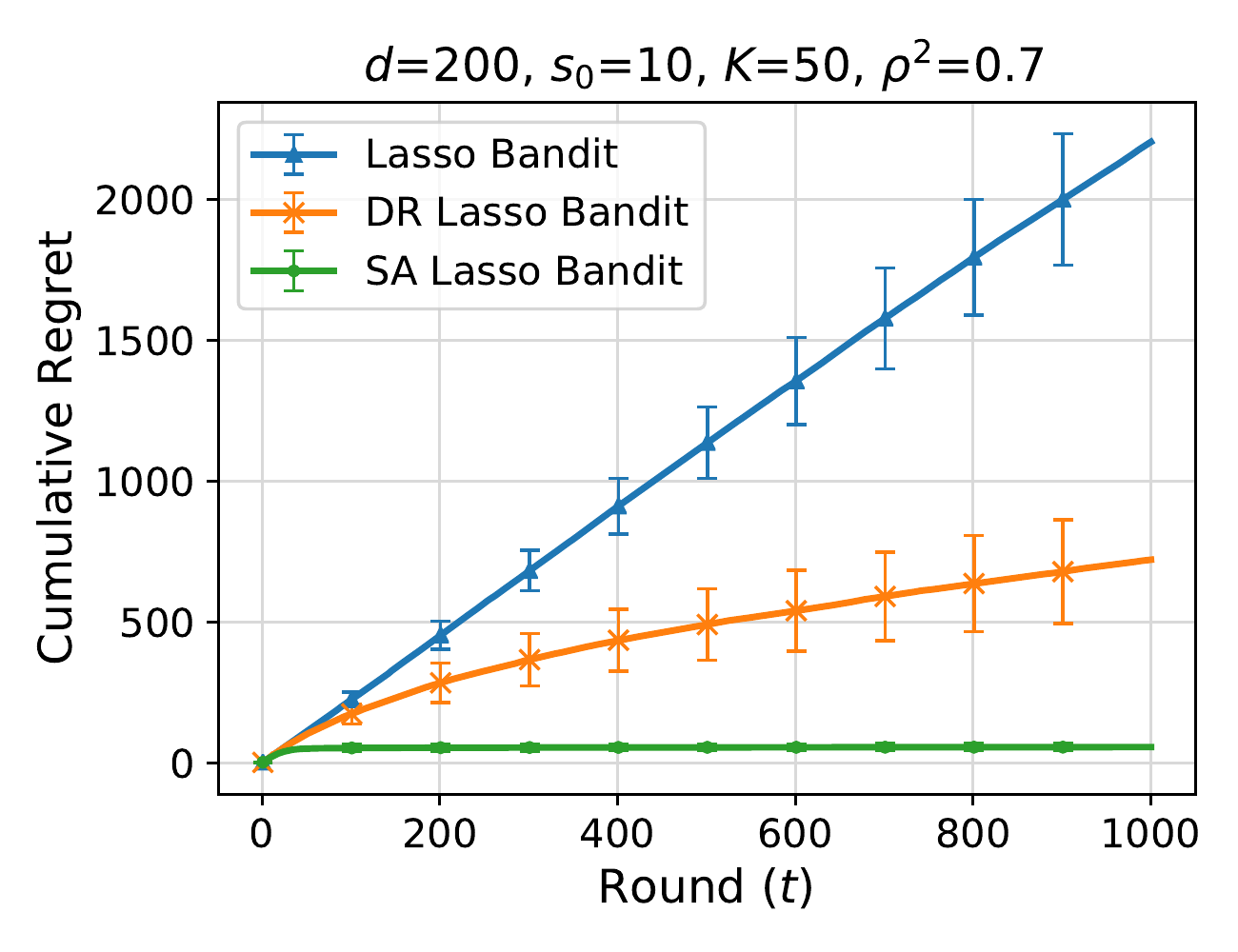}
    \end{subfigure}
    \begin{subfigure}[b]{0.24\textwidth}
        \includegraphics[width=\textwidth]{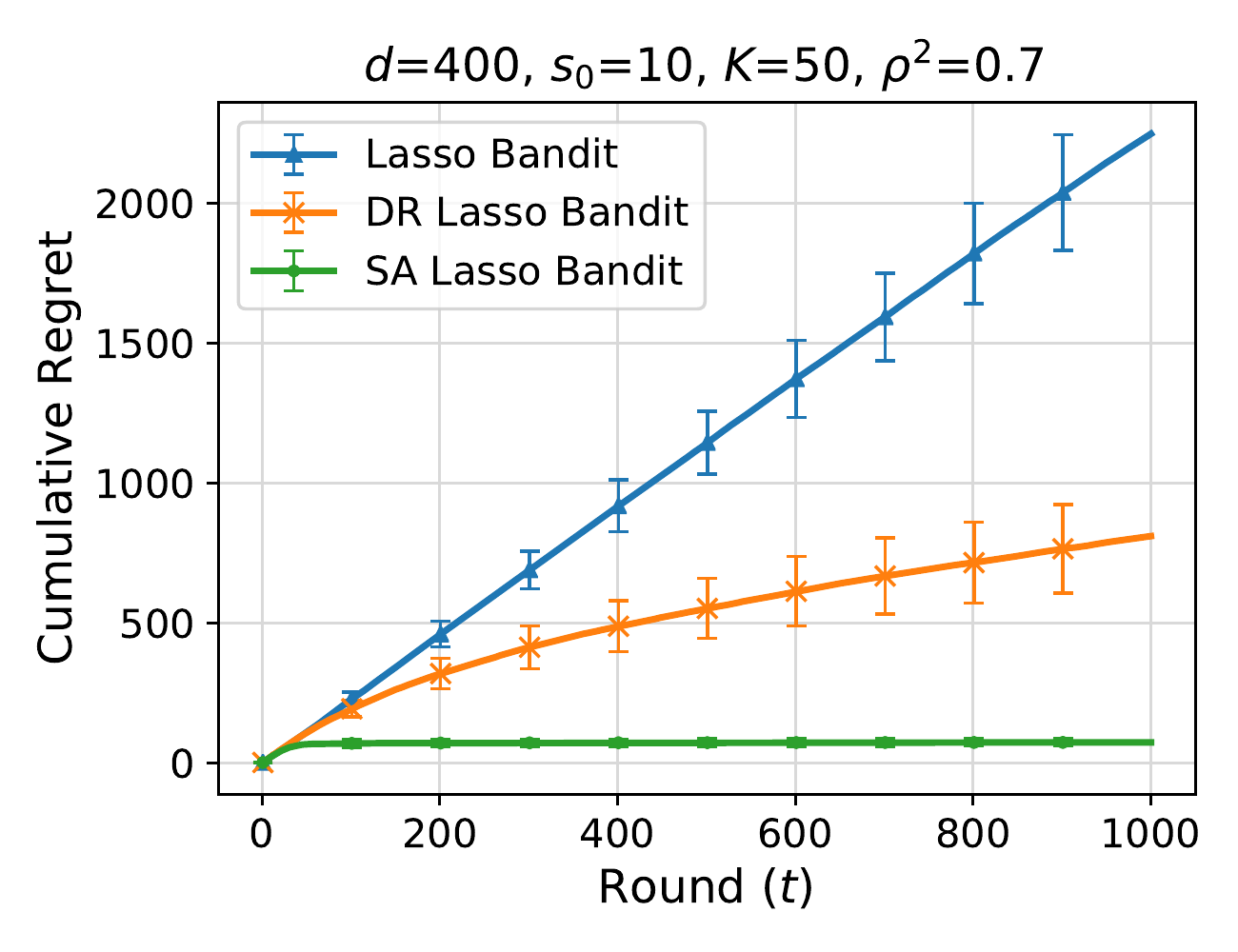}
    \end{subfigure}
    \begin{subfigure}[b]{0.24\textwidth}
        \includegraphics[width=\textwidth]{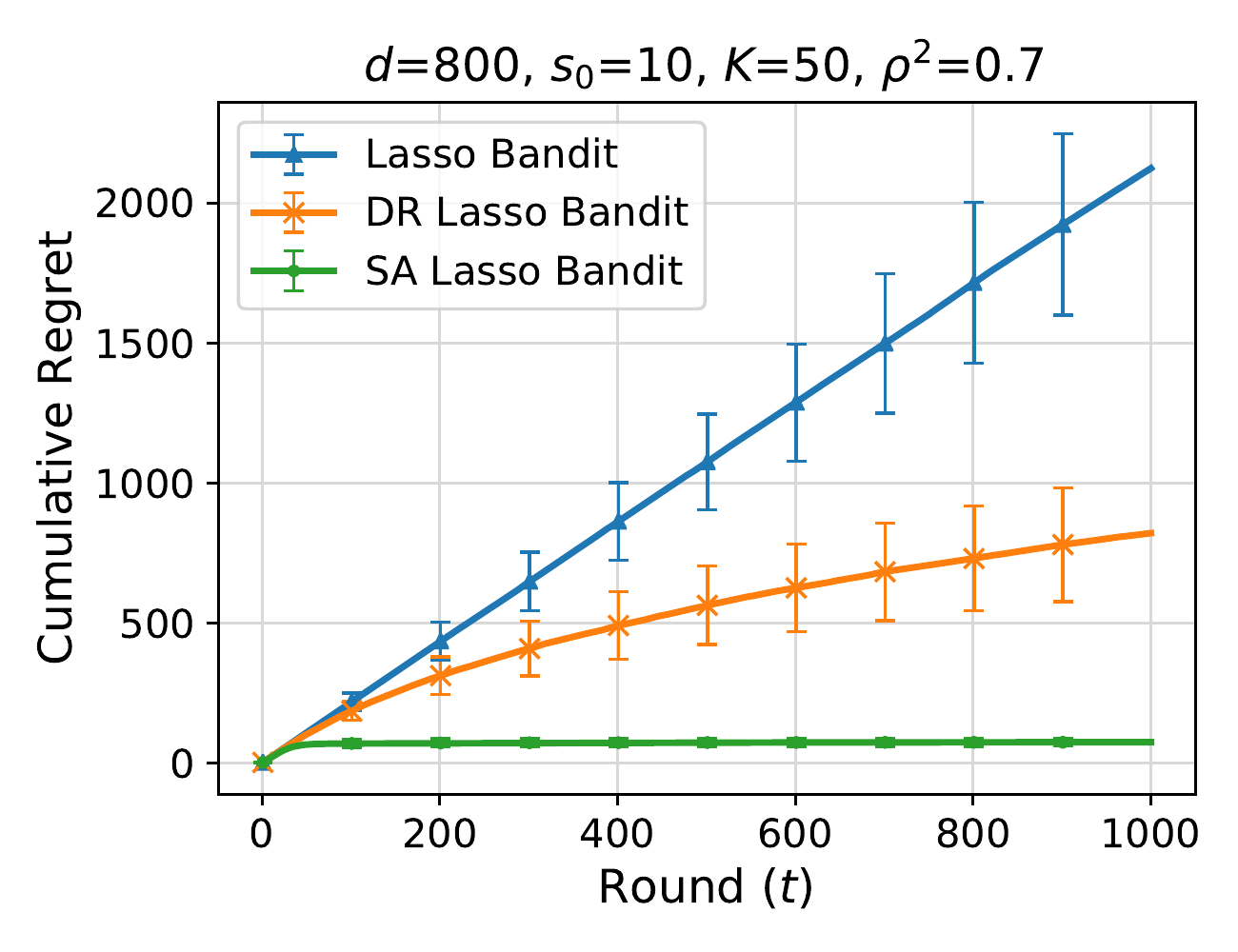}
    \end{subfigure}\\
        \begin{subfigure}[b]{0.24\textwidth}
        \includegraphics[width=\textwidth]{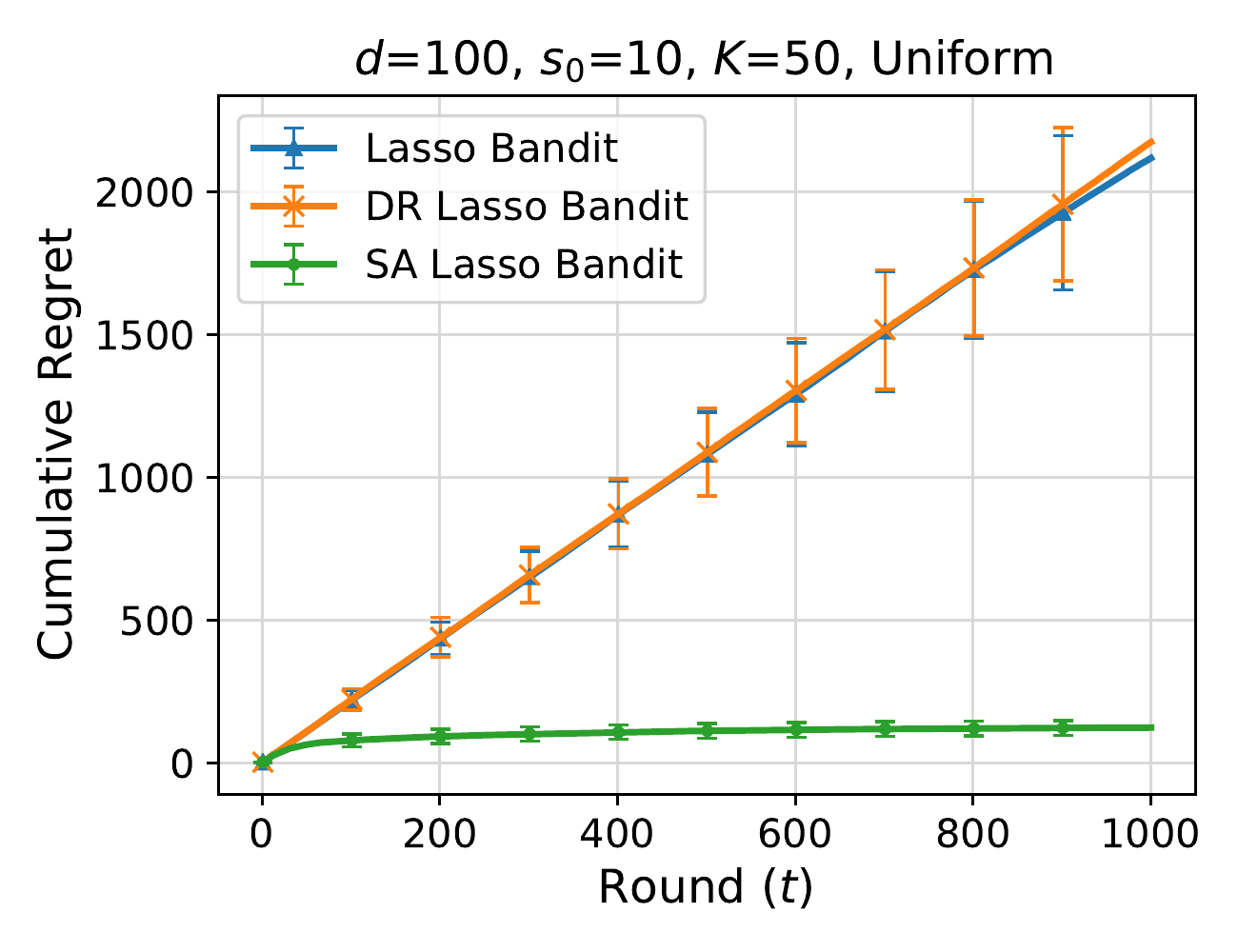}
    \end{subfigure}
    \begin{subfigure}[b]{0.24\textwidth}
        \includegraphics[width=\textwidth]{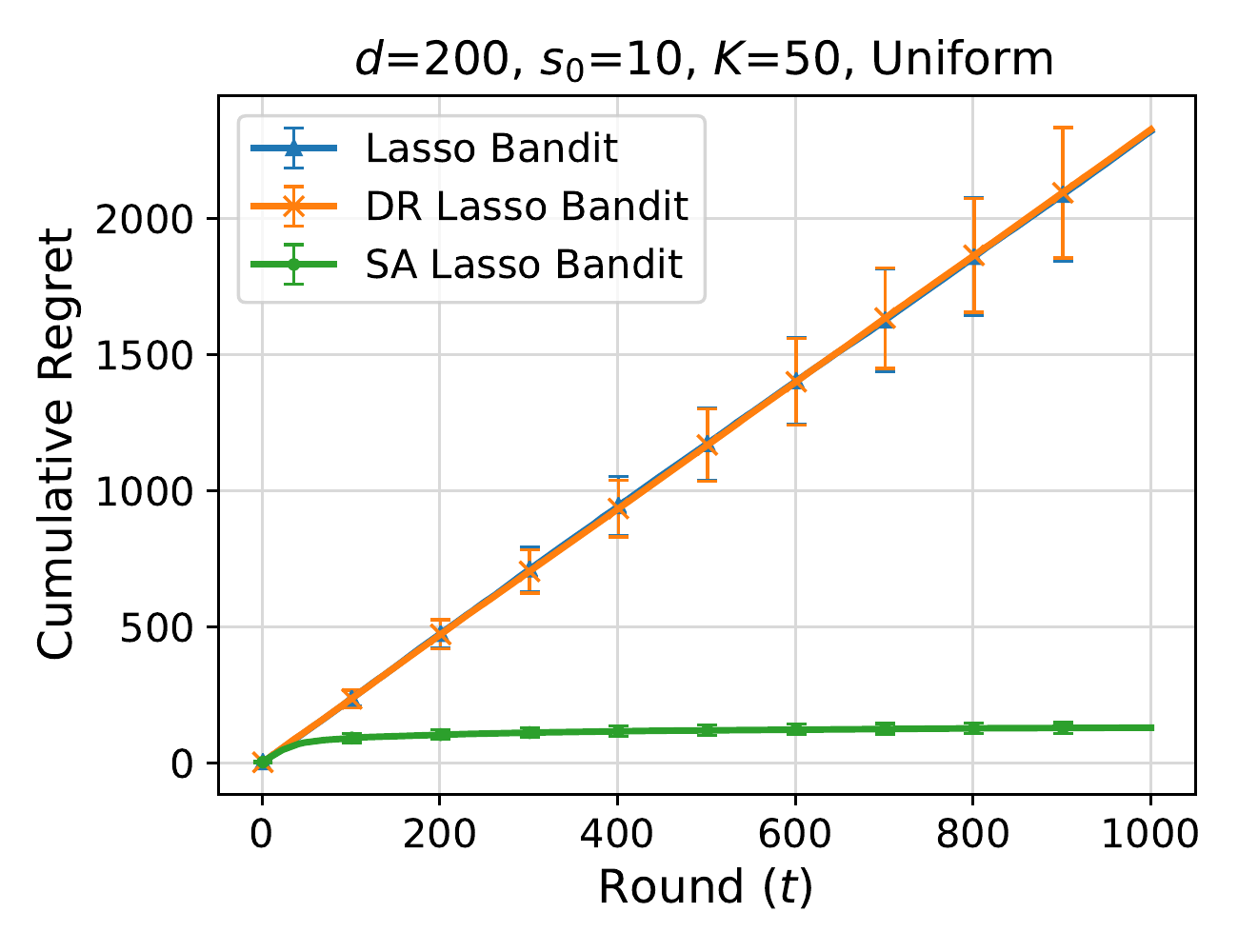}
    \end{subfigure}
    \begin{subfigure}[b]{0.24\textwidth}
        \includegraphics[width=\textwidth]{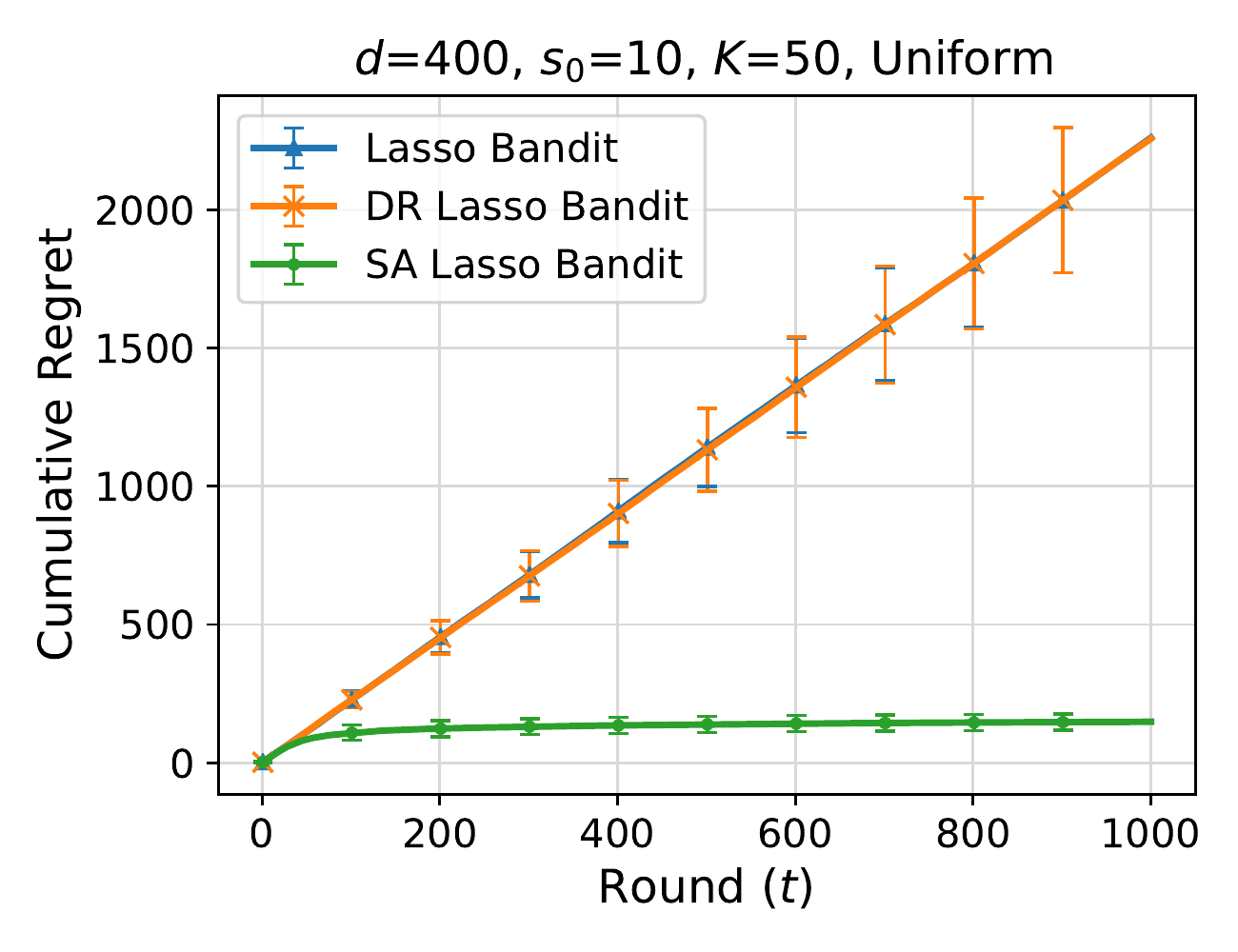}
    \end{subfigure}
    \begin{subfigure}[b]{0.24\textwidth}
        \includegraphics[width=\textwidth]{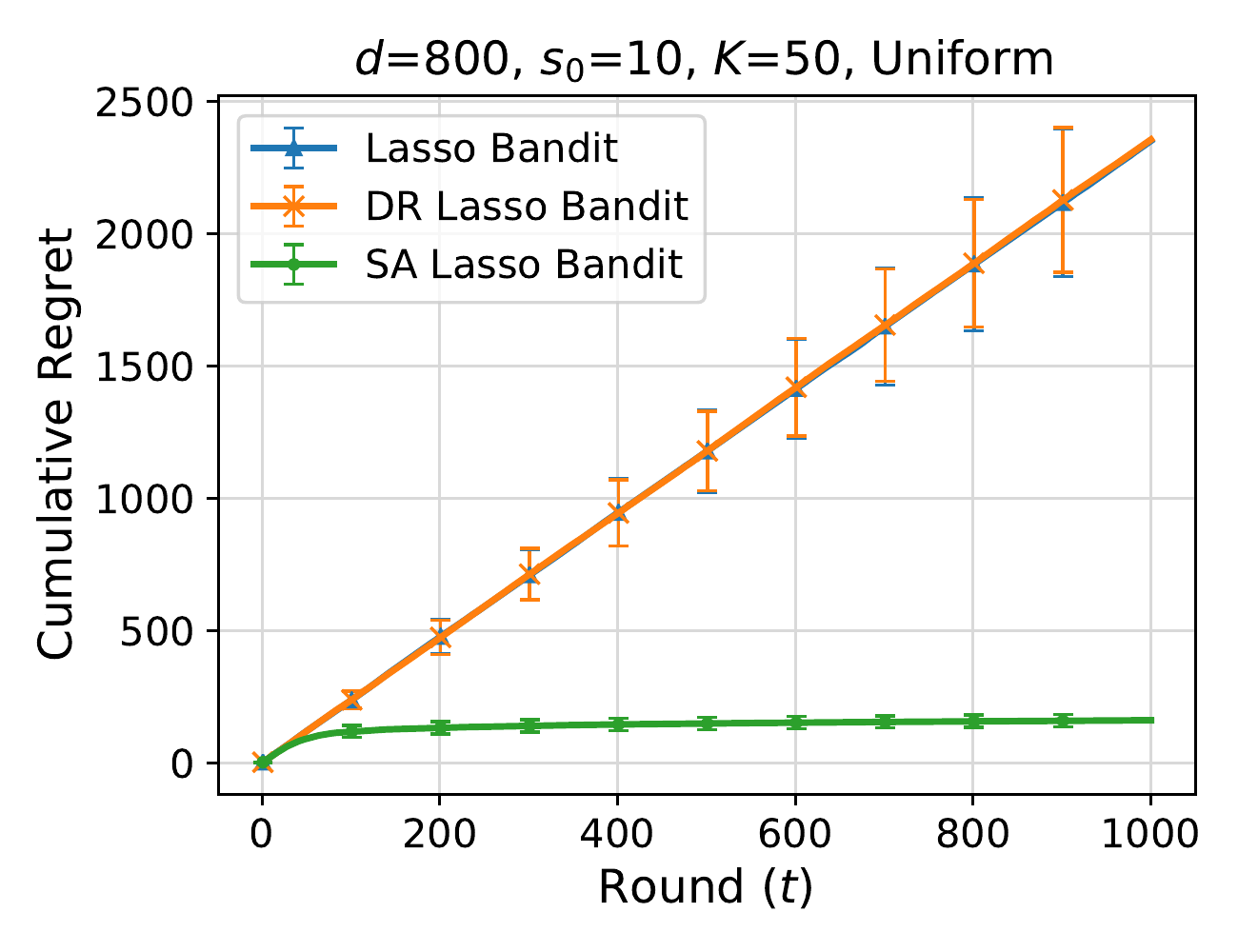}
    \end{subfigure}\\
    \begin{subfigure}[b]{0.24\textwidth}
        \includegraphics[width=\textwidth]{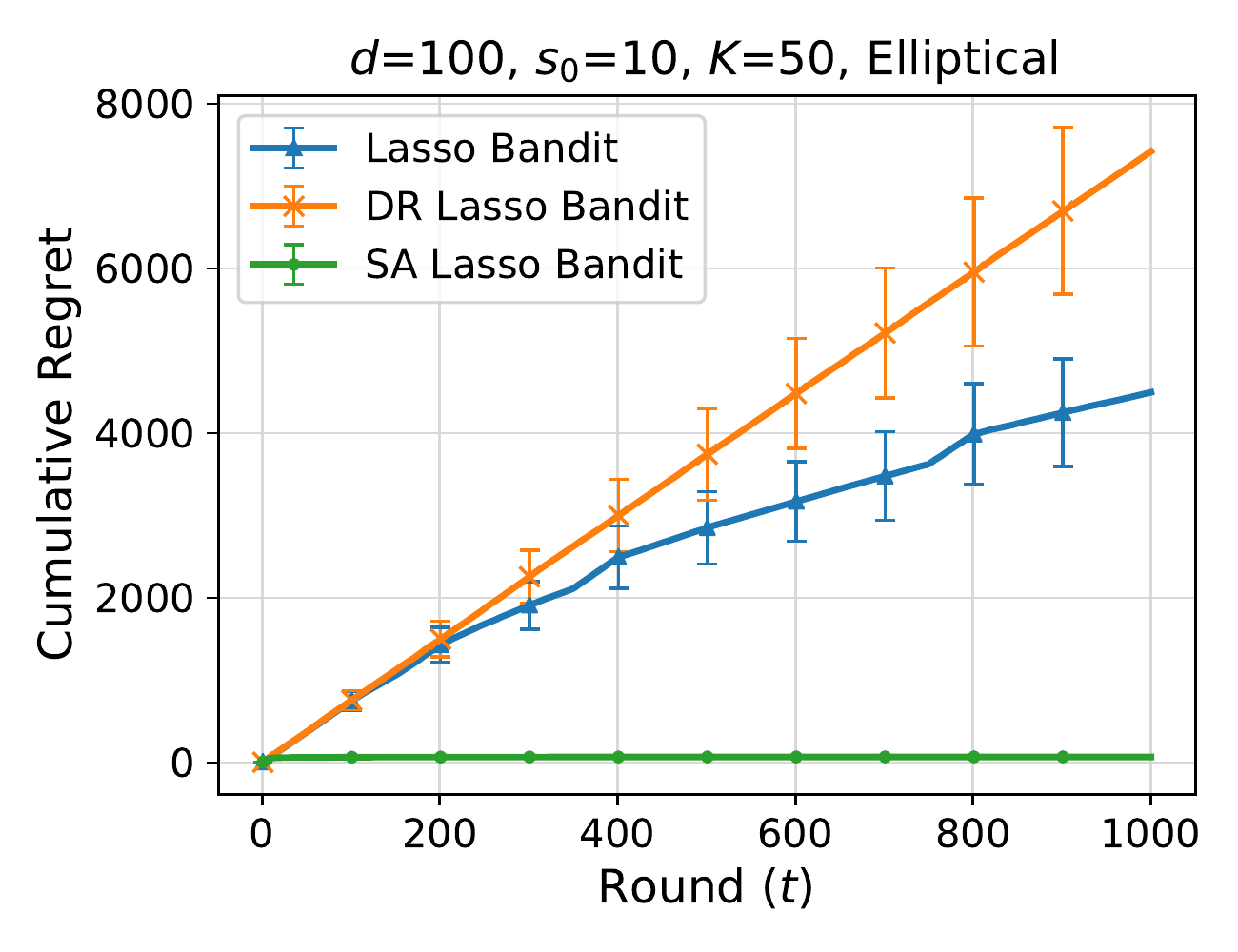}
    \end{subfigure}
    \begin{subfigure}[b]{0.24\textwidth}
        \includegraphics[width=\textwidth]{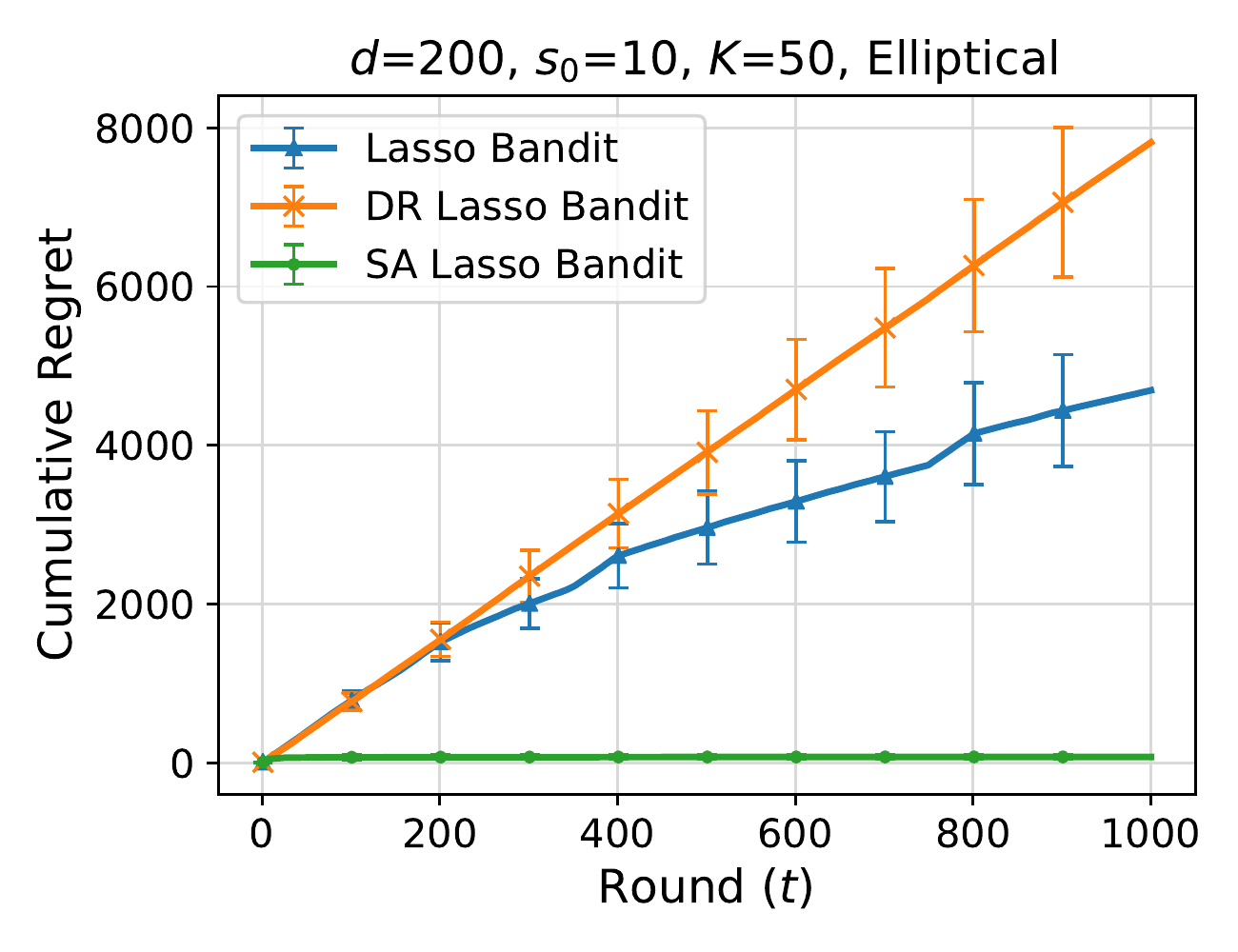}
    \end{subfigure}
    \begin{subfigure}[b]{0.24\textwidth}
        \includegraphics[width=\textwidth]{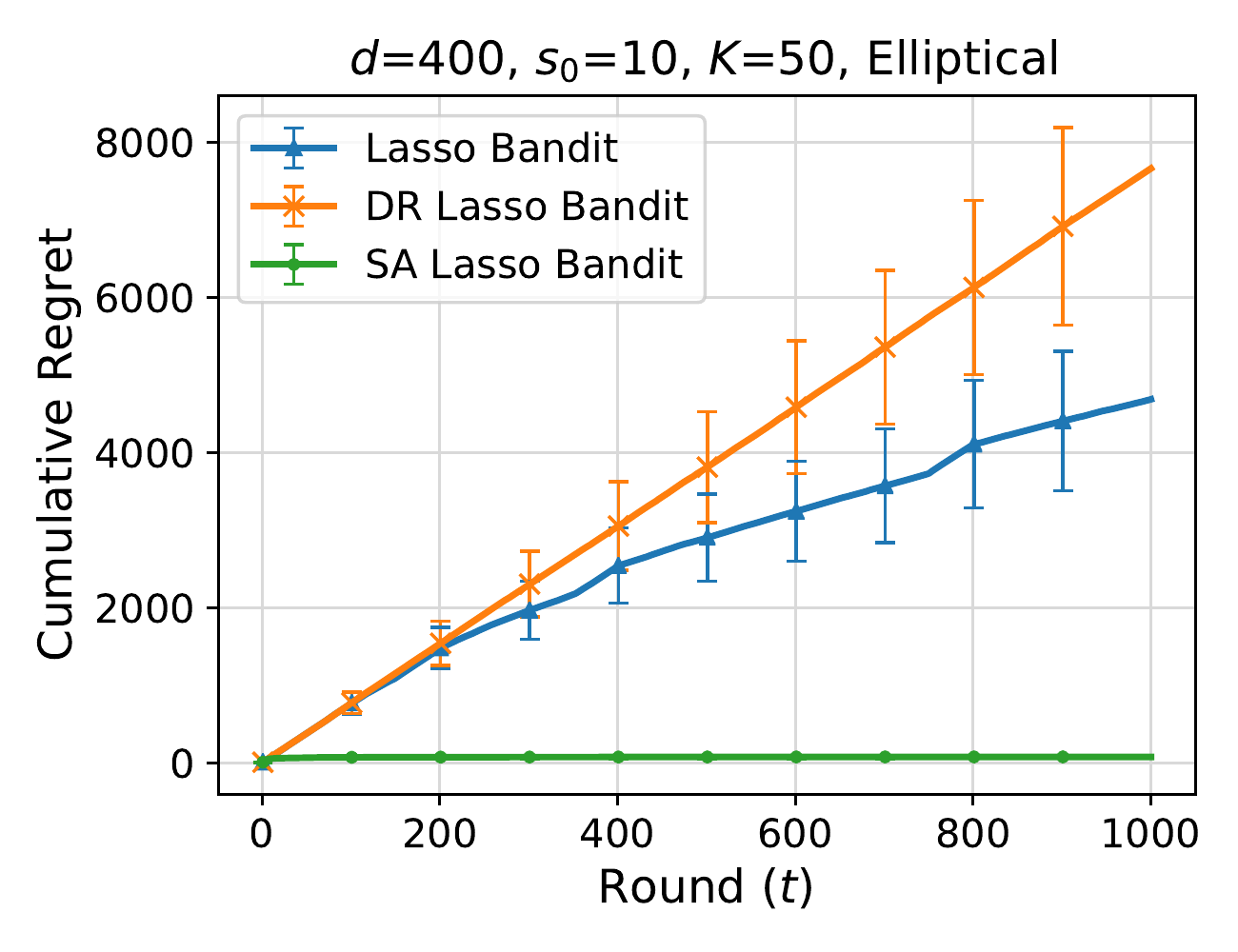}
    \end{subfigure}
    \begin{subfigure}[b]{0.24\textwidth}
        \includegraphics[width=\textwidth]{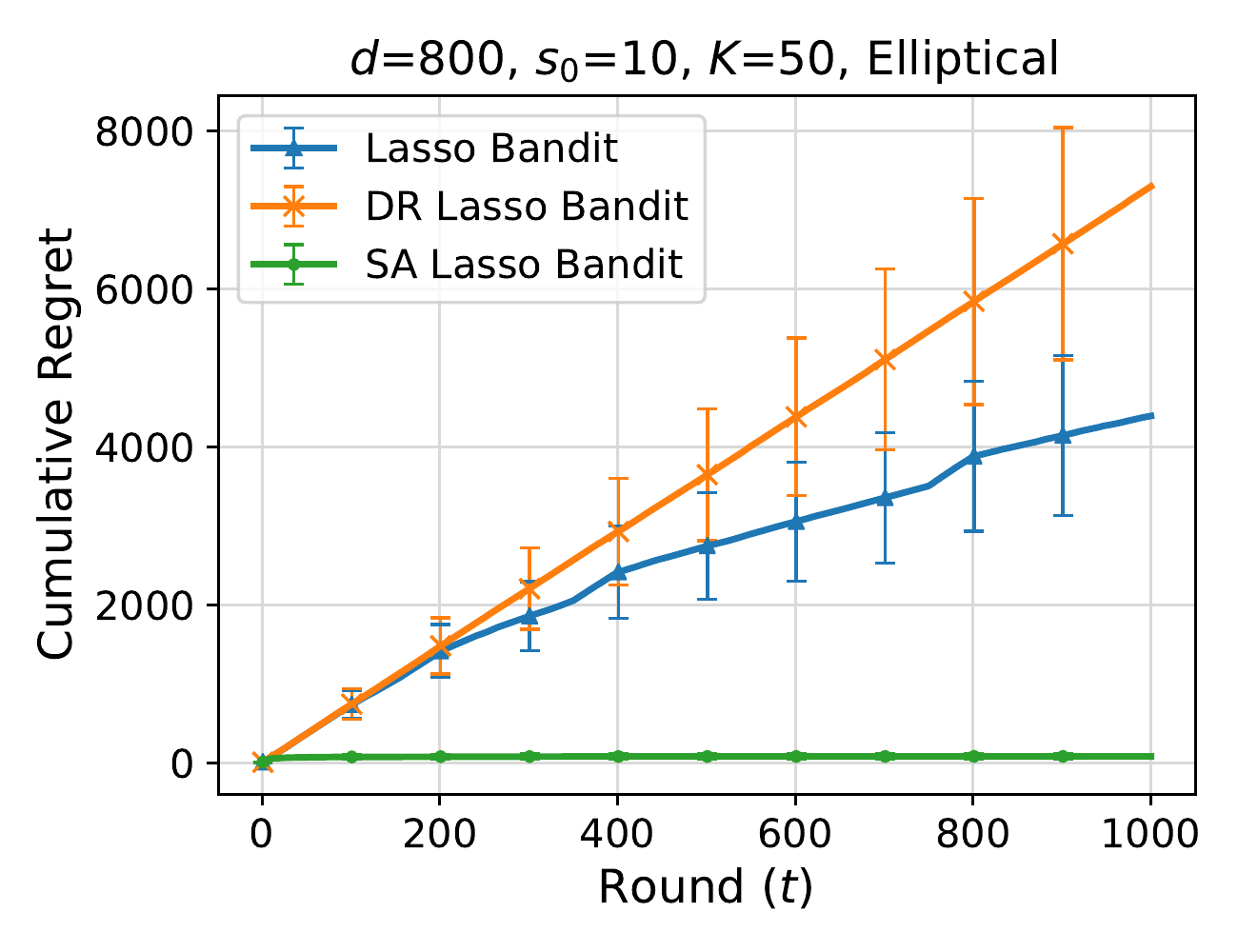}
    \end{subfigure}
    \caption{\small The plots show the $t$-round regret of \textsc{SA Lasso Bandit} (Algorithm~\ref{algo:SA_Lasso_bandit}), \textsc{DR Lasso Bandit} \citep{kim2019doubly}, and \textsc{Lasso Bandit} \citep{bastani2020online} for $K = 50$ and $s_0 = 10$. The first three rows are the results with features drawn from multivariate Gaussian distributions with varying levels of correlation between arms $\rho^2 \in \{0, 0.3, 0.7\}$. In the fourth row, features are drawn from a multi-dimensional uniform distribution. In the fourth row, features are drawn from a non-Gaussian elliptical distribution. For each row, we present evaluations for varying feature dimensions, $d \in \{100, 200, 400, 800\}$.} 
     \label{fig:K_arm_additional}
\end{figure*}


\end{document}